\documentclass[acmsmall]{acmart}

\AtBeginDocument{%
  \providecommand\BibTeX{{%
    \normalfont B\kern-0.5em{\scshape i\kern-0.25em b}\kern-0.8em\TeX}}}

\setcopyright{acmlicensed}
\copyrightyear{2024}
\acmYear{2024}
\acmDOI{XXXXXXX.XXXXXXX}

\acmISBN{978-1-4503-XXXX-X/18/06}
\usepackage{amsmath, amssymb, amsfonts}
\usepackage{graphicx}
\usepackage{float}
\usepackage{soul}
\usepackage{wrapfig}
\usepackage{dblfloatfix}
\usepackage{enumitem}
\usepackage{bm}
\usepackage{mathrsfs}
\usepackage[ruled,vlined]{algorithm2e}
\usepackage{multirow}
\usepackage{nicematrix}
\usepackage{pifont}
\usepackage[skip=3pt]{caption}
\usepackage{booktabs}
\usepackage{varwidth}
\usepackage{subcaption}
\usepackage{balance}

\SetArgSty{textnormal}

\renewcommand\vec{\mathbf}
\newtheorem{theorem}{Theorem}[section]

\newtheorem{example}{Example}[section]

\newcommand*{\expec}{\mathbb{E}}
\begin{document}

%\title{PathletRL: Trajectory Pathlet Dictionary Construction using Reinforcement Learning}
%\title{PathletRL++: Enhancing Trajectory Pathlet Extraction and Dictionary Formation via Reinforcement Learning}
%\title{PathletRL++: Memory-efficient Trajectory Pathlet Extraction and Dictionary Formation}
%\title{PathletRL++: Memory-efficient Scalable Trajectory Pathlet Extraction and Dictionary Formation}
%\title{PathletRL++: Optimizing Trajectory Pathlet Extraction and Dictionary Formation}
%\title{PathletRL++: A Bottom-up Approach to Trajectory Pathlet Extraction and Dictionary Formation}
%\title{PathletRL++: Optimizing Trajectory Pathlet Dictionaries via Reinforcement learning}
%\title{PathletRL++: A Reinforcement Learning-based Method for Optimizing Trajectory Pathlet Dictionary Construction}
%\title{PathletRL++: Memory-efficient Trajectory Pathlet Extraction and Dictionary Formation via Reinforcement Learning}
\title{PathletRL++: Optimizing Trajectory Pathlet Extraction and Dictionary Formation via Reinforcement Learning}


\author{Gian Alix}
\affiliation{
    \institution{York University}
    \city{Toronto}
    \country{Canada}
}
\email{gcalix@eecs.yorku.ca}
\authornote{Both authors contributed equally to this research.}

\author{Arian Haghparast}
\affiliation{
    \institution{York University}
    \city{Toronto}
    \country{Canada}
}
\email{arianhgh@my.yorku.ca}
\authornotemark[1]

\author{Manos Papagelis}
\affiliation{
    \institution{York University}
    \city{Toronto}
    \country{Canada}
}
\email{papaggel@eecs.yorku.ca}

\renewcommand{\shortauthors}{Alix, Haghparast and Papagelis}

\begin{abstract}
Advances in tracking technologies have spurred the rapid growth of large-scale trajectory data. Building a compact collection of pathlets, referred to as a \textit{trajectory pathlet dictionary}, is essential for supporting mobility-related applications. Existing methods typically adopt a top-down approach, generating numerous candidate pathlets and selecting a subset, leading to high memory usage and redundant storage from overlapping pathlets. To overcome these limitations, we propose a bottom-up strategy that incrementally merges basic pathlets to build the dictionary, reducing memory requirements by up to 24,000 times compared to baseline methods. The approach begins with unit-length pathlets and iteratively merges them while optimizing utility, which is defined using newly introduced metrics of \textit{trajectory loss} and \textit{representability}. We develop a deep reinforcement learning framework, \textsc{PathletRL}, which utilizes Deep Q-Networks (\textsc{DQN}) to approximate the utility function, resulting in a compact and efficient pathlet dictionary. Experiments on both synthetic and real-world datasets demonstrate that our method outperforms state-of-the-art techniques, reducing the size of the constructed dictionary by up to 65.8\%. Additionally, our results show that only half of the dictionary pathlets are needed to reconstruct 85\% of the original trajectory data. Building on \textsc{PathletRL}, we introduce \textsc{PathletRL++}, which extends the original model by incorporating a richer state representation and an improved reward function to optimize decision-making during pathlet merging. These enhancements enable the agent to gain a more nuanced understanding of the environment, leading to higher-quality pathlet dictionaries. \textsc{PathletRL++} achieves even greater dictionary size reduction, surpassing the performance of \textsc{PathletRL}, while maintaining high trajectory representability.
\end{abstract}



%%\received{20 February 2007}
%%\received[revised]{12 March 2009}
%%\received[accepted]{5 June 2009}

\maketitle
\begin{figure}[t]
    \setlength{\belowcaptionskip}{-7pt}
    \centering
    \includegraphics[width=0.8\textwidth]{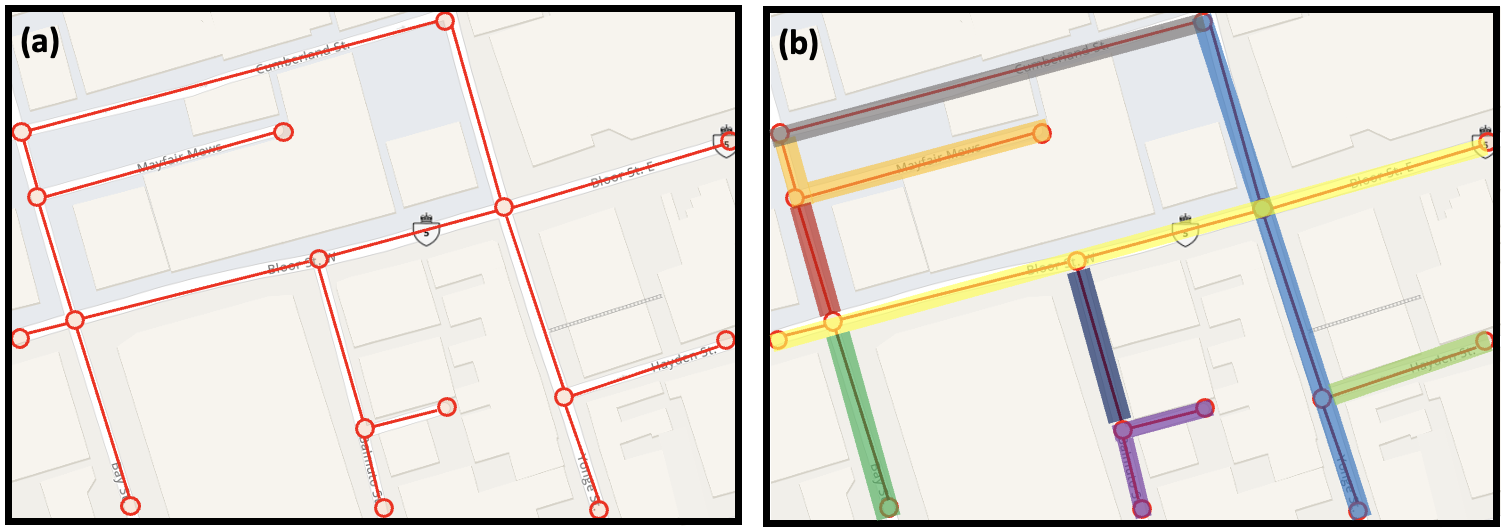}
    \caption{(a) Graph representation of a small area in Toronto\protect\footnotemark; (b) Example of various-length edge-disjoint pathlets in (a).}
    \label{fig:pathlets-graph}
\end{figure}

\section{Introduction}
\label{sec:intro}

\smallskip\noindent\textbf{Motivation \& Problem of Interest.} The development of technology for gathering and tracking location data has led to the accumulation of vast amounts of trajectory data, consisting of spatial and temporal information of moving objects, such as persons or vehicles. Mining trajectory data to find interesting patterns is of increased research interest due to a broad range of useful applications, including analysis of transportation systems \cite{nematichari2022evaluating}, human mobility  \cite{sawas2019versatile}, location-based services \cite{yassin2015lbs}, spatiotemporal epidemics \cite{pechlivanoglou2018centrality, alix2022stripe, pechlivanoglou2022microscopic} and more. There are several technical problems in trajectory data mining that researchers and practitioners have focused on in recent years, including trajectory similarity \cite{fang2022similarity}, clustering \cite{han2022clustering}, classification \cite{alsaeed2023tulhor},  prediction \cite{xue2022auxmoblcast} and simplification \cite{wang2021simplification}.
\footnotetext{Downtown Toronto maps taken from \url{https://www.mapquest.com/}}

% , outlier detection.
A few comprehensive surveys on the topic can be found in Zheng \cite{zheng2015trajectory}, Alturi et al. \cite{alturi2018spatiotemporal}, and Hamdi et al.  \cite{hamdi2022spatiotemporal}. In this research, we focus on the problem of constructing a small set of basic building blocks that can represent a wide range of trajectories, known as a {\em (trajectory) pathlet dictionary} (PD). The term {\em pathlet} appears in the literature by many names, such as \textit{subtrajectories}, \textit{trajectory segments}, or \textit{fragments} \cite{chen2013pathlet, agarwal2018subtrajectory, li2019personalized, zhao2018destination, sankararaman2013segmentation, panagiotakis2012representativeness}. For consistency, we will use the term \textit{pathlets} to denote these building blocks. 

\smallskip\noindent\textbf{The Broader Impact.} Effectively constructing pathlet dictionaries is of increased research and practical interest due to a broad range of tasks and applications that can use it, such as route planning \cite{yadav2020routeplanning}, travel time prediction \cite{han2021mdarnn}, personalized destination prediction \cite{xu2021destination}, trajectory prediction \cite{xue2022auxmoblcast}, and trajectory compression \cite{zhao2018trajcompression} (see Appendix \ref{sec:applications} for a supplementary discussion of these applications).

\smallskip\noindent\textbf{The State of the Art \& Limitations.} Many existing works frame the problem of analyzing and deriving pathlets as a (sub)trajectory clustering problem, where (sub)trajectory clusters represent popular paths (the pathlets) \cite{lee2007traculus,  vankrevald2011wellvisited, agarwal2018subtrajectory}. A few works considered an integer programming formulation with constraints to solve the problem \cite{chen2013pathlet, li2019personalized}. Some works designed their pathlets based on a route ``representativeness'' criterion \cite{panagiotakis2012representativeness, wang2022representative}. 
% While there has been a considerable number of works related to pathlets, 
Unfortunately, these existing works suffer some limitations. For example, Chen et al. \cite{chen2013pathlet} assumes that the datasets used are noise-free. Zhou et al.'s \cite{zhou2008bagofsegments} bag-of-segments method requires that trajectory segments are of fixed length. Van Krevald et al.  \cite{vankrevald2011wellvisited} demands input trajectories to have the same start/endpoints. The cluster centroids in (sub)trajectory clustering methods \cite{lee2007traculus, vankrevald2011wellvisited, agarwal2018subtrajectory, wang2019kpaths} do not necessarily reflect real roads in the road network. In addition, Wang et al. \cite{wang2022representative} demonstrated empirically that these clustering methods are computationally slow.  In spite of runtime improvements, \cite{wang2022representative} also requires the user to provide some budget constraint $B$ in the route representative discovery task, a domain-specific parameter that requires domain expert knowledge. Another related method is {\sc Traculus} \cite{lee2007traculus} that requires pathlets to be straight line segments, which is not always the case in real road maps.
In addition, all these works do not constraint pathlets to be {\em edge-disjoint}; two pathlets are said to be edge-disjoint if they don't share any edge. Therefore, existing works allow pathlets in the dictionary to (partially) overlap. These methods, by design, follow a top-down approach in constructing a dictionary. 
% This consists first forming all possible pathlet candidates, by considering pathlets of different configurations and sizes, and then eliminating candidates to form a smaller size pathlet dictionary that consists of only the most important ones (e.g., the most popular). 
This involves forming all possible pathlet candidates first, by considering pathlets of various configurations and sizes, and then eliminating candidates to form a smaller sized dictionary that consists of only the most important ones (e.g., the most popular). 
While simple and intuitive, its main limitation is the need for a large memory to initially store the large number of pathlets, most of which are redundant. This also limits its applicability in real-world settings, particularly when dealing with large road networks and trajectory data, as the number of initial candidates can quickly become overwhelming.
% Thus, our approach that considers disjoint pathlets utilizes a bottom-up strategy, where our initial PD consists of all possible unit-length sized pathlets and the PD is reduced by aggregating together certain (neighboring) pathlets that aim to maximize utility, according to concepts introduced in the maximal utility theory \cite{aleskerov2007util, mccormick1997util}. As such, our method, in contrast to prior works, does not require a large memory storage to store all its initial pathlets.

%\smallskip\noindent\textbf{Our approach.} We propose \textsc{PathletRL} (\textbf{\underline{Pathlet}} dictionary construction using trajectories with \textbf{\underline{R}}einforcement \textbf{\underline{L}}earning) as our solution to the problem of interest. Our methods utilize a merging-based algorithm to aggregate disjoint pathlets to form longer ones. Longer pathlets are important (over shorter ones) because they hold more spatiotemporal information, such as mobility patterns in trajectories \cite{chen2013pathlet}. We iteratively merge (neighboring) pathlets to form longer and higher-ordered pathlets based on some utility function. In our work, we employ a reinforcement learning (RL) method based on Deep Q Networks (DQN) to compute (approximate) such utility function, expressed as the RL's reward function. Using deep learning also distinguishes our work with existing methods. 

\begin{figure}[t]
    \centering
    \includegraphics[width=0.6\textwidth]{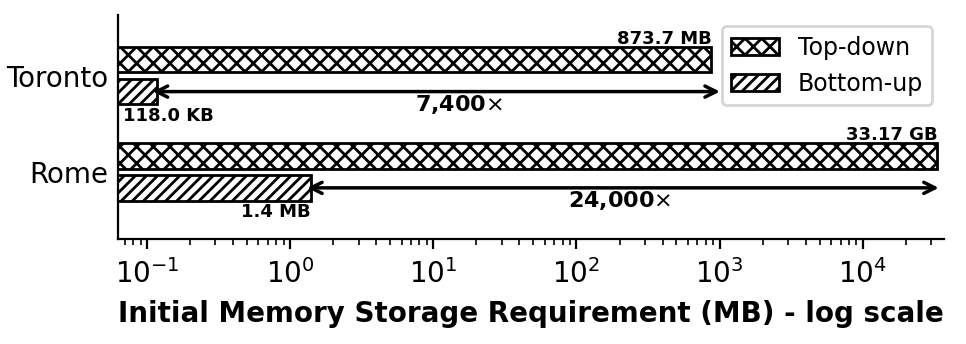}
    \caption[Initial Memory Storage Requirements (Horizontal Bar Plot)]{The memory required by top-down (existing) methods that use overlapping pathlets can be reduced by our proposed bottom-up solution that use edge-disjoint pathlets. \vspace{6pt}}
    \label{fig:memory-plot}
\end{figure}

% \vspace{-3.5pt}
\smallskip\noindent\textbf{Our Approach \& Contributions.} To address these limitations, we propose a bottom-up approach for constructing a pathlet dictionary that complies with edge-disjoint pathlets (see Fig. \ref{fig:pathlets-graph}) and reduces memory storage requirement. In Fig. \ref{fig:memory-plot} for instance, we illustrate how our proposed approach saves up to $\sim$24K$\times$ less memory space than existing methods for storing the initial pathlets (see Experiment \textbf{(Q2)} for full details, with Appendix \ref{sec:memory-proof} presenting a more theoreotical proof). The key idea of our approach is to initialize unit-length pathlets \& iteratively merge them to form longer, higher-order ones, while maximizing utility \cite{aleskerov2007util, mccormick1997util}. Longer pathlets are preferred (over shorter ones) as they hold more spatiotemporal information, such as mobility patterns in trajectories \cite{chen2013pathlet}. A deep reinforcement learning method is proposed to approximate the utility function.

A summary of our contributions is provided below:
\begin{itemize}[leftmargin=*]
    \itemsep 0em
    
    \item We introduce a more strict definition of a pathlet than in previous works to comply with edge-disjoint pathlets. This enables a bottom-up approach for constructing pathlet dictionaries that reduces memory storage needs.

    \item We introduce two novel metrics, namely \textit{trajectory loss} and \textit{trajectory representability}, which allow us to more comprehensively evaluate the utility of a pathlet and the overall quality of a constructed pathlet dictionary.

    \item We formulate the problem of \textit{pathlet dictionary construction} as a utility maximization problem, where shorter pathlets are merged to form a set of longer ones with higher utility.

    \item We propose \textsc{PathletRL}, a deep reinforcement learning method that utilizes a Deep $Q$ Network (\textsc{DQN}) policy to approximate the utility function of constructing a pathlet dictionary. To the best of our knowledge, this is the first attempt to employ a deep learning method for the problem.

    \item We address the limitations of the original \textsc{PathletRL} model by introducing optimization techniques to improve the quality and stability of the constructed pathlet dictionaries, ensuring better trade-offs between key objectives in trajectory modeling.

    \item We introduce \textsc{PathletRL++}, which extends \textsc{PathletRL} by incorporating a richer state representation and a more robust reward function to optimize decision-making during pathlet merging. This extension significantly improves the stability and quality of the constructed pathlet dictionary.

    \item We demonstrate empirically that the dictionary constructed by our \textsc{PathletRL} and its extensions is of superior quality compared to those constructed by traditional non-learning-based methods. Our method reduces the size of the dictionary by up to 65.8\% compared to other methods. Moreover, using only half of the pathlets in the dictionary suffices to reconstruct 85\% of the original trajectory data.

    \item We open-source our code to encourage reproducibility\protect\footnotemark.
\end{itemize}

\footnotetext{\url{https://github.com/Arianhgh/PathletRL}}

%\smallskip\noindent\textbf{Paper organization.} We organize the remainder of the paper as follows. Preliminary definitions and formalization of the problem is stated in Sec. \ref{sec:prelim}. In Sec. \ref{sec:methodology}, we provide thorough details of our methods. We then describe our experimental setup and provide discussions based on the these results in Sec. \ref{sec:evaluation}. We then review some related work in Sec. \ref{sec:related-work}, and finally conclude in Sec. \ref{sec:conclusions}.

\smallskip\noindent\textbf{Paper Organization.} The remainder of the paper is organized as follows. Section \ref{sec:prelim} presents preliminaries and a formal definition of the problem. Section \ref{sec:methodology} outlines our methodology and the details of the proposed approach. Section \ref{sec:evaluation} describes the experimental setup, presents the results, and provides insightful discussions. Section \ref{sec:pathletrl-optimization} introduces optimizations and extensions to the original \textsc{PathletRL} framework. We review related work in Section \ref{sec:related-work} and conclude in Section \ref{sec:conclusions}.

% \footnotetext{The link to the source code repository is in stealth mode due to anonymity.}
% \footnotetext{ Github repository released upon paper acceptance. } 

\section{Preliminaries \& Problem Definition}
%\section{Preliminaries and the Problem}
\label{sec:prelim}

In this section, we briefly introduce some definitions and notations (see Table \ref{tab:nomenclature}). Then we formally define the problem of interest.

% \setlength\belowcaptionskip{0pt}
% \setlength{\textfloatsep}{0pt}
% \begin{table}[t]
%     \centering
%      \setlength{\tabcolsep}{1pt}
%     \begin{tabular}{cl}
%         \toprule
%          \textbf{Symbol} & \multicolumn{1}{c}{\textbf{Definition}}  \\
%          \midrule
%          $\tau$ and $\mathcal{T}$         & A trajectory $\tau$ and a set of trajectories $\mathcal{T}$ \\
%         $\mathcal{G} \langle \mathcal{V}, \mathcal{E} \rangle$ & Road network with intersections $\mathcal{V}$ and road segments  $\mathcal{E}$ \\
%          $\rho$ and $\mathcal{P}$         & A pathlet $\rho$ and a pathlet set $\mathcal{P}$ \\
%          $\mathcal{G}_p \langle \mathcal{V}_p, \mathcal{E}_p \rangle$ & The pathlet graph representation of road network $\mathcal{G}$ \\
%          $\Phi(\tau)$                     & The pathlet-based representation of a trajectory $\tau$ \\
%          $\Lambda(\rho)$                  & The trajectory traversal set of a pathlet $\rho$ \\
%          $\mu(\tau)$                      & The trajectory representability of trajectory $\tau$ \\
%          $L_{traj}$                       & The trajectory loss \\
%          $\mathbb{S}$                     & The pathlet dictionary \\
%          $\phi$                           & The avg \# of pathlets representing each $\tau$ in $\mathcal{T}$ \\
%          \bottomrule
%     \end{tabular}
%     \caption[Summary of notation]{Summary of notation used in this work}
%     \label{tab:nomenclature}
% \end{table}

\setlength{\textfloatsep}{0pt}
\begin{table}[t]
    \centering
     \setlength{\tabcolsep}{1pt}
    \begin{tabular}{cl}
        \toprule
         \textbf{Symbol} & \multicolumn{1}{c}{\textbf{Definition}}  \\
         \midrule
         $\tau$ and $\mathcal{T}$         & A trajectory $\tau$ and a set of trajectories $\mathcal{T}$ \\
        $\mathcal{G} \langle \mathcal{V}, \mathcal{E} \rangle$ & Road network with intersections $\mathcal{V}$ and segments  $\mathcal{E}$ \\
         $\rho$ and $\mathcal{P}$         & A pathlet $\rho$ and a pathlet set $\mathcal{P}$ \\
         $\mathcal{G}_p \langle \mathcal{V}_p, \mathcal{E}_p \rangle$ & The pathlet graph representation of road network $\mathcal{G}$ \\
         $\Psi(\rho)$                     & The neighbors of pathlet $\rho$ \\
         $\Phi(\tau)$                     & The pathlet-based representation of a trajectory $\tau$ \\
         $\Lambda(\rho)$                  & The trajectory traversal set of a pathlet $\rho$ \\
         $\mu(\tau)$                      & The trajectory representability of trajectory $\tau$ \\
         $L_{traj}$                       & The trajectory loss \\
         $\mathbb{S}$                     & The pathlet dictionary \\
         $\phi$                           & The avg \# of pathlets representing each $\tau$ in $\mathcal{T}$ \\
         \bottomrule
    \end{tabular}

    \caption[Summary of notation]{Summary of notation used in this work}
    \label{tab:nomenclature}
    
\end{table}

\subsection{Primary Definitions and Notations}
\label{sec:primary-def-not}

\smallskip\noindent\textbf{Definition 2.1 (Trajectory)}. Let $\vec{O} = \{ o_1, o_2, ..., o_{|\vec{O}|} \}$ be a set of moving objects in a certain geographic map $\mathcal{M} \subset \mathbb{R}^2$. A \textit{trajectory} $\tau$ of a single object $o \in \vec{O}$ can be represented as a sequence of time-enabled geo-coordinate points: $\tau = \left\langle \left( x_1, y_1, t_1 \right),  ...,  ( x_{|\tau|}, y_{|\tau|}, t_{|\tau|} )\right\rangle$, where each $x_i$ and $y_i$ represents $o$'s longitudinal and latitudinal coordinates at a specific time instance $t_i \in [0,T]$. We let $|\tau|$ be its length, or the \# of time-enabled points for the trajectory of $o$. Moreover, we let \textit{trajectory (data) set} $\mathcal{T}$ consist of all trajectories of all $o \in \vec{O}$: $\mathcal{T} = \bigcup_{\forall o \in \vec{O}} \, \mathcal{T}_o$, with $\mathcal{T}_o$ as the set of all $o$'s trajectories.

% Moreover, we can define a \textit{trajectory (data) set}. It is denoted as $\mathcal{T}$, and it consists of all the trajectories of all entities $p \in \vec{P}$, which is represented as: $\mathcal{T} = \bigcup_{\forall p \in \vec{P}} \, \tau_p =  \tau_1 \cup \tau_2 \cup ... \cup \tau_{|\vec{P}|}$.

% Moreover, we can define a \textit{trajectory (data) set}, denoted as $\mathcal{T}$. It consists of all the trajectories of all entities $o \in \vec{O}$, which is represented as: $\mathcal{T} = \bigcup_{\forall o \in \vec{O}} \, \mathcal{T}_o$; here, $\mathcal{T}_o$ is the set of all $o$'s trajectories.

\smallskip\noindent\textbf{Definition 2.2 (Road Network)}. We denote by $\mathcal{G} \langle \mathcal{V}, \mathcal{E} \rangle$ the \textit{road network} within map $\mathcal{M}$, where $\mathcal{V}$ and $\mathcal{E} \subseteq \mathcal{V} \times \mathcal{V}$ represents $\mathcal{G}$'s set of road intersections (nodes) and segments (edges) respectively.

% Trajectory points (GPS traces) outside $\mathcal{M}$ are filtered out as a preprocessing step. In addition, the remaining trajectories require to be map-matched, a common task that identifies the path on the road an object has taken given a sequence of GPS locations \cite{newson2009hmm}. Ideally, we prefer highly accurate map-matched data from GPS trajectory traces; however, this task is itself involved and is outside the paper's scope. Thus, we rely on existing methods \cite{lou2009stmatching, yuan2010ivmm} to handle map-matching. With map-matched data, we can now formalize the most fundamental building block in this work.

Trajectory points (GPS traces) outside $\mathcal{M}$ are filtered out as a preprocessing step. The remaining trajectories also require to be map-matched (see Appendix \ref{sec:map-matching} for details). With map-matched data, we can now formalize the fundamental building block in this work.

% \smallskip\noindent\textbf{Definition 2.3 (Pathlet)}. A \textit{pathlet}, denoted by $\rho$, is defined as any sub-path in the road network $\mathcal{G}$. In particular, we denote by $\mathcal{P}$ the set of pathlets of the road network $\mathcal{G}$.
\smallskip\noindent\textbf{Definition 2.3 (Pathlet)}. A \textit{pathlet} $\rho$ is defined as any sub-path in the road network $\mathcal{G}$, with $\mathcal{P}$ being the set of all such pathlets.

In our work, we consider \textit{edge-disjoint} pathlets, s.t. no two $\rho_1, \rho_2 \in \mathcal{P}$ share any edge. For simplicity, we assume \textit{discrete} pathlets -- meaning they begin and end at an intersection (a node in the graph, with either start/endpoints at $\rho.s$/$\rho.e$), but the work can easily be generalized to include \textit{continuous} pathlets that drop this restriction.

\smallskip\noindent\textbf{Definition 2.4 (Pathlet Length\footnote{Not to confuse with a road segment's length that represents the measure depicting its actual physical distance, the pathlet length can be derived based on graph context.})}. Denoted by $\ell$, this represents pathlet $\rho$'s path length in the road network.

% The smallest unit of the pathlet has pathlet length $\ell = 1$. Moreover, we restrict pathlets by limiting the length of all $\rho \in \mathcal{P}$ to be $\ell \leq k$, for some user-defined parameter $k$. In this case, we say that $\mathcal{P}$ is a \textbf{$k$-order pathlet set}.
The smallest unit of the pathlet has length $\ell = 1$. Moreover, we restrict all pathlets $\rho \in \mathcal{P}$ to be of length $\ell \leq k$, for some user-defined $k$. In this case, we say that $\mathcal{P}$ is a \textbf{$k$-order pathlet set}.

\smallskip\noindent\textbf{Definition 2.5 (Pathlet Graph)}. The \textit{pathlet graph} $\mathcal{G}_p \langle \mathcal{V}_p, \mathcal{E}_p \rangle$ of a road network $\mathcal{G} \langle \mathcal{V}, \mathcal{E} \rangle$ depicts the road network's pathlets, where the road intersections represent the nodes $\mathcal{V}_p \subseteq \mathcal{V}$ and the road segments connecting road intersections as the edges $\mathcal{E}_p \subseteq \mathcal{E}$. 
% (usually but not necessarily with traffic lights) 

Fig. \ref{fig:pathlets-graph}(a), for example, represents the pathlet graph representation of a small area in Toronto. In our work, we consider an initial pathlet graph where each pathlet has length $\ell = 1$. 

\smallskip\noindent\textbf{Definition 2.6 (Pathlet Neighbors)}. Given a pathlet $\rho \in \mathcal{P}$, its neighbor pathlets, denoted by $\Psi(\rho)$, are all other pathlets $\rho' \in \mathcal{P} \setminus \{\rho\}$ who share the same start/endpoints with that of $\rho$:
$$\Psi(\rho) = \bigcup_{\rho' \in \mathcal{P} \setminus \{\rho\}, \, (\rho.s \in \{\rho'.s, \rho'.e\}) \lor (\rho.e \in \{\rho'.s, \rho'.e\})} \rho'$$
For example, the grey pathlet in Fig. \ref{fig:pathlets-graph}(b) has two neighbors: the orange \& blue pathlets.

\smallskip\noindent\textbf{Definition 2.7 (Pathlet-based Representation of a Trajectory)}. A trajectory $\tau \in \mathcal{T}$ can be represented based on some subset of pathlets $\mathcal{P}_{sub} \subseteq \mathcal{P}$. Moreover, the pathlets in $\mathcal{P}_{sub}$ can be concatenated in some sequence resulting into the path traced by $\tau$ on the road network $\mathcal{G}$. We denote this by $\Phi(\tau) = \{ \rho^{(1)}, \rho^{(2)}, ..., \rho^{(|\mathcal{P}_{sub}|)} \}$, where $\rho^{(i)} \in \mathcal{P}_{sub}$ denotes the $i$th pathlet in the sequence that represents the pathlet-based representation for $\tau$.

Based on this, it is also possible to define a trajectory's pathlet length, which we initially set before constructing the pathlet graph. Each trajectory $\tau \in \mathcal{T}$ has pathlet length equal to $\sum_{\forall \rho \in \Phi(\tau)} \ell(\rho)$, whose value remains static for the rest of our algorithm. 

% Based on this, it is also possible to define a trajectory's pathlet length -- which we initially set up before constructing the pathlet graph. Each trajectory $\tau \in \mathcal{T}$ would have a pathlet length equal to $\sum_{\forall \rho \in \Phi(\tau)} \ell(\rho)$, whose value remains static for the rest of our algorithm. 
% We then define the \textit{trajectory reconstruction set} of a pathlet -- that is related to the pathlet-based representation of a trajectory.

% \smallskip\noindent\textbf{Definition 2.8 (Trajectory Reconstruction Set of a Pathlet)}. We denote by the function $\Lambda(\rho)$ to be the set of all trajectories $\forall \tau \in \mathcal{T}$ that can be reconstructed from the given pathlet $\rho \in \mathcal{P}$. This can be written as $\Lambda(\rho) = \{ \tau \, | \, \forall \tau \in \mathcal{T}, \, \rho \in \Phi(\tau)  \}$. In other words, all trajectories in $\Lambda(\rho)$ \textit{pass through} pathlet $\rho$. Therefore, function $\Lambda(\cdot)$ is the inverse operation of $\Phi(\cdot)$.
% \smallskip\noindent\textbf{Definition 2.8 (Trajectory Spanning Set of a Pathlet)}. Let $\Lambda(\rho)$ be the set of all trajectories $\tau \in \mathcal{T}$ that \textit{pass through} or \textit{span} pathlet $\rho \in \mathcal{P}$. This can also be written as $\Lambda(\rho) = \{ \tau \, | \, \forall \tau \in \mathcal{T}, \, \rho \in \Phi(\tau)  \}$.
\smallskip\noindent\textbf{Definition 2.8 (Trajectory Traversal Set of a Pathlet)}. Let $\Lambda(\rho)$ be the set of all trajectories $\tau \in \mathcal{T}$ that \textit{pass} or \textit{traverse} pathlet $\rho \in \mathcal{P}$. This can also be written as $\Lambda(\rho) = \{ \tau \, | \, \forall \tau \in \mathcal{T}, \, \rho \in \Phi(\tau)  \}$.
% In other words, all trajectories in $\Lambda(\rho)$ \textit{pass through} pathlet $\rho$. Therefore, function $\Lambda(\cdot)$ is the inverse operation of $\Phi(\cdot)$.

% Related to this is the road segments' weights in the road network (i.e., the \textit{pathlet weights} $\omega$). 
% We can also assigned weights $\omega$ to pathlets. In the unweighted case, all pathlets are weighed equally. Meanwhile in the weighted version, pathlets are weighed equal to the \# of trajectories passing through a pathlet, i.e., $\omega(\rho) = |\Lambda(\rho)|$, or $\omega(\rho) = |\Lambda(\rho)|/|\mathcal{T}|$ when normalized. These weights indicate each pathlet's importance in the road network/pathlet graph. 

% We can also assign weights $\omega$ to pathlets. 
We can also assign weights $\omega$ to pathlets. In the unweighted case, all pathlets are weighed equally; while in the weighted version, pathlets are weighed equal to the \# of trajectories traversing a pathlet, i.e., $\omega(\rho) = |\Lambda(\rho)|$, or  $\frac{|\Lambda(\rho)|}{|\mathcal{T}|}$ when normalized. These weights indicate each pathlet's importance in the road network/pathlet graph.
%  $\omega(\rho) = |\Lambda(\rho)|/|\mathcal{T}|$

% This then brings us two novel metrics related to trajectories, namely the \textit{trajectory representability} and the \textit{trajectory loss}, that are unique to our work in contrast to existing methods.

\subsection{Novel Trajectory Metrics}
\label{sec:novel-metrics}

We can now introduce some novel metrics to allow us to more comprehensively evaluate our pathlets and pathlet dictionaries.

\smallskip\noindent\textbf{Definition 2.9 (Trajectory Representability\protect\footnotemark)}. The (trajectory) representability $\mu \in [0\%,100\%]$ of a trajectory $\tau$ denotes the \% of $\tau$ that can be represented using pathlets in pathlet set $\mathcal{P}$.

\footnotetext{Not to be confused with \textit{representativeness} that describes the capability of a trajectory to represent other similar nearby trajectories, \textit{representability} depicts how much a trajectory can be reconstructed given our pathlets.}

% Clearly, the pathlet-based representation of $\tau$ is directly related to its representability, i.e., $\mu(\tau) = |\Phi(\tau)|/\sum_{\forall \rho \in \Phi(\tau)} \ell(\rho)$. However, in the weighted case, some pathlets are weighed more than others. Therefore, a trajectory $\tau$'s representability may need to be redefined based on the pathlet's weightings, i.e., $\mu(\tau) = \sum_{\forall \rho \in \Phi(\tau)} \omega(\rho)$.

Clearly, the pathlet-based representation of $\tau$ is directly related to its representability, i.e., \\ {$\mu(\tau) = \sum_{\forall \rho \in \Phi(\tau)} \frac{\ell(\rho)}{|\tau|}$}, for the unweighted case and $\mu(\tau)$ = $\sum_{\forall \rho \in \Phi(\tau)} \omega(\rho)$, for the weighted version.

\smallskip\noindent\textbf{Definition 2.10 (Trajectory Loss)}. We define the \textit{trajectory loss} $L_{traj}$ to be the \# of trajectories $\forall \tau \in \mathcal{T}$ that have representability value $\mu = 0\%$, i.e., $L_{traj} = |\{ \tau | \tau \in \mathcal{T}, \, \mu(\tau) = 0 \} |$. We can also describe these trajectories as "lost" or "discarded" from the given trajectory set $\mathcal{T}$, and we may also depict this number as a \%.

% \smallskip\noindent We refer the reader to a toy example of how relevant and impactful these metrics are in pathlets and in trajectories in Appendix \ref{sec:toy-examples}. 

The relevance and impact of these two metrics will become clear as we go over the methodology in finer details.

\subsection{Problem Definition}
\label{sec:problem-def}

Before formalizing the problem, we first introduce the \textit{pathlet dictionary} and some optimization definitions.

% \smallskip\noindent\textbf{Definition 2.11 (Pathlet Dictionary)}. A (trajectory) pathlet dictionary (PD) is a data structure composed of pathlets $\rho \in \mathcal{P}$ (keys), and each one's trajectory reconstruction set $\Lambda(\rho)$ (values).
\smallskip\noindent\textbf{Definition 2.11 (Pathlet Dictionary)}. A (trajectory) pathlet dictionary (PD) is a data structure that stores pathlets $\rho \in \mathcal{P}$ (keys), and their associated trajectory traversal set $\Lambda(\rho)$ (values).

\begin{figure*}[t]
    \centering
    \includegraphics[width=\textwidth]{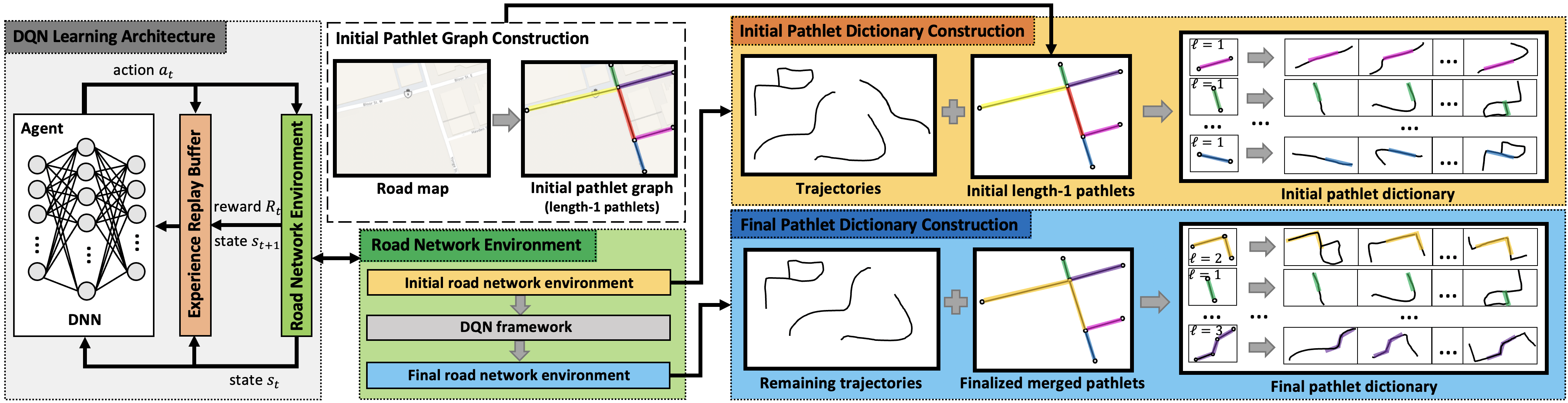}
    \caption{The overall architecture (including the constructed PDs) of our proposed \textsc{PathletRL} model \vspace{12pt}}
    \label{fig:architecture}
\end{figure*}

% Refer to Fig. \ref{fig:architecture} (the righthand boxes inside the orange and blue panels) for an illustrative example of a PD. We are more interested in the construction of a PD that aims to achieve one or a combination of the following objectives:
See Fig. \ref{fig:architecture} (the righthand boxes inside the orange \& blue panels) for an illustrative example of a PD. We are interested in constructing a PD that aims to achieve one or a combination of the following:

\begin{enumerate}[label=\textbf{(O\arabic*)}]
    \item Minimal size of candidate pathlet set $\mathbb{S}$, or the candidate set with the least possible number of pathlets (i.e., $\min |\mathbb{S}|$)
    \item Minimal $\phi$, or the avg \# of pathlets representing each trajectory $\tau \in \mathcal{T}$ (i.e., $\min \phi = \min \frac{1}{|\mathcal{T}|} \sum_{\tau \in \mathcal{T}} |\Phi(\tau)|$)
    \item Minimal trajectory loss $L_{traj}$  (i.e., $\min L_{traj}$)
    \item Maximal $\bar{\mu}$, or the average representability values of the remaining trajectories in $\mathcal{T}$ (i.e., $\max \bar{\mu} = \max \frac{1}{|\mathcal{T}|} \sum_{\tau \in \mathcal{T}} \mu(\tau)$)
\end{enumerate}

In other words, the objective function that we wish to optimize is based on the four objectives above -- which can be modelled by:
% is given by the following mathematical formulation:
\begin{equation}
\small
    \min_{\sum_i \alpha_i = 1} \left( \alpha_1 |\mathbb{S}| + \alpha_2 \frac{1}{|\mathcal{T}|} \sum_{\tau \in \mathcal{T}} |\Phi(\tau)| + \alpha_3 L_{traj} - \alpha_4 \frac{1}{|\mathcal{T}|} \sum_{\tau \in \mathcal{T}} \mu(\tau) \right)
    \label{eq:opt-constraint}
\end{equation}
where the $\alpha_i$'s are user-defined objective weights.
% where $\{ \alpha_i \}$, with $\sum_{i=1}^4 \alpha_i$ = 1, are user-defined parameters depicting the weights of our four objectives.  We now formalize our problem.
% of interest.
% We are then ready to formalize our problem of interest.

\smallskip\noindent \textbf{Problem 1 (Pathlet Dictionary Construction)}. Given a road network $\mathcal{G}\langle \mathcal{V}, \mathcal{E} \rangle$ of a specific map $\mathcal{M}$, a trajectory set $\mathcal{T}$, max pathlet length $k$, max trajectory loss $M$, and avg trajectory representability threshold $\hat{\mu}$, construct a $k$-order pathlet dictionary $\mathbb{S}$. The dictionary $\mathbb{S}$ consists of edge-disjoint pathlets with lengths of at most $k$, and achieves the max possible utility according to some utility function as depicted in Equation (\ref{eq:opt-constraint}), such that $L_{traj} < M$ and $\bar{\mu} \geq \hat{\mu}$.

\setlength{\textfloatsep}{0pt}
{
% \SetAlgoNoLine% 
% \SetAlgoLined\SetArgSty{}
\LinesNumbered

\begin{algorithm}[H]
\caption{PathletRL: Trajectory Pathlet Dictionary Construction}
\label{alg:pathlet_rl}
\KwIn{
  Road Network $\mathcal{G}(V, E)$, 
  Trajectory Set $\mathcal{T}$,  
  Maximum Pathlet Length $k$, 
  Trajectory Loss Threshold $M$, 
  Average Representability Threshold $\mu_{\text{threshold}}$, 
}
\KwOut{Pathlet Dictionary $\mathcal{S}$ for Representing $\mathcal{T}$.}

\textbf{Initialization:} \\
Set the candidate set $\mathcal{C} = \emptyset$ for storing processed pathlets \\
Construct the pathlet graph $\mathcal{G}_p$ from the input road network $\mathcal{G}$ \\
Identify neighbors for each pathlet and compute initial pathlet weights $\omega_p$ \\
For each trajectory $t_i \in \mathcal{T}$, compute initial trajectory coverage $\phi$, representability $\mu$ \\
Randomly select an initial pathlet $p_{\text{current}} \in \mathcal{G}_p \setminus \mathcal{C}$ \\

\While{termination conditions not met}{
    \If{Trajectory loss $L_{\text{traj}}$ exceeds $M$ or $\mu < \mu_{\text{threshold}}$ or $\mathcal{G}_p \setminus \mathcal{C} = \emptyset$}{
        \textbf{Terminate} the episode and store the final pathlet dictionary $\mathcal{S}$ \;
    }
    
    Observe current state $s_t$ of $\mathcal{G}_p$ \;
    Select action $a_t$ using the $\epsilon$-greedy policy from DQN \;
    
    \eIf{$a_t = 0$ (\textbf{Skip action})}{
        Mark current pathlet $p_{\text{current}}$ as processed by adding it to $\mathcal{C}$ \;
        \If{$\mathcal{G}_p \setminus \mathcal{C} \ne \emptyset$}{
            Randomly select a new pathlet $p_{\text{current}} \in \mathcal{G}_p \setminus \mathcal{C}$ \;
        } \Else{
            \textbf{Terminate} the episode and store the final pathlet dictionary $\mathcal{S}$ \;
        }
    }{
        Select neighboring pathlet $p_{\text{neighbor}}$ based on action $a_t$ \;
        \If{Merging $p_{\text{current}}$ and $p_{\text{neighbor}}$ results in path length $\leq k$}{
            Merge pathlets to form a new pathlet $p_{\text{merged}}$ \;
            Update traversal set $\Lambda(p_{\text{merged}})$ and pathlet graph $\mathcal{G}_p$ \;
            Recompute trajectory coverage $\phi$, representability $\mu$, and trajectory loss $L_{\text{traj}}$ \;
            Remove trajectories with zero coverage from $\mathcal{T}$ \;
            Update the set of unprocessed pathlets $\mathcal{G}_p \setminus \mathcal{C}$ \;
            Set $p_{\text{current}} = p_{\text{merged}}$ \;
        } \Else{
            Mark $p_{\text{current}}$ as processed by adding it to $\mathcal{C}$ \;
            \If{$\mathcal{G}_p \setminus \mathcal{C} \ne \emptyset$}{
                Randomly select a new pathlet $p_{\text{current}} \in \mathcal{G}_p \setminus \mathcal{C}$ \;
            } \Else{
                \textbf{Terminate} the episode and store the final pathlet dictionary $\mathcal{S}$ \;
            }
        }
    }
    
    Compute reward $R_t$ based on changes in size $|\mathcal{S}|$, representability $\mu$, trajectory loss $L_{\text{traj}}$, and average number of pathlets per trajectory $\phi$ \;
    Store transition $(s_t, a_t, R_t, s_{t+1})$ in experience replay buffer \;
    Perform DQN update on the Q-network using sampled experiences \;
    
    Set next state $s_{t+1}$ \;
}
\textbf{Output:} Pathlet dictionary $\mathcal{S}$ 
\end{algorithm}

\section{Methodology}
\label{sec:methodology}

We now describe our methods to address the problem of interest (see Fig. \ref{fig:architecture} for its architecture). In particular, we describe the components of the proposed \textsc{PathletRL} model (\textbf{\underline{Pathlet}} dictionary construction using trajectories with \textbf{\underline{R}}einforcement \textbf{\underline{L}}earning). There are two main components: (1) the method responsible for extracting the candidate pathlet sets through a merging-based process, and (2) a deep reinforcement learning-based architecture for approximating the utility function of the merging process of (1).
% that is utilized in the algorithm presented by 
% that utilizes a Deep $Q$ Network policy

\subsection{Extracting Candidate Pathlets}
\label{sec:extract-candidate-pathlets}

In this section, we describe the algorithmic details for merging our edge-disjoint pathlets. The high-level idea of the algorithm is based on the theory of maximal utility \cite{mccormick1997util, aleskerov2007util}, i.e., iteratively merging (neighboring) pathlets until this brings forth little to no improvement on the \textit{utility} (details of the utility are given later). The algorithm takes in as input a road network $\mathcal{G}$, a trajectory set $\mathcal{T}$ operating within $\mathcal{G}$, the max threshold for the trajectory loss $M$, the trajectory representability threshold $\hat{\mu}$, and a positive integer $k$ denoting the desired $k$-order pathlet graph. As output, it returns a pathlet dictionary (PD) that holds pathlet information as described in Definition 2.11. The extracted PD aims to satisfy the four objectives \textbf{(O1)-(O4)}. See Algorithm \ref{alg:pathlet_rl} for the pseudocode.

\smallskip\noindent \textbf{Initialization}. The algorithm begins by setting up an empty candidate set $\mathcal{C} = \emptyset$ to keep track of processed pathlets. It constructs the initial pathlet graph $\mathcal{G}_p$ from the input road network $\mathcal{G}$, which consists of initial pathlets of length 1. For each trajectory $t_i \in \mathcal{T}$, it calculates the initial trajectory coverage $\phi$ and representability $\mu$. An initial pathlet $p_{\text{current}}$ is randomly chosen from the set of unprocessed pathlets $\mathcal{G}_p \setminus \mathcal{C}$.

\smallskip\noindent \textbf{An Iterative Algorithm}. The algorithm operates in a loop where, at each iteration, the agent observes the current state $s_t$ of the environment and selects an action $a_t$ using an $\epsilon$-greedy policy from the Deep Q-Network (DQN). The actions are defined as follows:

\begin{itemize}
    \item \textbf{Skip Action ($a_t = 0$)}: The agent chooses to skip the current pathlet $p_{\text{current}}$, marking it as processed by adding it to $\mathcal{C}$. If there are still unprocessed pathlets remaining (i.e., $\mathcal{G}_p \setminus \mathcal{C} \ne \emptyset$), the agent randomly selects a new pathlet from this set to be the next $p_{\text{current}}$. Otherwise, the episode terminates, and the final pathlet dictionary $\mathcal{S}$ is produced.
    \item \textbf{Merge Action ($a_t > 0$)}: The agent selects a neighboring pathlet $p_{\text{neighbor}}$ based on the action $a_t$ to merge with the current pathlet $p_{\text{current}}$. If merging these two pathlets results in a new pathlet whose length does not exceed the maximum allowed length $k$, they are merged to form a new pathlet $p_{\text{merged}}$. The traversal set $\Lambda(p_{\text{merged}})$ and the pathlet graph $\mathcal{G}_p$ are updated accordingly. Trajectories that now have zero coverage are removed from the trajectory set $\mathcal{T}$. The metrics $\phi$, $\mu$, and trajectory loss $L_{\text{traj}}$ are recalculated. The agent then updates $p_{\text{current}}$ to be the newly formed $p_{\text{merged}}$. If merging is not possible due to length constraints, the agent treats this as a skip action. At each iteration, the agent computes a reward $R_t$ based on changes in the size of the pathlet dictionary $|\mathcal{S}|$, representability $\mu$, trajectory loss $L_{\text{traj}}$, and the average number of pathlets per trajectory $\phi$. The transition $(s_t, a_t, R_t, s_{t+1})$ is stored in the experience replay buffer, and the DQN updates its Q-network using sampled experiences. The loop continues until one of the termination conditions is met: the trajectory loss exceeds the threshold $M$, the average representability $\mu$ falls below the threshold $\mu_{\text{threshold}}$, or there are no unprocessed pathlets left. The final pathlet dictionary $\mathcal{S}$.
\end{itemize}

\smallskip\noindent \textbf{The Utility Function}. To complete the description of the algorithm, we discuss the formulation of the utility function that we approximated using a learning-based method.

In particular, a reinforcement learning method is utilized to learn the (sequence of) actions (i.e., merge or don't merge pathlets) that would yield the highest possible utility. Specifically, we frame the utility function as a \textit{reward function} that we aim to optimize (i.e., maximize). We discuss the details of this design in the next section.

\begin{figure}[t]
    \centering
    \includegraphics[width=\textwidth]{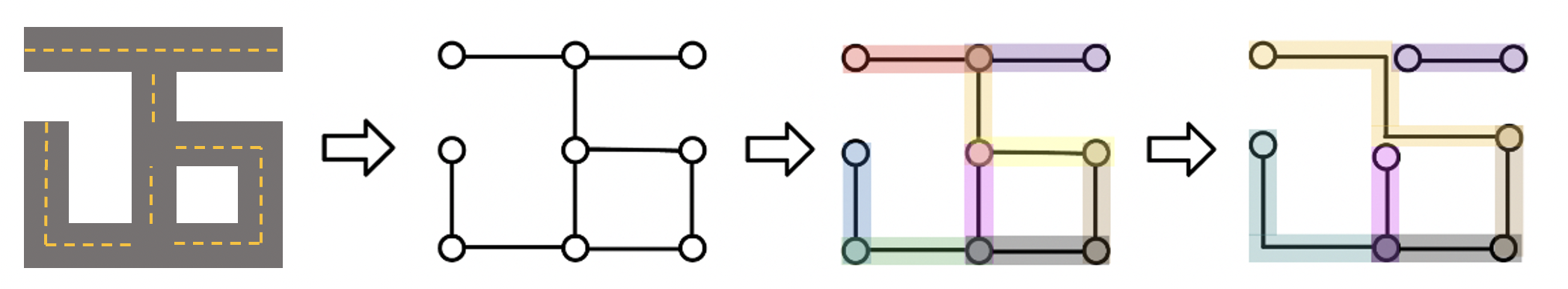} \\
    (a) \hspace{32mm} (b) \hspace{32mm} (c) \hspace{32mm} (d)
    \caption[Road Network to Pathlet Graph Representation for Example \ref{ex:pathlet-merge}]{An illustrative example of Example \ref{ex:pathlet-merge}: (a) A toy example of a simple road network; (b) A grid representation of (a); (c) The initial pathlet graph representation (of length-1 pathlets) for the road network in (a); (d) The final (merged) pathlet graph representation after the completion of the pathlet-merging algorithm in Algorithm \ref{alg:pathlet_rl}.}
    \label{fig:pathlet-merge-pic1}
\end{figure}

\begin{figure}[t]
    \centering
    \includegraphics[width=\textwidth]{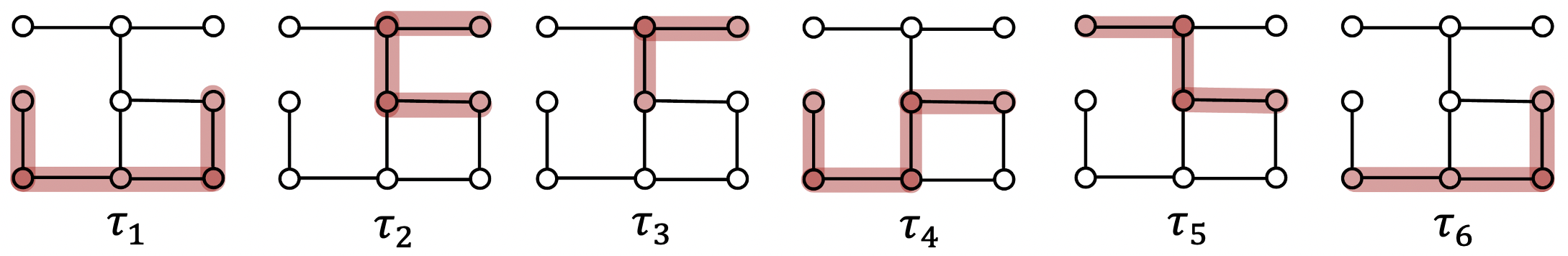} 
    \caption[Trajectory Paths (Road Segments) for Example \ref{ex:pathlet-merge}]{An illustrative example of the paths (road segments) traversed by six trajectories $\{ \tau_1, \tau_2, \tau_3, \tau_4, \tau_5, \tau_6 \}$ (highlighted by maroon) from the road network of Example \ref{ex:pathlet-merge} as seen in Figure \ref{fig:pathlet-merge-pic1}; see Table \ref{tab:pathlet-rep-set-toy} that lists the sequential pathlet-based representations of each trajectory. \vspace{12pt}}
    \label{fig:pathlet-merge-pic2}
\end{figure}

With the pathlet dictionary constructed following this process of pathlet merging based on the utility theory, we can attain something like the one depicted in Figure \ref{fig:architecture} (bottom right figure). Now we provide a simple toy example of the algorithm.

\begin{example}
    As edge-disjoint pathlets are non-overlapping, then a decision would have to be made when a pathlet is about to merge with one of its neighbor pathlets. As such, there is some advantages or gains by merging with a certain pathlet, but such action also comes with some cost. This is where trajectory representabilities and losses come in. More specifically, a trajectory's representability will drop when a portion of its trajectory cannot be represented due to a pathlet merging with one of its neighbors. To give a more concrete example, consider the snippet of a road network in Figure \ref{fig:pathlet-merge-pic1}(a). Figure \ref{fig:pathlet-merge-pic1}(b) illustrates its grid representation, and Figure \ref{fig:pathlet-merge-pic1}(c) the (initial) pathlet graph representation, where we highlighted using various colors the nine pathlets in question. See the left image of Figure \ref{fig:pathlet-labels-toy} for the labels of these pathlets and their color code in Table \ref{tab:pathlet-color-code}. Initially, our pathlet dictionary is composed of the following based on the six trajectories as highlighted by the maroon color on the grid graph in Figure \ref{fig:pathlet-merge-pic2} (their pathlet-based representations in Table \ref{tab:pathlet-rep-set-toy}): 

    \quad \quad $\rho_1 : \{ \tau_5 \}$  \quad $\rho_2 : \{ \tau_2, \tau_3 \}$ \quad $\rho_3 : \{ \tau_2, \tau_3, \tau_5 \}$ \quad $\rho_4 : \{ \tau_2, \tau_4, \tau_5 \}$ \quad $\rho_5 : \{ \tau_1, \tau_4 \}$
    
    \quad \quad  $\rho_6 : \{ \tau_4 \}$ \quad $\rho_7 : \{ \tau_1, \tau_6 \}$ \quad  $\rho_8 : \{ \tau_1, \tau_4, \tau_6 \}$ \quad $\rho_9 : \{ \tau_1, \tau_6 \}$

    \noindent After the entire merging process, the algorithm returns the following final dictionary -- with its illustration in Figure \ref{fig:pathlet-merge-pic1}(d), with their labels in Figure \ref{fig:pathlet-labels-toy}, their color codes in Table \ref{tab:pathlet-color-code} and the trajectories' updated representabilities and pathlet-based representations in Table \ref{tab:pathlet-rep-set-toy}:

    \begin{minipage}{0.3\textwidth}
    \begin{table}[H]
        \centering
        \begin{tabular}{cc}
            \toprule[1.5pt]
            \textbf{Pathlet} & \textbf{Color} \\
            \midrule
            $\rho_1$ & red \\
            $\rho_2$ & purple \\
            $\rho_3$ & orange \\
            $\rho_4$ & yellow \\
            $\rho_5$ & blue \\
            $\rho_6$ & pink \\
            $\rho_7$ & brown \\
            $\rho_8$ & green \\
            $\rho_9$ & grey \\
            $\rho_{134}$ & tangerine \\
            $\rho_{58}$ & aquamarine \\
            \bottomrule[1.5pt]
        \end{tabular}
        \caption[Color Coding of Pathlets for Example \ref{ex:pathlet-merge}]{Color coding of the pathlets for the toy example in Example \ref{ex:pathlet-merge}}
        \label{tab:pathlet-color-code}
    \end{table}
\end{minipage}
\hfill
\begin{minipage}{0.6\textwidth}
    \begin{table}[H]
        \centering
        \begin{tabular}{ccc}
            \toprule[1.5pt]
            \multirow{3}{*}{\textbf{Traj}} & \multicolumn{2}{c}{\textbf{Pathlet-based representation}} \\
            & \multicolumn{2}{c}{\textbf{set} $\Phi$ (\textbf{Representability} $\mu$)} \\
            \cmidrule{2-3}
            & \textbf{(Before merge)} & \textbf{(After merge)} \\
            \midrule
            \\[-0.75em]
            $\tau_1$ & $\{ \rho_5, \rho_8 , \rho_9, \rho_7 \}$ (100\%)  & $\{ \rho_{58}, \rho_9, \rho_7 \}$ (100\%) \\
            \\[-0.75em]
            $\tau_2$ & $\{ \rho_2, \rho_3, \rho_4 \}$ (100\%)  & $\{ \rho_2 \}$ (33\%) \\
            \\[-0.75em]
            $\tau_3$ & $\{ \rho_3, \rho_2 \}$ (100\%)  & $\{ \rho_2 \}$ (50\%) \\
            \\[-0.75em]
            $\tau_4$ & $\{ \rho_4, \rho_6, \rho_8, \rho_5 \}$ (100\%)  & $\{ \rho_6, \rho_{58} \}$ (75\%) \\
            \\[-0.75em]
            $\tau_5$  & $\{ \rho_4, \rho_3, \rho_1 \}$ (100\%)  & $\{ \rho_{134} \}$ (100\%) \\
            \\[-0.75em]
            $\tau_6$ & $\{ \rho_7, \rho_9, \rho_8 \}$ (100\%)   & $\{ \rho_7, \rho_9 \}$ (67\%) \\
            \\[-0.75em]
            \bottomrule[1.5pt]
        \end{tabular}
        \caption[Pathlet-based Representation of Trajectories for Example \ref{ex:pathlet-merge}]{Pathlet-based representation set (and trajectory representabilities) of the provided toy example in Example \ref{ex:pathlet-merge}}
        \label{tab:pathlet-rep-set-toy}
    \end{table}
\end{minipage}

\vspace{10pt}
    \quad \quad $\rho_{134} : \{ \tau_5 \}$ \quad $\rho_2 : \{ \tau_2, \tau_3 \}$ \quad $\rho_{58} :  \{ \tau_1, \tau_4 \}$ \quad $\rho_6 : \{ \tau_4 \}$ \quad $\rho_7 : \{ \tau_1, \tau_6 \}$ \quad $\rho_9 : \{ \tau_1, \tau_6 \}$
    
    \noindent Note that pathlets $\rho_1$, $\rho_3$, and $\rho_4$ from initial pathlet graph merged to become pathlet $\rho_{134}$ in the final pathlet graph (it is irrelevant in this example whether pathlet $\rho_3$ merged with $\rho_1$ or $\rho_4$ first). The same can be said for pathlets $\rho_5$ and $\rho_8$ to form pathlet $\rho_{58}$.

    Initially, all trajectories have $\mu = 100\%$ representatbility values because all six trajectories can be represented by pathlets in the initial pathlet dictionary. However, after the entire merging process, we are left with some of the original six trajectories to have a lower $\mu$ value than the original representability value. Notice for example trajectory $\tau_1$; its $\mu$ value did not drop because all the pathlets in its original pathlet-based representation $\Phi$ either have never merged, or have merged with a neighboring pathlet that also belongs to $\Phi(\tau_1)$. In other words, the entirety of trajectory $\tau_1$ can still be represented by the pathlets in the final pathlet dictionary; this results into its representability being maintained at 100\%. A similar story can be told for trajectory $\tau_5$. For the rest of the trajectories, the representabilities are lower. Looking at trajectory $\tau_2$ for example whose $\Phi(\tau_2) = \{ \rho_2, \rho_3, \rho_4 \}$. However, after the algorithm, we only have $\Phi(\tau_2) = \{ \rho_2 \}$ left; the reason being is that pathlets $\rho_1$, $\rho_3$, and $\rho_4$ have all merged together to form $\rho_{134}$. But note that since $\rho_1$ does not represent $\tau_2$, then the merged $\rho_{134}$ is not part of the $\Phi(\tau_2)$ after completing the iterative algorithm. As a result, it is expected for its trajectory representability to drop. The same story can be said for trajectories $\tau_3$, $\tau_4$ and $\tau_6$.
    \label{ex:pathlet-merge}
\end{example}
\begin{figure}
    \includegraphics[width=0.5\textwidth]{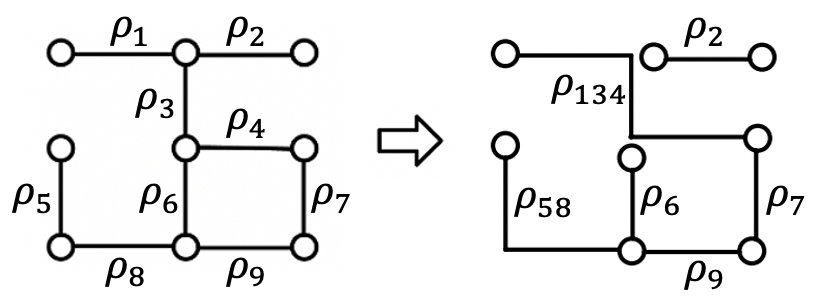}
    \caption[Pathlet Labels for Example \ref{ex:pathlet-merge}]{Pathlet labels for the initial and final pathlet graph representations}
    \label{fig:pathlet-labels-toy}
\end{figure}

\smallskip\noindent \textbf {Remark.}
    One can imagine that the representabilities of the trajectories can potentially drop at each step of the iterative algorithm, until it drops to zero. In the event that a trajectory's representabilty reaches zero, it is removed from the trajectory set and counted as a \textit{trajectory loss}. Now this is a soft version of what is considered to be a trajectory loss. There is a harder, stricter variation, where the notion of trajectory representability is eliminated from the picture, (i.e., a trajectory losing only a small portion of its representability due to a pathlet merge action meant that the entire trajectory is considered lost), entails throwing excessive trajectories and can be fatal to the performance of the model and the algorithm. Intuitively, one might not desire discarding an entire trajectory when (say only 1\% of the trajectory) cannot be represented due to the pathlet merge. The experimental evaluations later will demonstrate the essence of this soft version that takes trajectory representability into account.
    \label{remark:no-rep}

\begin{theorem}[Trajectory Representability Theorem]
    At any step $i$ of the iterative Algorithm \ref{alg:pathlet_rl}, then the trajectory representability $\mu$ of some trajectory $\tau \in \mathcal{T}$ by the end of that iteration $i$ is equal to:
    \begin{equation}
        \mu_i(\tau) = \dfrac{\sum_{\rho' \in \Phi_i(\tau)} \ell(\rho') }{ \sum_{\rho \in \Phi_0(\tau)} \ell(\rho) }
        \label{eq:rep-eq}
    \end{equation}
    where $\Phi_0$ and $\Phi_i$ are the pathlet-based representation of trajectory $\tau$ in the initial (iteration 0) and iteration $i$ of the iterative algorithm respectively.
    \label{thm:traj-rep-iter}
\end{theorem}

The above theorem provides a formula for how to compute a trajectory's representability value $\mu$ at some iteration $i$ of the algorithm. We refer the reader to Appendix \ref{sec:proof-thm-traj-rep-iter} where we provide a complete proof for this theorem.

\subsection{Background on Reinforcement Learning}
\label{sec:rl_background}

For readers already familiar with reinforcement learning and Deep Q-Networks (DQN), you may proceed directly to Section~\ref{sec:rl}. For those new to the topic, we provide a brief overview of essential concepts to facilitate understanding of our framework.

\smallskip\noindent \textbf{Markov Decision Process (MDP)}. An MDP models decision-making where outcomes are partly random and partly under the control of an agent. It consists of:
\begin{itemize}
    \item $\mathcal{S}$: Set of possible states.
    \item $\mathcal{A}$: Set of actions available to the agent.
    \item $\mathcal{P}(s' \,|\, s, a)$: Transition probability from state $s$ to $s'$ after action $a$.
    \item $\mathcal{R}(s, a, s')$: Immediate reward received after transitioning to state $s'$ from $s$ using action $a$.
    \item $\gamma \in [0, 1]$: Discount factor for future rewards.
\end{itemize}
The agent aims to learn a policy $\pi$ that maximizes the expected cumulative reward.

\smallskip\noindent \textbf{Policy ($\pi$)}. A policy maps states to actions. It can be deterministic or stochastic, assigning probabilities to actions. The goal is to find a policy that maximizes the expected return.

\smallskip\noindent \textbf{Value Functions}. The value function $V^{\pi}(s)$ estimates the expected return starting from state $s$ under policy $\pi$:
$$ V^{\pi}(s) = \mathbb{E}_{\pi} \left[ \sum_{k=0}^{\infty} \gamma^k r_{t+k} \,\bigg|\, s_t = s \right] $$
Similarly, the action-value function $Q^{\pi}(s, a)$ estimates the expected return after taking action $a$ in state $s$:
$$ Q^{\pi}(s, a) = \mathbb{E}_{\pi} \left[ \sum_{k=0}^{\infty} \gamma^k r_{t+k} \,\bigg|\, s_t = s, a_t = a \right] $$

\smallskip\noindent \textbf{Iterative Learning in Reinforcement Learning}. Reinforcement learning is inherently iterative. The agent interacts with the environment in a sequence of time steps. At each time step $t$:
\begin{enumerate}
    \item The agent observes the current state $s_t$.
    \item Based on its policy $\pi$, the agent selects an action $a_t$.
    \item The environment transitions to a new state $s_{t+1}$ according to the transition probability $\mathcal{P}(s_{t+1} \,|\, s_t, a_t)$.
    \item The agent receives a reward $r_t = \mathcal{R}(s_t, a_t, s_{t+1})$.
    \item The agent updates its policy $\pi$ to improve future actions, often using the observed reward and state transitions.
\end{enumerate}
Through this iterative process, the agent learns from the consequences of its actions, adjusting its policy to maximize cumulative rewards over time. This learning continues until the agent's policy converges to an optimal or satisfactory level of performance.

\smallskip\noindent \textbf{Q-Learning and Deep Q-Networks (DQN)}. Q-learning is an algorithm where the agent learns the optimal action-value function $Q^*(s, a)$ by iteratively updating its estimates based on the Bellman equation:
$$ Q(s_t, a_t) \leftarrow Q(s_t, a_t) + \alpha \left[ r_t + \gamma \max_{a'} Q(s_{t+1}, a') - Q(s_t, a_t) \right] $$
where $\alpha$ is the learning rate.

Deep Q-Networks (\textsc{Dqn}) extend Q-learning to handle large or continuous state spaces by using a neural network to approximate the $Q$-function. The iterative learning involves:
\begin{itemize}
    \item Collecting experiences $(s_t, a_t, r_t, s_{t+1})$ by interacting with the environment.
    \item Storing experiences in a replay buffer.
    \item Sampling mini-batches from the replay buffer to train the neural network.
    \item Updating the network parameters to minimize the difference between predicted and target $Q$-values.
\end{itemize}
This process allows the agent to learn effective policies even in complex environments.

\subsection{Reinforcement Learning Framework}
\label{sec:rl}

Building upon the reinforcement learning concepts introduced earlier, we now detail how we apply RL to the pathlet merging problem. Our goal is to train an agent to decide whether to merge or keep specific pathlets in the graph, optimizing the trade-off between the objectives mentioned in section \ref{sec:problem-def}.

\smallskip\noindent \textbf{Agent}. The agent interacts with the pathlet graph by selecting actions that alter its structure. It aims to learn a policy that maximizes cumulative rewards, effectively simplifying the graph while maintaining high trajectory representability.

\smallskip\noindent \textbf{States}. Each state $s_t$ captures the current configuration of the pathlet graph. We represent the state as a 4-tuple:
$$(S_1, S_2, S_3, S_4)$$
where:
\begin{itemize}
    \item $S_1$: Number of pathlets in the graph ($|\mathbb{S}|$).
    \item $S_2$: Average number of pathlets needed to represent trajectories.
    \item $S_3$: Trajectory loss metric $L_{traj}$, indicating the proportion of trajectories that cannot be accurately represented.
    \item $S_4$: Average trajectory representability $\bar{\mu}$.
\end{itemize}
This state representation encapsulates key factors affecting our optimization objectives.

\smallskip\noindent \textbf{The Actions}. At each time $t$, the agent has a choice of two discrete actions on the currently processed pathlet $\rho$, as expressed by the action space $\mathcal{A} = \{ \textsc{keep}, \textsc{merge}\}$. In other words, $\textsc{keep}$ action suggests that the current pathlet $\rho$ should be kept and not be merged with any one of its neighbors. As a result, Algorithm \ref{alg:pathlet_rl} puts the current pathlet $\rho$ in the processed set and then selects a new pathlet to process, performing one of the two actions in the action space on that new pathlet. The $\textsc{merge}$ action should however merge the current pathlet $\rho$ with one of its $|\Psi(\rho)|$ neighbors. For that, the agent would need to decide on which neighbor in the set $\Psi(\rho)$ to merge with. Thus, the action space can succinctly be written as:
$$\mathcal{A} = \bigcup_{\forall \hat{\rho} \in \Psi(\rho)} \textsc{merge}(\rho, \hat{\rho}) \cup \{ \textsc{keep}(\rho) \}$$

\vspace{-3pt}
\smallskip\noindent \textbf{The Reward Function}. We formulate our reward function $R$ based on the optimization equation defined in Equation (\ref{eq:opt-constraint}):
\begin{equation}
\footnotesize
    \max_{a_t} \expec \left[ \left( - \alpha_1 |\mathbb{S}| - \alpha_2  \frac{1}{|\mathcal{T}|} \sum_{\tau \in \mathcal{T}} |\Phi(\tau)| -\alpha_3 L_{traj} + \alpha_4 \frac{1}{|\mathcal{T}|} \sum_{\tau \in \mathcal{T}} \mu(\tau) \right) \right]
    \label{eq:reward}
\end{equation}
Whenever the RL agent performs an action $a_t$ in the pathlet graph environment, the environment provides back feedback to it in the form of instantaneous rewards $\{r_t\}_{t=0}^T$:
    $$r_t = -\alpha_1 \Delta |\mathbb{S}| - \alpha_2  \Delta \phi -\alpha_3 \Delta L_{traj} + \alpha_4 \Delta \bar{\mu}$$
where $\Delta \odot$ represents the change of $\odot$'s value in the previous and current timesteps. In the end, the agent receives the total sum of these instantaneous rewards, plus the final reward as depicted in Equation (\ref{eq:reward}). Note that in order for the agent to realize the importance of both immediate and long-term future rewards, a user-defined \textit{discounted rate factor} $\gamma \in [0,1]$ was introduced.

\smallskip\noindent \textbf{The Policy and DQN Implementation}. To handle the challenge of a continuous state space, we employ a Deep Q-Network (\textsc{Dqn}) to approximate the optimal action-value function. The agent's policy $\pi$ aims to select actions that maximize the expected cumulative reward:
    $$Q^{\pi}(s_t, a_t) = \mathbb{E} \left[ R_t \, | \, s_t, a_t \right]$$
Instead of maintaining a $Q$-table, which is impractical due to the size of our state and action spaces, the \textsc{Dqn} uses a neural network to estimate $Q$-values. The network takes the state $s_t$ as input and outputs estimated $Q$-values for possible actions $a_t$. By training this network, the agent learns an effective policy for deciding when to merge or keep pathlets, aiming to maximize the cumulative reward defined by our objectives.

\smallskip\noindent \textbf{Experience Replay and Training}. During training, the agent interacts with the environment and collects experiences in the form of transitions $(s_t, a_t, r_t, s_{t+1})$. To improve learning stability and efficiency, we utilize an experience replay buffer \cite{fedus2020replay} where these transitions are stored. The agent samples mini-batches from this buffer to train the neural network, which helps break the temporal correlations between sequential data. An $\epsilon$-greedy strategy is employed during data collection to balance exploration and exploitation: the agent selects a random action with probability $\epsilon$, and the action with the highest estimated $Q$-value otherwise. The network is trained to minimize the Huber loss between the predicted $Q$-values and the target $Q$-values computed from the sampled experiences. This loss function is distinct from our trajectory loss metric, which assesses the number of trajectories that cannot be represented by the current set of pathlets.

\subsection{Space Complexity Analysis}
\label{sec:space-analysis}

One of the main motivations for why edge-disjoint pathlets have been used together with bottom-up approaches is due to the reduced memory storage requirements that is necessary to initialize pathlets, in contrast with previous works that utilize top-down schemes with overlapping, redundant pathlets. In fact, we provide a space complexity analysis of top-down approaches and compare this with the proposed bottom-up methods through the following theorem (see Appendix \ref{sec:memory-proof} for the proof).

\begin{theorem}[Initial Memory Storage Requirement Theorem]
    The memory space that is required by top-down methods for initializing a pathlet dictionary has a quadratic $\Theta(n^2)$ bound, with $n$ as the number of segments of the road network. Bottom-up schemes on the other hand, such as the proposed \textsc{PathletRL}, requires only an initial $\Theta(n)$ amount of memory space, with $n$ as the number of initial length-1 pathlets.  
    \label{thm:space-analysis}
\end{theorem}

% \setlength\belowcaptionskip{0pt}
% \begin{table}[t]
% \centering
%   \begin{tabular}{c|lcc}
%     \toprule
%     \multicolumn{1}{c}{} & \multicolumn{1}{c}{\textbf{Feature}} & \textbf{\textsc{Toronto}} & \textbf{\textsc{Rome}} \\
%     \midrule
%     \parbox[t]{1mm}{\multirow{2}{*}{\rotatebox[origin=c]{90}{\textit{\scriptsize{roadmap}}}}} & \textbf{\# nodes} & 1.9K & 7.5K \\
%     \\[-0.8em]
%     & \textbf{\# edges / initial pathlets} & 2.5K & 15.4K \\
%     % \\[-0.8em]
%     \midrule
%     \parbox[t]{1mm}{\multirow{3}{*}{\rotatebox[origin=c]{90}{\textit{\scriptsize{trajectories}}}}} & \textbf{Trajectory type} & realistic synthetic & real world  \\
%     & \textbf{Object} & cars & taxis  \\
%     & \textbf{\# Total trajectories} & 169K & 3.8M \\
%   \bottomrule 
% \end{tabular}
% \caption{Dataset attributes}
% \label{tab:data-attributes}
% \end{table}

\section{Evaluation}
\label{sec:evaluation}

In this section, we present the details of our experimental setup for evaluating our proposed method. We aim to analyze and evaluate our models based on the following research questions:

\begin{enumerate}[leftmargin=*, label=\textbf{(Q\arabic*)}]
    \item How does \textsc{PathletRL} compare with the \textsc{SotA} methods, in terms of the quality of the extracted PDs?
    \item How much memory does the bottom-up approach save compared to top-down methods? 
    \item How much improvement and how much more effective is our \textsc{PathletRL} model against its ablation variations?
    \item What is the distribution of pathlet lengths in the obtained dictionary in our \textsc{PathletRL} model?
    \item How effective is the constructed PD in reconstructing the original trajectories?
    \item What is the sensitivity of the user-defined parameters [$\alpha_1$, $\alpha_2$, $\alpha_3$, $\alpha_4$] in the performance of our \textsc{PathletRL} model?
    \item How do the hyper-parameters such as the maximum length of pathlets and the average trajectory representability threshold impact the performance and efficiency of \textsc{PathletRL}?
\end{enumerate}

\subsection{Datasets}
\label{sec:datasets}

We utilize two datasets that each depict a different map scenario (see Appendix \ref{sec:dataset-stats} for its complete statistics, and Appendix \ref{sec:privacy-ethics} for a brief discussion about the data's privacy concerns). We used real world maps of two metropolitan cities, Toronto\footnote{\url{https://www.toronto.ca/city-government/data-research-maps/open-data/}} and Rome through the OpenStreetMaps\footnote{\url{https://www.openstreetmap.org/}}. A realistic synthetic vehicular mobility datasets for the \textsc{Toronto} map was generated using the \textsc{Sumo} mobility app simulator\footnote{\textbf{\underline{S}}imulation of \textbf{\underline{U}}rban \textbf{\underline{MO}}bility: \url{https://www.eclipse.org/sumo/}} (3.7 hrs). Moreover, larger-scale, real-world taxi cab trajectories (first week of February 2014) were taken from \textsc{Crawdad} \cite{roma-taxi-20140717}, an archive site for wireless network and mobile computing datasets, to form the \textsc{Rome} dataset. We split our trajectory sets into 70\% training and 30\% testing, where the training data was used to construct our pathlet dictionaries and the remainder for evaluation.

% \begin{table}[t]
%     \setlength{\tabcolsep}{3pt}
%     \centering
%     \begin{NiceTabular}{lccc}
%         \toprule
%         \Block[c]{}{\textbf{\textsc{PathletRL}} \\ \textbf{\textsc{Algorithm}}} & \Block[c]{}{\textbf{Representability} \\ \textbf{Measure}} & \Block[c]{}{\textbf{Weighted} \\ \textbf{Networks}} & \Block[c]{}{\textbf{Deep Learning} \\ \textbf{Policy}} \\
%         \midrule
%         \textsc{PathletRL-Nr} & \ding{55} & \ding{51} & \ding{51} \\
%         \textsc{PathletRL-Rnd} & \ding{51} & \ding{51} & \ding{55} \\
%         \textsc{PathletRL-Unw} & \ding{51} & \ding{55} & \ding{51} \\
%         \textsc{PathletRL (ours)} & \ding{51} & \ding{51} & \ding{51} \\
%         \bottomrule \\
%     \end{NiceTabular}
%     \caption{Features of the proposed \textsc{PathletRL} models, alongisde with its ablation baselines.}
%     \label{tab:pathletrl-features}
% \end{table}

\subsection{Experimental Parameters}
\label{sec:parameters}

Refer to Appendix \ref{sec:imp} for full details of the implementation. To implement the RL architecture, we used a deep neural network that consists of the following parameters. It comprises of three hidden fully-connected layers of 128, 64 and 32 hidden neurons. The ReLU activation function has been employed, optimized by Adam with a learning rate of 0.001. We also use a 0.2 dropout in the network, together with the Huber loss function. More specific to the DQN's parameters, we have $m=5$ episodes for each of the $n=100$ iterations. The size of the experience replay buffer is 100,000 and the memory minibatch size is 64. Our agent also uses a discount factor $\gamma = 0.99$. Moreover, we use $k=10$ for the $k$-order candidate set, $M$ = 25\% maximum trajectory loss and $\hat{\mu}$ = 80\% average representability threshold. We also set $\alpha_i$ = \textonequarter, $\forall i$, which denotes equal importance for each of the four objectives as depicted in Equation (\ref{eq:reward}). 

\setlength\belowcaptionskip{-10pt}
\begin{table}[t]
    \setlength{\tabcolsep}{2pt}
    \centering
    \begin{NiceTabular}{lccc}
        \toprule
        \Block[c]{}{\textbf{\textsc{PathletRL}} \\ \textbf{\textsc{Algorithm}}} & \Block[c]{}{\textbf{Representability} \\ \textbf{Measure}} & \Block[c]{}{\textbf{Weighted} \\ \textbf{Networks}} & \Block[c]{}{\textbf{Deep Learning} \\ \textbf{Policy}} \\
        \midrule
        \textsc{PathletRL-Nr} & \ding{55} & \ding{51} & \ding{51} \\
        \textsc{PathletRL-Rnd} & \ding{51} & \ding{51} & \ding{55} \\
        \textsc{PathletRL-Unw} & \ding{51} & \ding{55} & \ding{51} \\
        \textsc{PathletRL (ours)} & \ding{51} & \ding{51} & \ding{51} \\
        \bottomrule \\
    \end{NiceTabular}
    \caption{Features of the proposed \textsc{PathletRL} models, alongisde its ablation baselines.}
    \label{tab:pathletrl-features}
\end{table}

\vspace{-3pt}

\subsection{Baselines}
\label{sec:baselines}

% We introduce the following baselines (Appendix \ref{sec:baselines-choice} discusses why the following baselines are selected). The first two are \textsc{SotA} baselines that utilize top-down approaches, while the last three are ablation versions of our proposed model (see Table \ref{tab:pathletrl-features} for a summary of each ablation model and the features withheld in each version to demonstrate the importance and effectiveness of such features).

% Shortened the previous version.
We introduce the following baselines (see Appendix \ref{sec:baselines-choice} for a discussion on baseline choice). The first two are \textsc{SotA} (top-down-based) baselines, while the last three are ablation versions of our proposed model. Table \ref{tab:pathletrl-features} depicts the features withheld in each ablation variation to demonstrate the feature's importance and effectiveness.

\smallskip\noindent \textbf{Chen et al.} \cite{chen2013pathlet}. The very first paper that introduces the notion of pathlets. This method frames the problem as an integer programming formulation, that is solvable using dynamic programming.

\smallskip\noindent \textbf{Agarwal et al.} \cite{agarwal2018subtrajectory}. Framing PD extraction as a subtrajectory clustering problem, where subtrajectory clusters are treated as pathlets, they use \textit{pathlet-cover} inspired from the popular \textit{set-cover} algorithm.

\smallskip\noindent \textbf{\textsc{Sgt}}. The \textit{Singleton} baseline considers all initial length-1 pathlets (the original road map), without merging any pathlets.

\smallskip\noindent \textbf{\textsc{PathletRL-Rnd}}. This version of \textsc{PathletRL} does not support Deep $Q$ Networks and does not utilize a \textsc{Dqn} agent. Actions at each episodic timestep are taken uniformly at random.

% \smallskip\noindent \textbf{\textsc{PathletRL-Nr}}. This \textsc{PathletRL} version does not consider trajectories' representability. If a portion of a trajectory can no longer be reconstructed due to a merge action, then that entire trajectory is discarded right away.
\smallskip\noindent \textbf{\textsc{PathletRL-Nr}}. Trajectory representability is absent under this ablation. If a pathlet traversed by some trajectory $\tau$ merges with another pathlet that is not traversed by this $\tau$, then there no longer exists a subset of pathlets in the pathlet set that can represent $\tau$; as a result, we immediately discard trajectory $\tau$.

% If for instance, some pathlet $\rho$ spanning a part of a trajectory $\tau$ merges with a neighboring pathlet $\rho'$ not spanning $\tau$, then trajectory $\tau$ under this ablation can no longer be represented given the new merged pathlet as it contains $\rho'$ that is not part of $\tau$ and must therefore be discarded immediately.

% If a part of a trajectory can no longer be reconstructed due to a merge action, then the trajectory is discarded immediately.

% \smallskip\noindent \textbf{\textsc{PathletRL-Unw}}. The variation of this \textsc{PathletRL} model is applied on a pathlet graph environment where each pathlet in the graph is weighed of equal importance.
\smallskip\noindent \textbf{\textsc{PathletRL-Unw}}. This version of \textsc{PathletRL} is applied to a pathlet graph environment where all pathlets are equally weighted.

\vspace{-3pt}

\subsection{Evaluation Metrics}
\label{sec:evaluation-metrics}

To evaluate model performance, we consider the following metrics that will measure the quality of the extracted pathlet dictionaries. Note that $[\downarrow]$ ($[\uparrow]$) indicate that lower (higher) values are better.

\begin{enumerate}[leftmargin=*]
    \item $|\mathbb{S}|$, the size of the pathlet dictionary $[\downarrow]$
    \item $\phi$, the avg \# of pathlets that represent each trajectory $[\downarrow]$
    \item $L_{traj}$, the \# of trajectories discarded (expressed in \%) $[\downarrow]$
    \item $\bar{\mu}$, the average representability across the remaining trajectories (expressed in \%) $[\uparrow]$
\end{enumerate}

Note that the third and fourth metrics above do not apply to Chen et al.'s \cite{chen2013pathlet} and Agarwal et al.'s \cite{agarwal2018subtrajectory} methods as such measures are only applicable to pathlet-merging methods. Moreover, the fourth metric does not apply to \textsc{PathletRL-Nr}. Under this model, all remaining trajectories in the dataset (and hence the average) are always 100\% representable, which is not so interesting.

\setlength\belowcaptionskip{0pt}
\begin{table*}[t]
 \setlength{\tabcolsep}{4pt} % Increase the column separation for better spacing
  \centering % Center the table on the page
  \begin{tabular*}{0.8\textwidth}{@{\extracolsep{\fill}} c|lcc|c|cccc @{}}
    \toprule
    \multicolumn{2}{c}{} & \multicolumn{2}{c|}{\textbf{Baselines}} & \textsc{Null} & \multicolumn{4}{c}{\textsc{PathletRL}} \\
    \cmidrule{3-9}
    \multicolumn{2}{c}{} & \cite{chen2013pathlet} & \cite{agarwal2018subtrajectory} & \textsc{Sgt} & \textsc{Rnd} & \textsc{Nr} & \textsc{Unw} & \textsc{(ours)} \\
    \midrule
    \parbox[t]{1mm}{\multirow{4}{*}{\rotatebox[origin=c]{90}{\textsc{\scriptsize{Toronto}}}}} & $[\downarrow] \hspace{1mm} |\mathbb{S}|$ & 13,886 & 7,982 & 2,563 & 2,454 & 1,896 & \underline{1,801} & \textbf{1,743} \\
    & $[\downarrow] \hspace{1mm} \phi$ & 7.02 & 5.97 & 4.76 & 3.77 & \textbf{2.89} & 3.98 & \underline{3.75} \\
    & $[\downarrow] \hspace{1mm} L_{traj}$ & \textsc{n/a} & \textsc{n/a} & 0\% & 19.7\% & 17.6\% & \textbf{15.1\%} & \underline{15.2\%} \\
    & $[\uparrow] \hspace{1mm} \bar{\mu}$ & \textsc{n/a} & \textsc{n/a} & 100\% & 79.9\% & \textsc{n/a} & \underline{80.0\%} & \textbf{80.0\%} \\
    \midrule
    \parbox[t]{1mm}{\multirow{4}{*}{\rotatebox[origin=c]{90}{\textsc{\scriptsize{Rome}}}}} & $[\downarrow] \hspace{1mm} |\mathbb{S}|$ & 59,396 & 31,017 & 15,465 & 9,718 & 7,003 & \underline{5,804} & \textbf{5,291} \\
    & $[\downarrow] \hspace{1mm} \phi$ & 202.91 & 188.33 & 230.15 & 173.04 & 158.18 & \underline{146.39} & \textbf{139.89} \\
    & $[\downarrow] \hspace{1mm} L_{traj}$ & \textsc{n/a} & \textsc{n/a} & 0\% & 24.9\% & \underline{21.1\%} & 22.9\% & \textbf{20.4\%} \\
    & $[\uparrow] \hspace{1mm} \bar{\mu}$ & \textsc{n/a} & \textsc{n/a} & 100\% & 82.7\% & \textsc{n/a} & \textbf{86.2\%} & \underline{85.6\%} \\
    \bottomrule 
\end{tabular*}
  \caption{Numerical results showing the attributes of the pathlet dictionaries extracted by each method for each dataset.}
\label{tab:pd-results}
\end{table*}

\subsection{Results and Discussion}
\label{sec:results-discussion}

In this section, we go over each of the six research questions, present the experimental results and provide some discussion.
% Visualizations of the pathlets in the extracted dictionaries can be found in Appendix \ref{sec:visualizations}.

% \subsubsection{Quality of the Extracted OPD against \textsc{SotA} Methods}
% \label{sec:quality-opd-sota}

\smallskip\noindent \textbf{(Q1) Quality of the Extracted PDs}. We evaluate the pathlet dictionary extracted by our \textsc{PathletRL} algorithm against the PDs extracted by \textsc{SotA} baselines. See Table \ref{tab:pd-results} for the numerical results, where the bold numbers indicate the result of the most superior model for the given PD metric and the underlined number is the result of the second-best performing model (note that we do not boldface or underline the numbers of \textsc{Sgt} as such model serves as the null model where nothing is done to the pathlet graph). Although the nature of the pathlet definition and the approaches are not necessarily the same, the algorithms of Chen et al. \cite{chen2013pathlet} and Agarwal et al. \cite{agarwal2018subtrajectory} are still comparable. We ultimately show that their top-down approaches are not as effective as our bottom-up strategies. First, we look at the size of the pathlet dictionary, $|\mathbb{S}|$, where the smaller the number the better is the result. Our \textsc{PathletRL} model was able to improve from \textsc{Sgt} by $\sim$32.0\% ($\sim$65.8\%) for the \textsc{Toronto} (\textsc{Rome}) dataset. These numbers are an $\sim$87.4\% ($\sim$91.1\%) improvement from Chen et al.'s \cite{chen2013pathlet} model on the \textsc{Toronto} (\textsc{Rome}) dataset. Our model also improves by $\sim$78.2\% ($\sim$82.9\%) from Agarwal et al.'s \cite{agarwal2018subtrajectory} method on the \textsc{Toronto} (\textsc{Rome}) dataset. These two observations indicate how superior our method is against the state-of-the-art; i.e., bottom-up methods being better than top-down schemes. Note as well that Chen et al.'s \cite{chen2013pathlet} and Agarwal et al.'s \cite{agarwal2018subtrajectory} PDs are larger than the initial \# of length-1 pathlets, as their methods are top-down -- which initially considers all possible pathlet sizes and configurations (including overlaps). Clearly, they do not exhibit ideal results, compared to our proposed \textsc{PathletRL} model, as well as in all of the ablation versions of \textsc{PathletRL}.

Then, we focus on the metric of the average pathlet number that represents each trajectory (which can go up or down at each step of the iterative algorithm). Similar to $|\mathbb{S}|$, a smaller $\phi$ indicates a more ideal dictionary. Our \textsc{PathletRL} model was able to extract dictionaries that improve from \textsc{Sgt} by $\sim$21.2\% ($\sim$39.2\%) on the \textsc{Toronto} (\textsc{Rome}) dataset. Meanwhile, Chen et al.'s \cite{chen2013pathlet} PD has $\phi$ quality $\sim$47.4\% higher than \textsc{Sgt} for \textsc{Toronto} dataset, and only $\sim$11.8\% lower than \textsc{Sgt} on \textsc{Rome} dataset. A similar trend can be seen in Agarwal et al.'s \cite{agarwal2018subtrajectory} dictionary, with $\sim$25.4\% higher and $\sim$18.2\% lower than the initial number in the \textsc{Toronto} and \textsc{Rome} datasets respectively. Clearly, our proposed \textsc{PathletRL} (and its ablation variations) outperforms these \textsc{SotA} baselines. 
% Comparing these baselines with our proposed \textsc{PathletRL}, one can clearly see how much superior ours is; with the ablation variatons's PDs with higher quality than the baselines on the PD $\phi$ metric.

Our \textsc{PathletRL} also improves from \textsc{Sgt} based on the $|\mathbb{S}|$ and $\phi$ metrics. Because no action is taken on the pathlet graph in \textsc{Sgt}, only the original numbers are shown; thus, $L_{traj}$ and $\bar{\mu}$ remains as 0\% and 100\% for both datasets. However, we can see the benefits of \textsc{PathletRL} by trading off these values to obtain smaller dictionaries with much less $\phi$ scores -- as controlled by the $\alpha$ parameters.

\smallskip\noindent \textbf{(Q2) Memory Efficiency}. Fig. \ref{fig:memory-plot} (in log scale) demonstrates how much more memory-efficient \textsc{PathletRL} is compared to the baselines \cite{chen2013pathlet, agarwal2018subtrajectory}. 
As can be seen in Fig. \ref{fig:memory-plot}, our method gets as input a set of trajectories that requires $\sim$900 MB ($\sim$30+ GB) to be stored in memory, and builds a trajectory pathlet dictionary that requires a mere $\sim$100 KB ($\sim$1 MB) for the \textsc{Toronto} (\textsc{Rome}) dataset. In fact, this represents a $\sim$7.4K$\times$ ($\sim$24K$\times$) saving. 
%
%Clearly, our method is more superior as it can be seen that the $\sim$900 MB ($\sim$30+ GB) memory storage requirement that is necessary to store the initial pathlets of the dictionary by the baselines can significantly be reduced by our model to a measly $\sim$100 KB ($\sim$1 MB) for the \textsc{Toronto} (\textsc{Rome}) dataset. Indeed, this depicts large amount of memory savings -- a $\sim$7.4K$\times$ ($\sim$24K$\times$) improvement. 
%
This considerable improvement can be attributed to the fact that our method uses a bottom-up approach where only edge-disjoint pathlets are considered. In contrast, the current baselines follow a top-down approach, where the trajectory pathlet dictionary consists of overlapping pathlets of various sizes and configurations, most of which are redundant.

\setlength\belowcaptionskip{0pt}
\begin{figure}[t]
    \centering
    \includegraphics[width=0.7\textwidth]{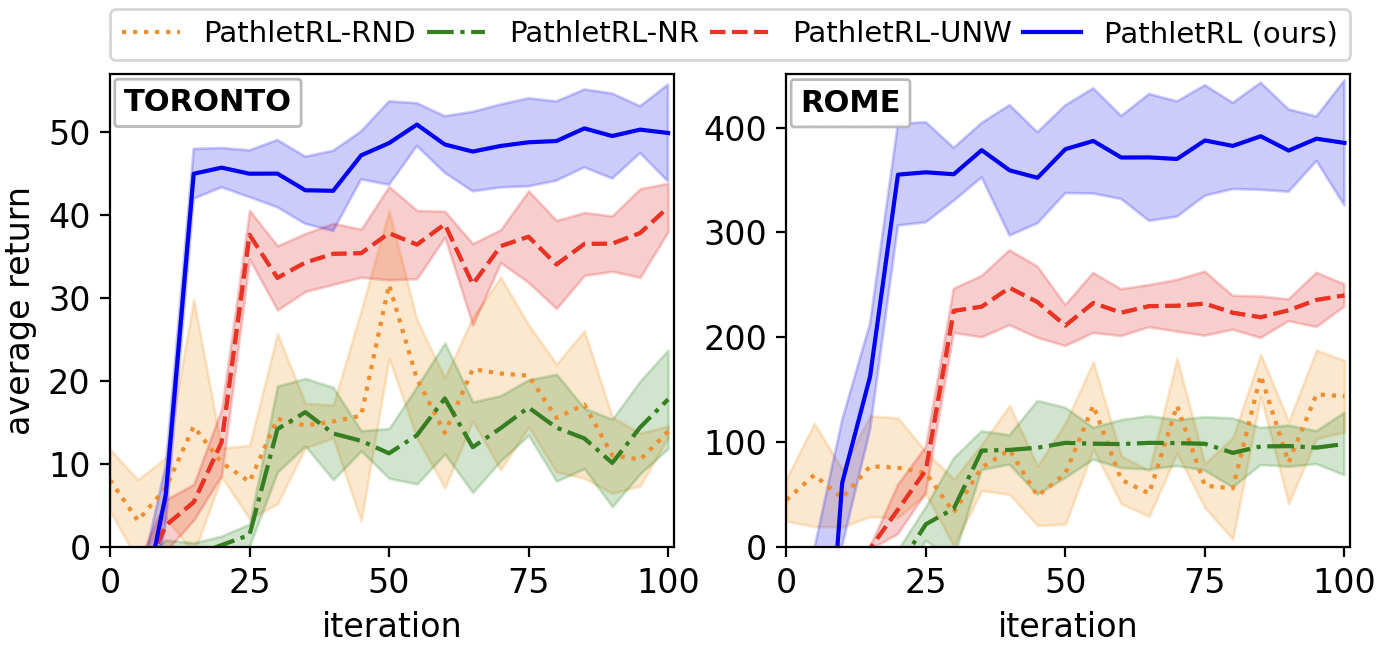}
    \caption{Performance evaluation of proposed and ablation \textsc{PathletRL} models, measured using the average return metric of $m = 5$ episodes across $n = 100$ iterations (run 10 times)}
    \label{fig:drl-performance}
\end{figure}

\smallskip\noindent \textbf{(Q3) Ablation Study}. Next, we perform an ablation experiment to see how well our proposed \textsc{PathletRL} model performs. Fig. \ref{fig:drl-performance} displays the average returns of \textsc{PathletRL} and its ablations across $n$ iterations on the two datasets. We can observe similar trends on both datasets. Notice that \textsc{PathletRL-Rnd} has the poorest performance, exhibiting a random RL policy that shows no learning at all. Meanwhile, all other models demonstrate that their average return value converges after some iterations (for example, 15 and 20 iterations for \textsc{PathletRL} on the \textsc{Toronto} and \textsc{Rome} datasets), and then fluctuating slightly within a small range. \textsc{PathletRL-Nr}, while it demonstrates some level of learning due to the \textsc{Dqn} policy, does not perform well compared to \textsc{PathletRL-Unw} -- which suggests that representability is an essential component. This unweighted version of \textsc{PathletRL} can be seen as a runner up to our (weighted) proposed model, which  indicates that there is some value to assigning pathlet weights than simply weighing all pathlets equally.
% mean that assigning pathlet weights are more important than just simply weighing all pathlets equally.

% In addition to comparing the trends of \textsc{PathletRL} models' performance evaluation, we can also look at the quality of their PDs (see Table \ref{tab:pd-results} for the numerical results). 
Besides comparing the trends of \textsc{PathletRL} models' performance evaluation, we can also look at the quality of their PDs (see Table \ref{tab:pd-results} for the results). Generally speaking, our proposed \textsc{PathletRL}'s PD is superior than the PDs extracted by its ablation variations (i.e., the $|\mathbb{S}|$ metrics for both the \textsc{Toronto} and \textsc{Rome} datasets, the $\bar{\mu}$ metric for the \textsc{Toronto} dataset, and the $\phi$ and $L_{traj}$ metrics for the \textsc{Rome} dataset). In other cases that \textsc{PathletRL} did not rank first, it was a runner up but this can be explained. For instance, consider the $\phi$ metric in the \textsc{Toronto} dataset. The reason for the higher quality of \textsc{PathletRL-Nr}'s PD than that of  \textsc{PathletRL}'s in terms of the $\phi$ metric is that because trajectory counts easily shrink faster in \textsc{PathletRL-Nr}; the average $\phi$ can easily go down should the number of pathlets representing each trajectory also dwindle in number. The $\bar{\mu}$ metric for the PD of \textsc{PathletRL-Unw} is higher than that of the PD of \textsc{PathletRL} on the \textsc{Rome} dataset, which could be because \textsc{PathletRL-Unw} has fewer trajectories remaining in its trajectory set and that it just so happened that those that remained have high representability $\mu$ values. Regardless, the differences in numbers between the PDs of \textsc{PathletRL} and \textsc{PathletRL-Unw} in terms of the $\bar{\mu}$ metric is small and still comparable. The same can be said for the $L_{traj}$ metric of the PDs of \textsc{PathletRL-Unw} and \textsc{PathletRL} on the \textsc{Toronto} dataset, which differs by only a measly $\sim$0.1\% -- demonstrating that \textsc{PathletRL} is still competitive.

\setlength\belowcaptionskip{0pt}
\begin{figure}[t]
    \centering
    \includegraphics[width=0.7\textwidth]{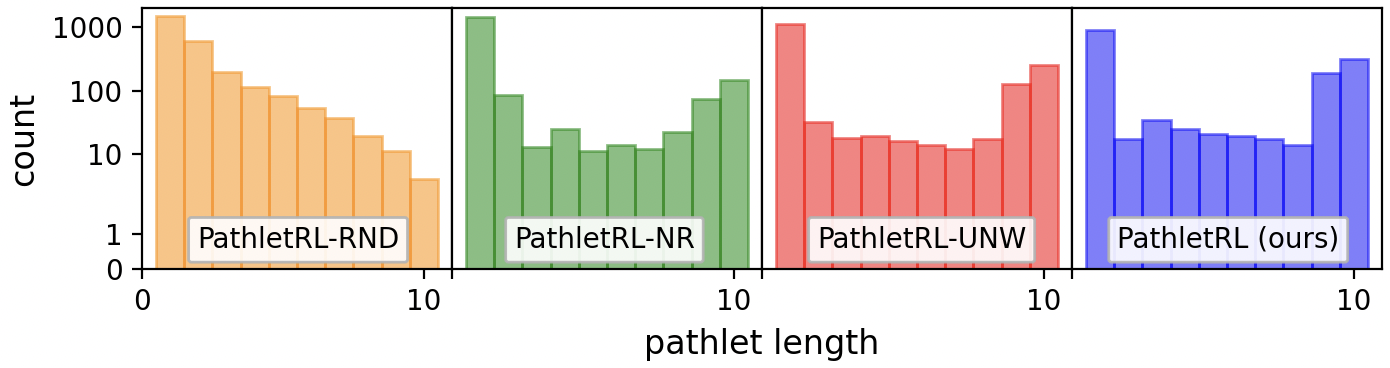}
    \includegraphics[width=0.7\textwidth]{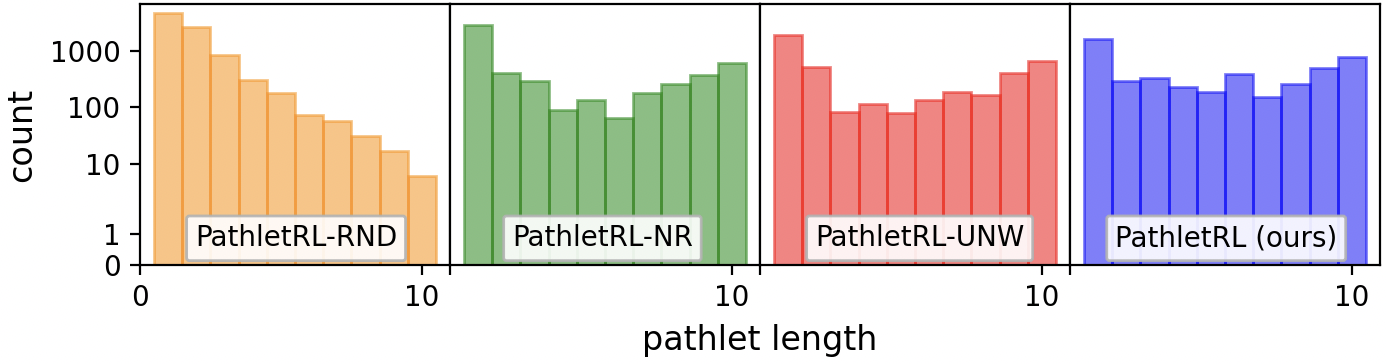}
    \caption{The pathlet length distribution of pathlet dictionaries obtained by \textsc{PathletRL} model and its ablation versions on the \textsc{Toronto} (top) and \textsc{Rome} (bottom) datasets.}
    \label{fig:distribution}
\end{figure}

\smallskip\noindent \textbf{(Q4) Pathlet Length Distribution}. We also analyze the length distribution of the pathlets in our dictionaries. The trend is similar for both datasets (Fig. \ref{fig:distribution} shows the pathlet length distribution, with the $y$-axis in log scale). The \textsc{PathletRL-Rnd} has a decreasing \# for longer pathlets, which is intuitive as the random policy blindly keeps \& merges pathlets. It is harder to maintain longer pathlets in this random probabilistic manner as pathlets that terminate their growth are already considered ``processed'' and cannot grow further. As a result, it is more rare to see higher-ordered pathlets than shorter pathlets in \textsc{PathletRL-Rnd}'s PD. The rest of the other RL models utilizing \textsc{Dqn} policy have longer-length pathlets as expected (with more length-9 and 10 pathlets in the dictionary). Our proposed PD can capture more of these higher-ordered pathlets -- indicating a smaller pathlet dictionary, as reflected by the results in Table \ref{tab:pd-results}. Meanwhile, we can still observe a large number of length-1 pathlets, which may be due to a number of reasons. One could be that some of the length-1 pathlets are still \textit{unprocessed} (as a result of early stopping caused by the various termination criteria in our algorithm). It could also be that some of the length-1 processed pathlets are unmerged due to the algorithm's recommendation based on the utility, or perhaps based on pathlets losing all neighbor pathlets to merge with. The latter case depicts a scenario for where a length-1 processed pathlet $\rho$ may have lost all its neighbor pathlets as a result of these neighbors merging amongst one another, leaving no way for $\rho$ to merge with any of these formed merged pathlets.

\begin{figure}[t]
    \centering
    \includegraphics[width=0.7\textwidth]{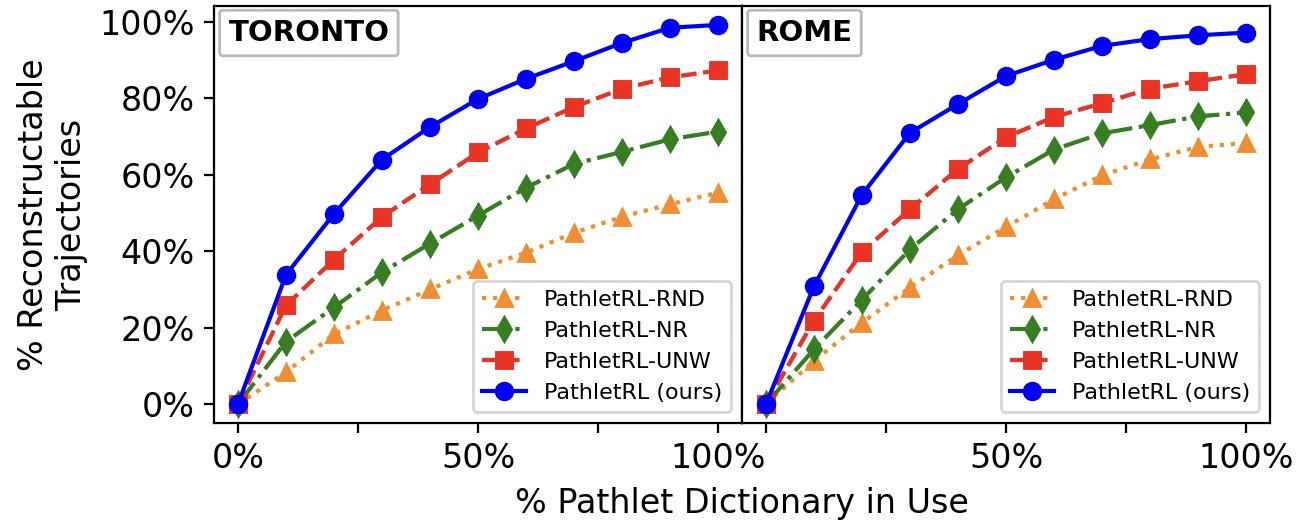}
    \caption{The \% of evaluation trajectories reconstructable from a sample set taken from extracted pathlet dictionaries.}
    \label{fig:partial-traj}
\end{figure}

\smallskip\noindent \textbf{(Q5) Partial Trajectory Reconstruction}. See Fig. \ref{fig:partial-traj} for a plot that displays the results of this experiment, where we determine how much of the dictionary is adequate enough to reconstruct most of our trajectories in our testing set. Here, we say that a trajectory is \textit{reconstructable} if its representability value $\mu \geq 0.75$ (i.e., 75\% of the trajectory can be represented by the PD). Anything less would mean that the trajectory is not reconstructable due to an excessive number of gaps. Ideally, we would like to take the top $x$\% of the pathlets in the PD that are the most traversed by the trajectories in the training set. However, we can further remove the bias in the experiment by choosing instead a random sample of $x \in [10\%,20\%,...,100\%]$ pathlets in the extracted PD, and measuring how much of the trajectories in the testing set are reconstructable by this pathlet sample set. As shown in the results, by using around half of the pathlets in \textsc{PathletRL}'s PD, a good $\sim$80\% ($\sim$85\%) of the trajectories in the \textsc{Toronto} (\textsc{Rome}) testing set can be reconstructed; and by using the whole dictionary, almost all trajectories in the set can be reconstructed. This shows that our proposed model is able to extract a high-quality pathlet dictionary. Comparing this with our ablation versions, they all follow a similar trend for both datasets -- i.e., the amount of trajectories that the dictionaries of the ablation models can reconstruct is less than the amount that our proposed \textsc{PathletRL}'s dictionary can do so. The worst being is \textsc{PathletRL-Rnd}'s dictionary that can reconstruct up to only $\sim$50\% ($\sim$65\%) of the trajectories in \textsc{Toronto} (\textsc{Rome}) given the entire dictionary.

\smallskip\noindent \textbf{(Q6) Parameter Sensitivity Analysis}. We then discuss how the values of the four $\alpha$ values ($\alpha_1$, $\alpha_2$, $\alpha_3$, $\alpha_4$) affect the output PD's quality in terms of its four attributes: $|\mathbb{S}|$, $\phi$, $L_{traj}$, $\bar{\mu}$ respectively\footnote{\textbf{Note}: the goal here is not to find an optimal value for $\alpha_i$, but intends on showing the trend of how $|\mathbb{S}|$, $\phi$, $L_{traj}$, $\bar{\mu}$ changes with varying values of $\alpha_1$, $\alpha_2$, $\alpha_3$, $\alpha_4$ respectively.}. There are four stacked plots, as seen in Fig. \ref{fig:param-sens}. The first one depicts the changes to the PD's $|\mathbb{S}|$ as we vary $\alpha_1 = [0.0, 0.1, ..., 1.0]$ (while keeping the other $\alpha$ terms equal $\alpha_2 = \alpha_3 = \alpha_4 = \frac{1-\alpha_1}{3}$). We notice, as expected, a decreasing trend; larger $\alpha_1$ values indicate higher importance to a smaller dictionary size. By varying $\alpha_2$ and keeping the other terms equal, we can also observe a decreasing trend; higher $\alpha_2$ values imply a stronger emphasis on lower $\phi$ scores -- that is, lower average number of pathlets representing trajectories. Then when term $\alpha_3$'s value is varied, while keeping other $\alpha$ terms the same, it also shows a trend that decreases as we increases $\alpha_3$. The greater the $\alpha_3$ is, the more importance we put into keeping the trajectory loss low. Finally, varying $\alpha_4$ while keeping the other $\alpha$ terms equal, shows an opposite trend than the other $\alpha$ terms. Here, we can observe that larger $\alpha_4$ values are indicative of greater importance towards possessing higher representability values.
\vspace{-3pt}

\begin{figure}[t]
    \centering
    \setlength\belowcaptionskip{2pt}
    \includegraphics[width=0.6\textwidth]{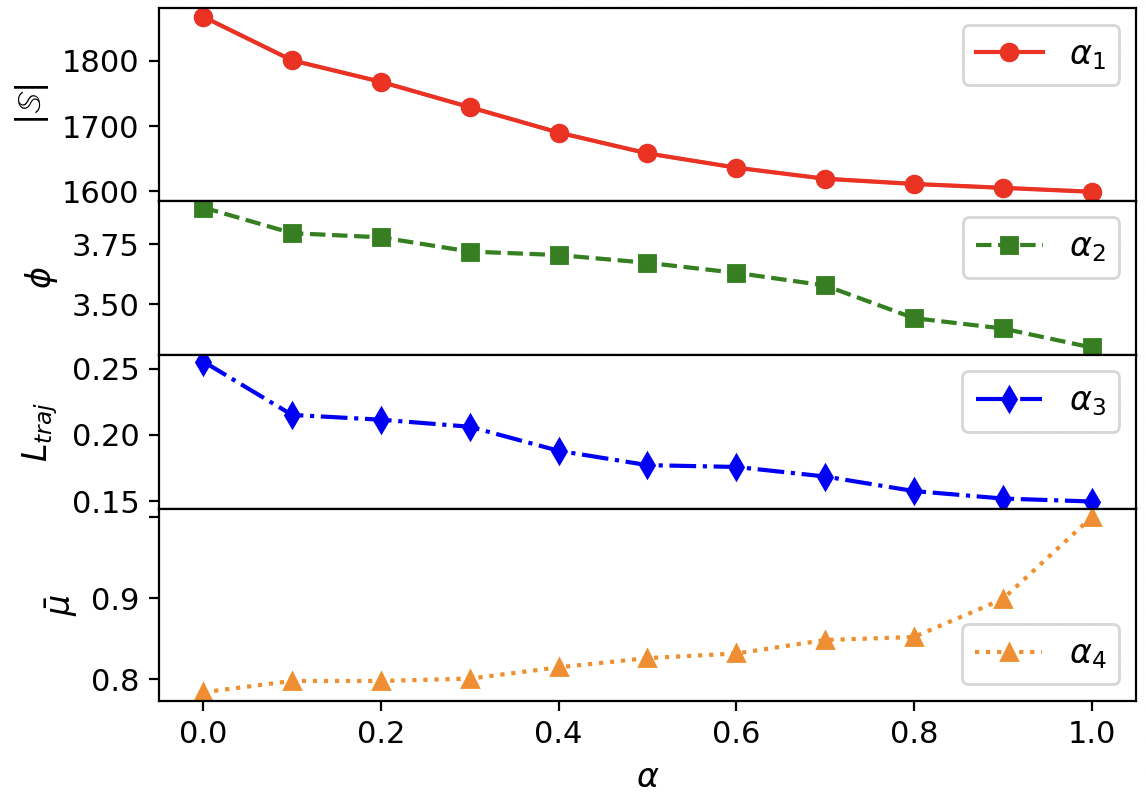}
    \caption{Parameter sensitivity experiment: the impact of $\alpha$'s on the quality of \textsc{PathletRL}'s pathlet dictionary}
    \label{fig:param-sens}
\end{figure}

\begin{figure}[h!]
    \centering
    \begin{subfigure}[b]{0.49\textwidth}
        \includegraphics[width=\textwidth]{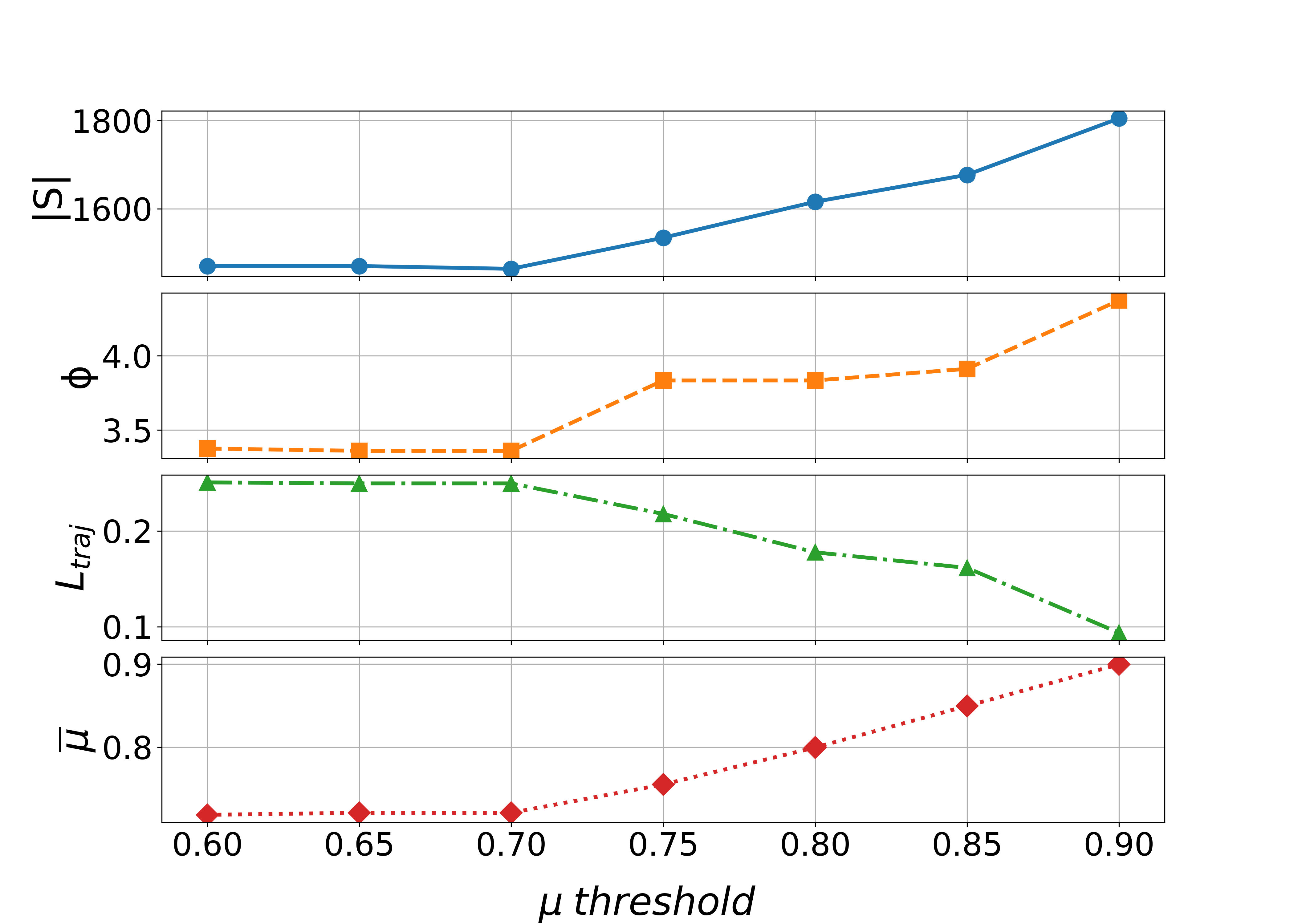}
        \caption{State variables vs.\ $\mu_{\text{threshold}}$ for $k = 10$.}
        \label{fig:plot1}
    \end{subfigure}
    \hfill
    \begin{subfigure}[b]{0.49\textwidth}
        \includegraphics[width=\textwidth]{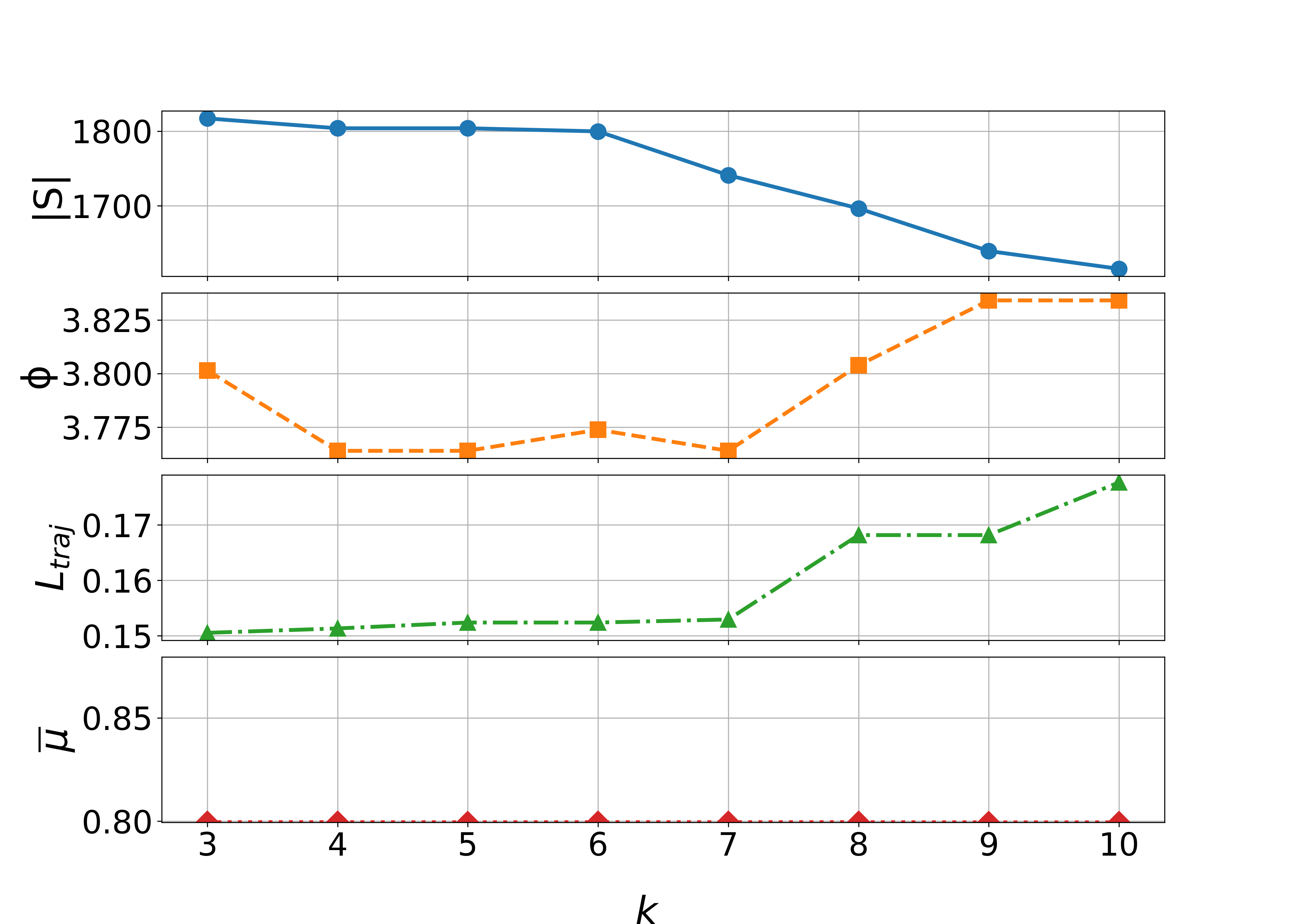}
        \caption{State variables vs.\ $k$ for $\mu_{\text{threshold}} = 0.8$.}
        \label{fig:plot2}
    \end{subfigure}
    \caption{Comparison of state variables under varying $\mu_{\text{threshold}}$ and $k$ hyper-parameters.}
    \label{fig:side_by_side_plots}
\end{figure}

\smallskip\noindent \textbf{(Q7) Hyper-Parameter Impact on Model Performance}. We analyze the impact of the hyper-parameters $\mu_{\text{threshold}}$ and $k$ on the attributes $|\mathbb{S}|$, $\phi$, $L_{\text{traj}}$, and $\bar{\mu}$. When varying $\mu_{\text{threshold}}$ with a fixed $k = 10$, we observe several trends. The size of the set ($|\mathbb{S}|$) remains stable at first but grows substantially as $\mu_{\text{threshold}}$ increases, indicating that, as anticipated, higher values of $\mu_{\text{threshold}}$ lead to larger pathlet counts. This is because the episodes become shorter as we the agent hits the threshold more quickly. The average number of pathlets per trajectory ($\phi$) remains relatively constant at lower thresholds but increases when $\mu_{\text{threshold}}$ is higher, indicating that more pathlets are required to represent each trajectory since fewer merges occur as the threshold is reached sooner. The relationship between the Trajectory Loss ($L_{\text{traj}}$) and the Average Representability ($\bar{\mu}$) exhibits opposing trends as the values of $\mu_{\text{threshold}}$ increase. This is because when $\mu_{\text{threshold}}$ is set below a certain value (70\% in this case), the trajectory loss threshold (set at 25\%) becomes the determining factor for termination, as it is reached before the lower $\mu_{\text{threshold}}$. However, when $\mu_{\text{threshold}}$ exceeds 70\%, $\bar{\mu}$ matches the $\mu_{\text{threshold}}$ because it now becomes the primary terminating criterion. As a result, the trajectory loss no longer dictates termination and begins to decrease, since fewer pathlets are merged, leading to a lower number of lost trajectories.

When fixing $\mu_{\text{threshold}} = 0.8$ and varying $k$, a distinct set of trends emerges. The size of the set ($|\mathbb{S}|$) decreases as $k$ increases, suggesting that using longer pathlets results in a more compact PD, as anticipated. The average number of pathlets per trajectory ($\phi$) shows a slight increase with higher $k$, indicating that trajectories are broken into more segments as $k$ rises. The trajectory loss ($L_{\text{traj}}$) increases as expected, reflecting a minor reduction in reconstruction quality as more pathlets are merged together. Additionally, the $\mu_{\text{threshold}}$ remains fixed at 80\%, consistent with the initial parameter setting mentioned earlier.

We analyze the distribution of pathlet lengths generated for various values of $k$ and a fixed $\mu_{\text{threshold}}$. It is evident that increasing the maximum pathlet length ($k$) encourages the agent to merge more pathlets, thereby reducing the number of pathlets with an initial length of 1. Furthermore, the plots reveal that although pathlets of varying lengths are present in each setting, the agent predominantly generates pathlets that are close to the maximum length $k$, rather than distributing them evenly across lengths from 1 to $k$. When examining the effect of different $\mu_{\text{threshold}}$ values, we see that lower $\mu_{\text{threshold}}$ values allow for greater merging, resulting in a higher proportion of merged pathlets, as expected.

\begin{figure}[t!]
    \centering
    % First row: mu_threshold = 80%
    \begin{center}  % Use center instead of subfigure to avoid row labeling
        \begin{subfigure}{0.24\textwidth}
            \includegraphics[width=\textwidth]{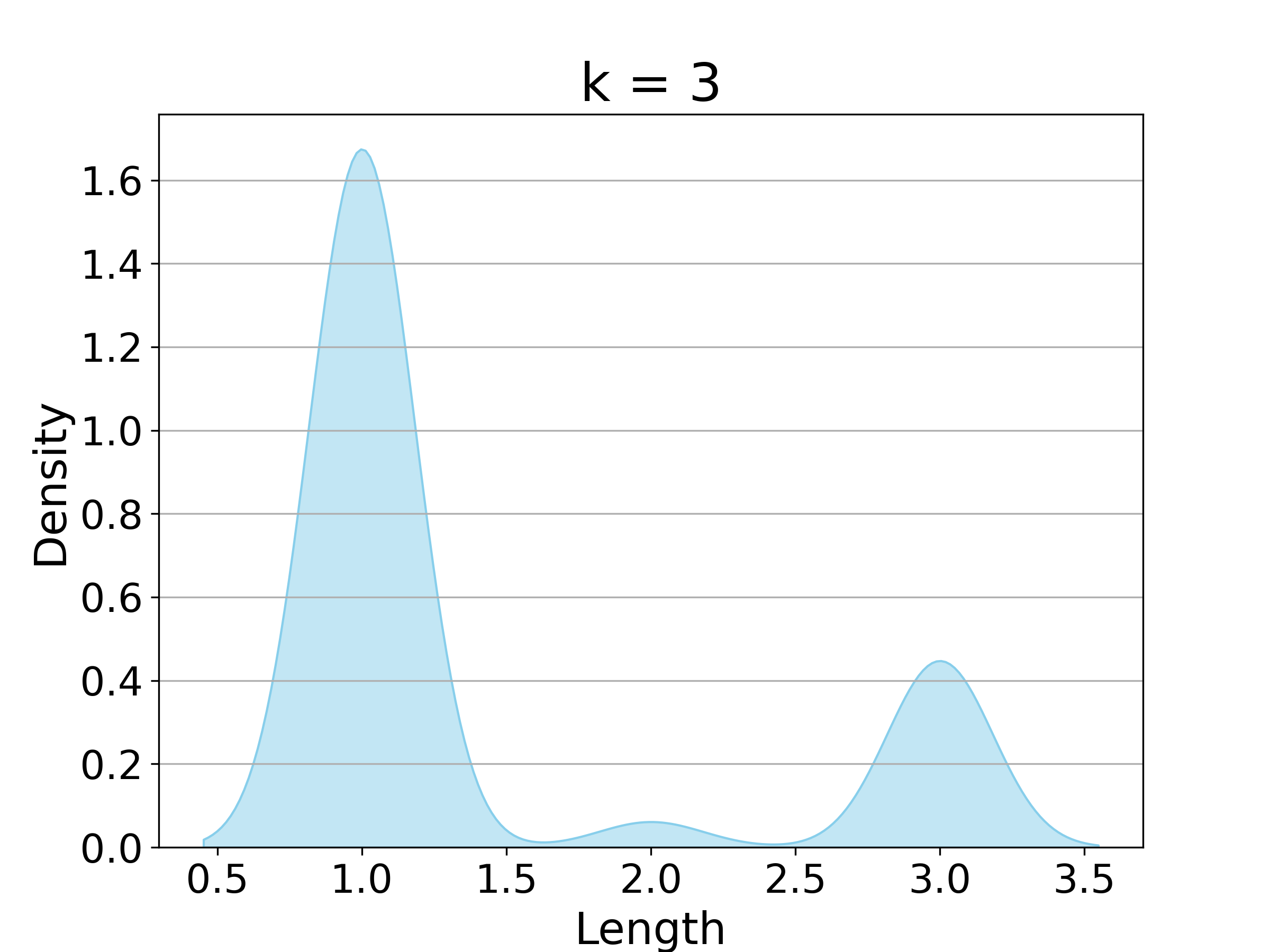}
            \caption{}
            \label{fig:plot1}
        \end{subfigure}
        \begin{subfigure}{0.24\textwidth}
            \includegraphics[width=\textwidth]{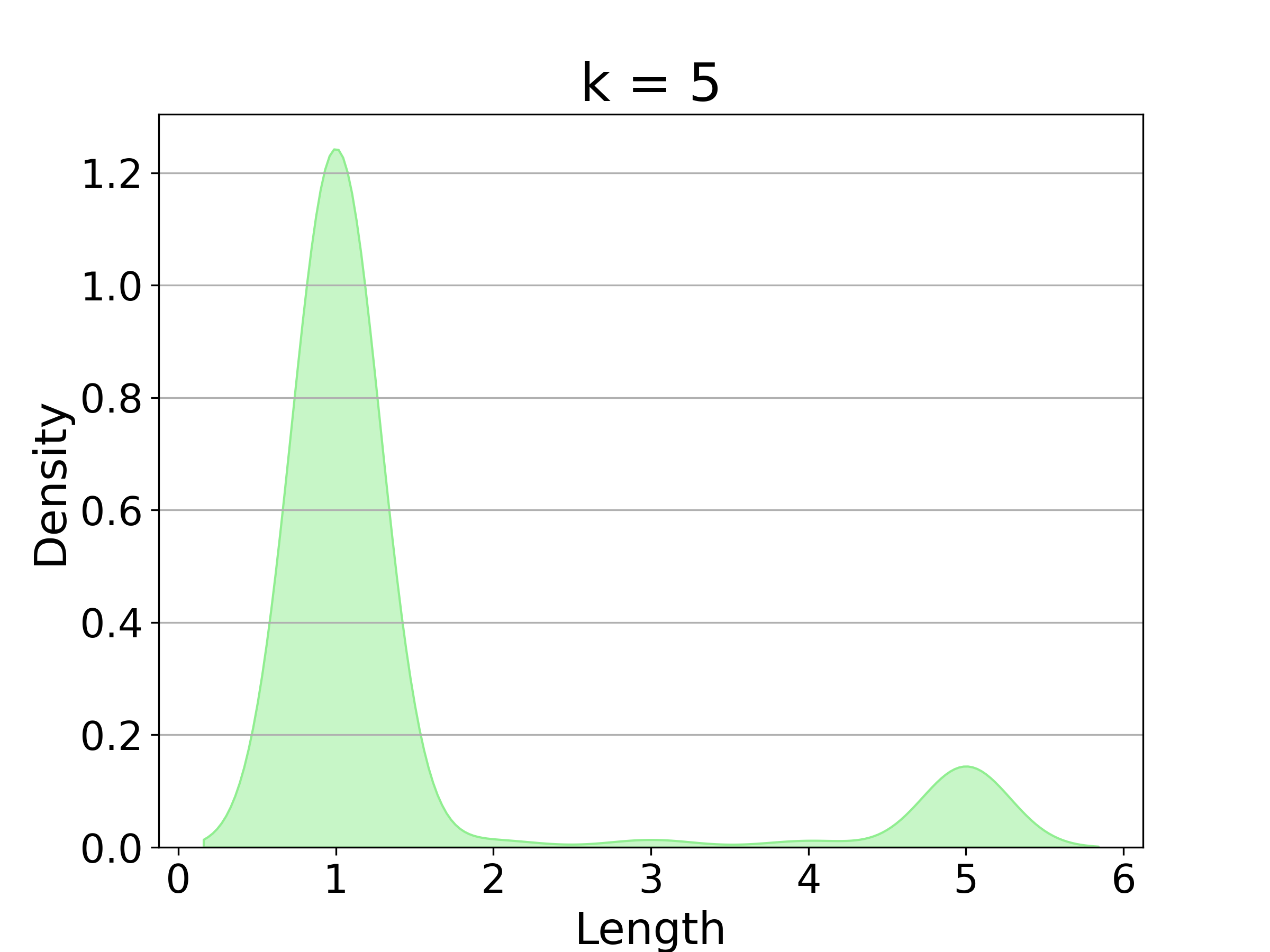}
            \caption{}
            \label{fig:plot2}
        \end{subfigure}
        \begin{subfigure}{0.24\textwidth}
            \includegraphics[width=\textwidth]{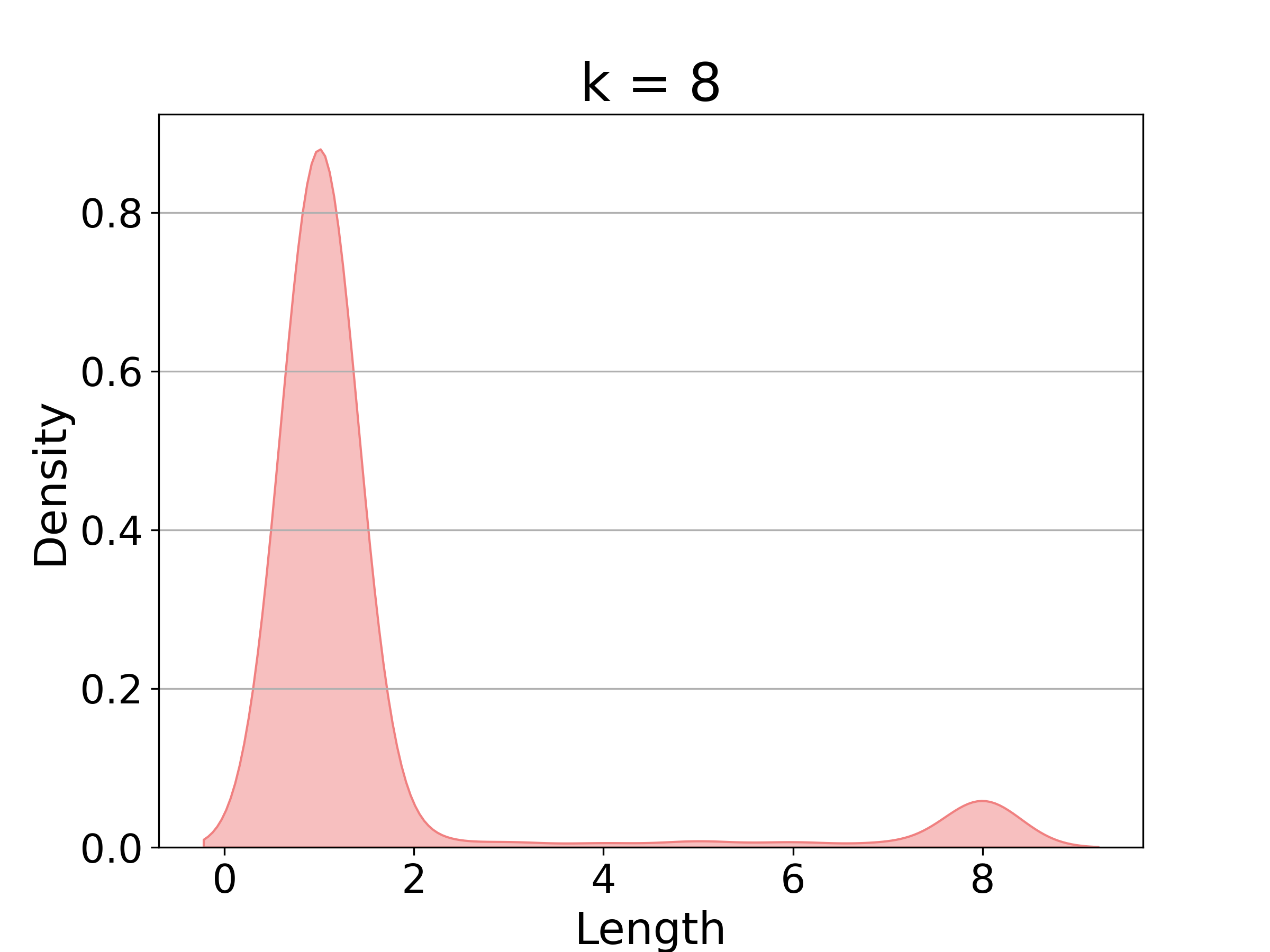}
            \caption{}
            \label{fig:plot3}
        \end{subfigure}
        \begin{subfigure}{0.24\textwidth}
            \includegraphics[width=\textwidth]{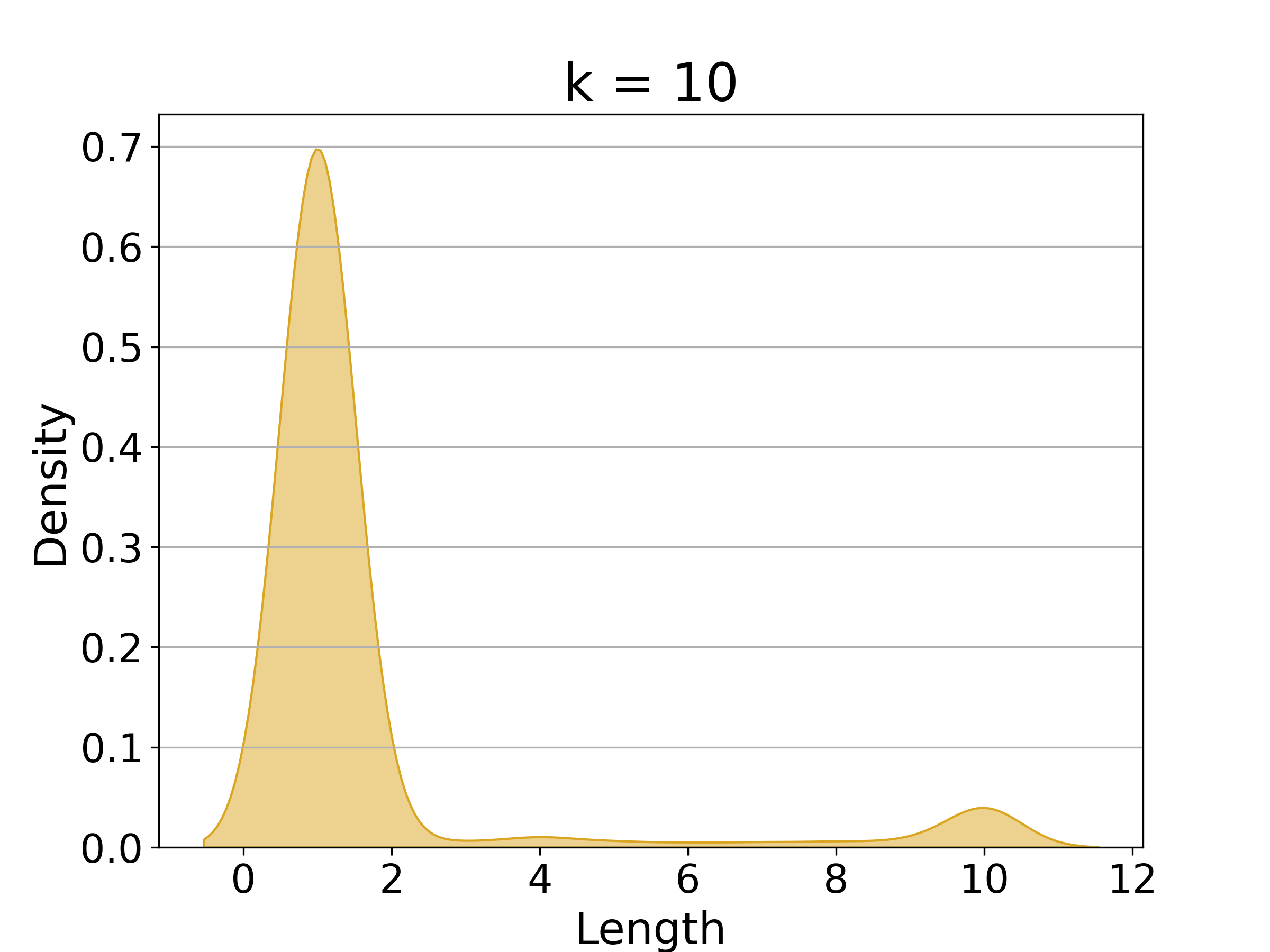}
            \caption{}
            \label{fig:plot4}
        \end{subfigure}
        \caption*{$\mu_{\text{threshold}} = 80\%$.}  % Unnumbered caption
    \end{center}
    
    % Second row: mu_threshold = 60%
    \begin{center}  % Use center instead of subfigure to avoid row labeling
        \begin{subfigure}{0.24\textwidth}
            \includegraphics[width=\textwidth]{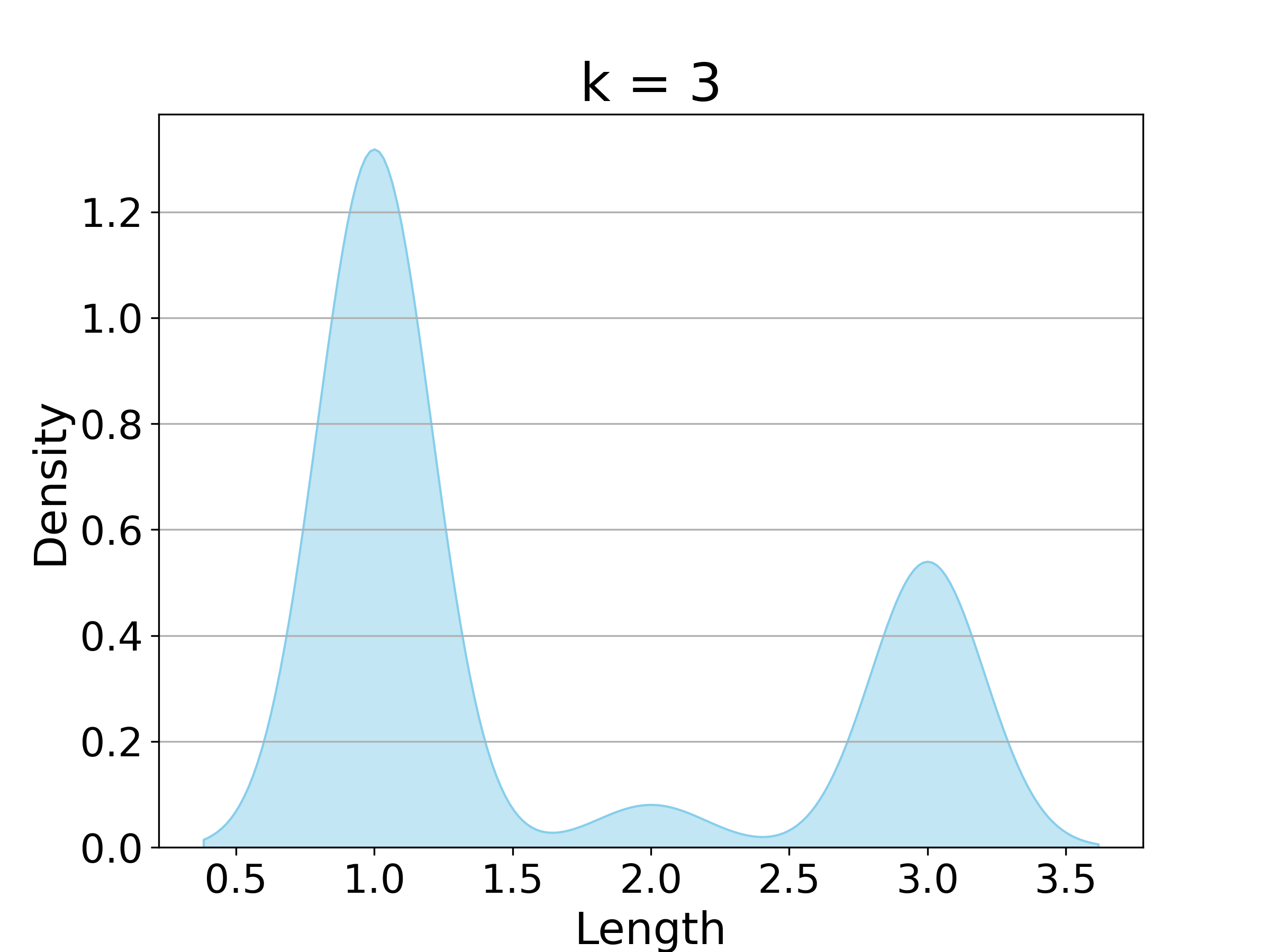}
            \caption{}
            \label{fig:plot5}
        \end{subfigure}
        \begin{subfigure}{0.24\textwidth}
            \includegraphics[width=\textwidth]{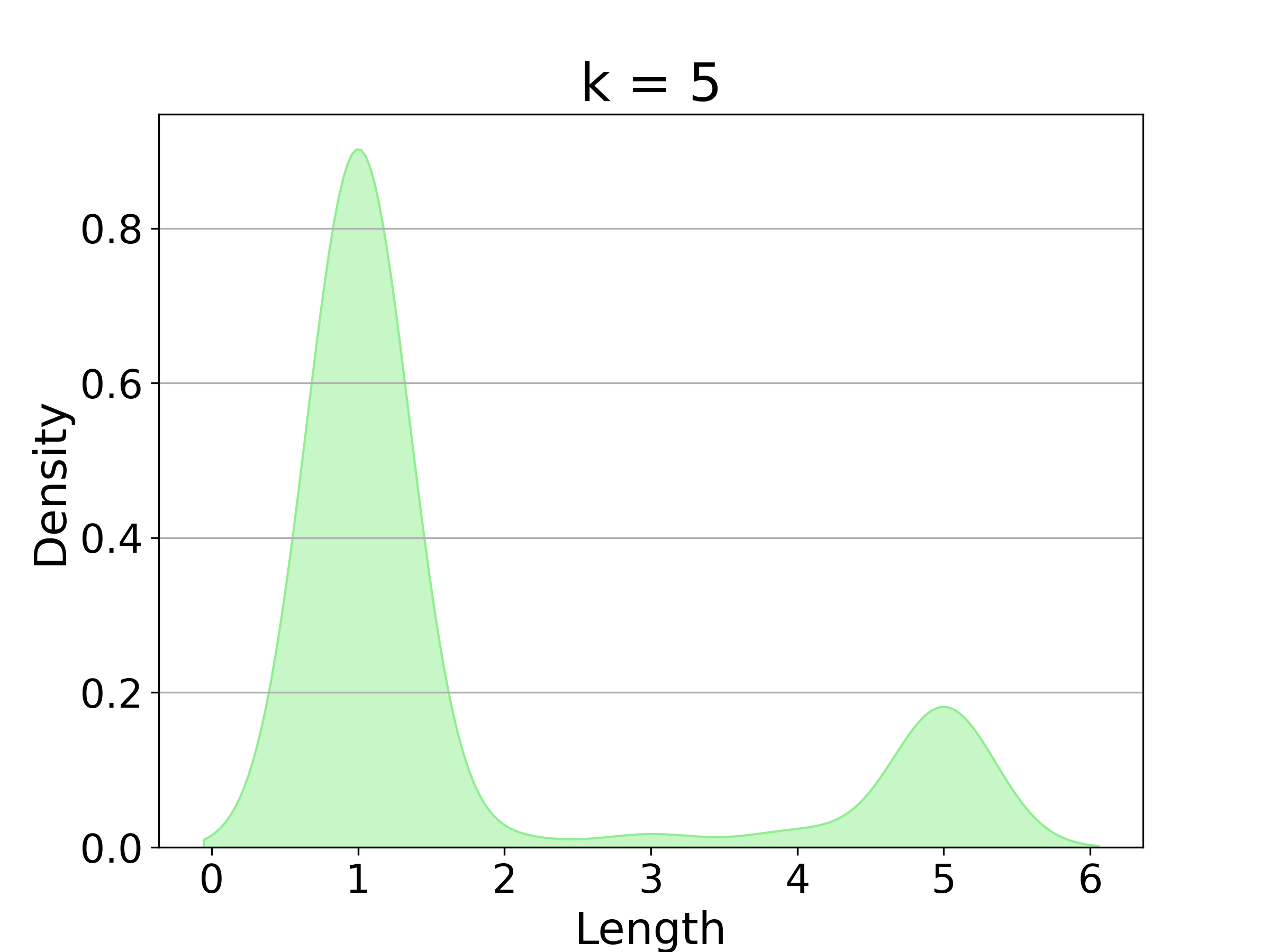}
            \caption{}
            \label{fig:plot6}
        \end{subfigure}
        \begin{subfigure}{0.24\textwidth}
            \includegraphics[width=\textwidth]{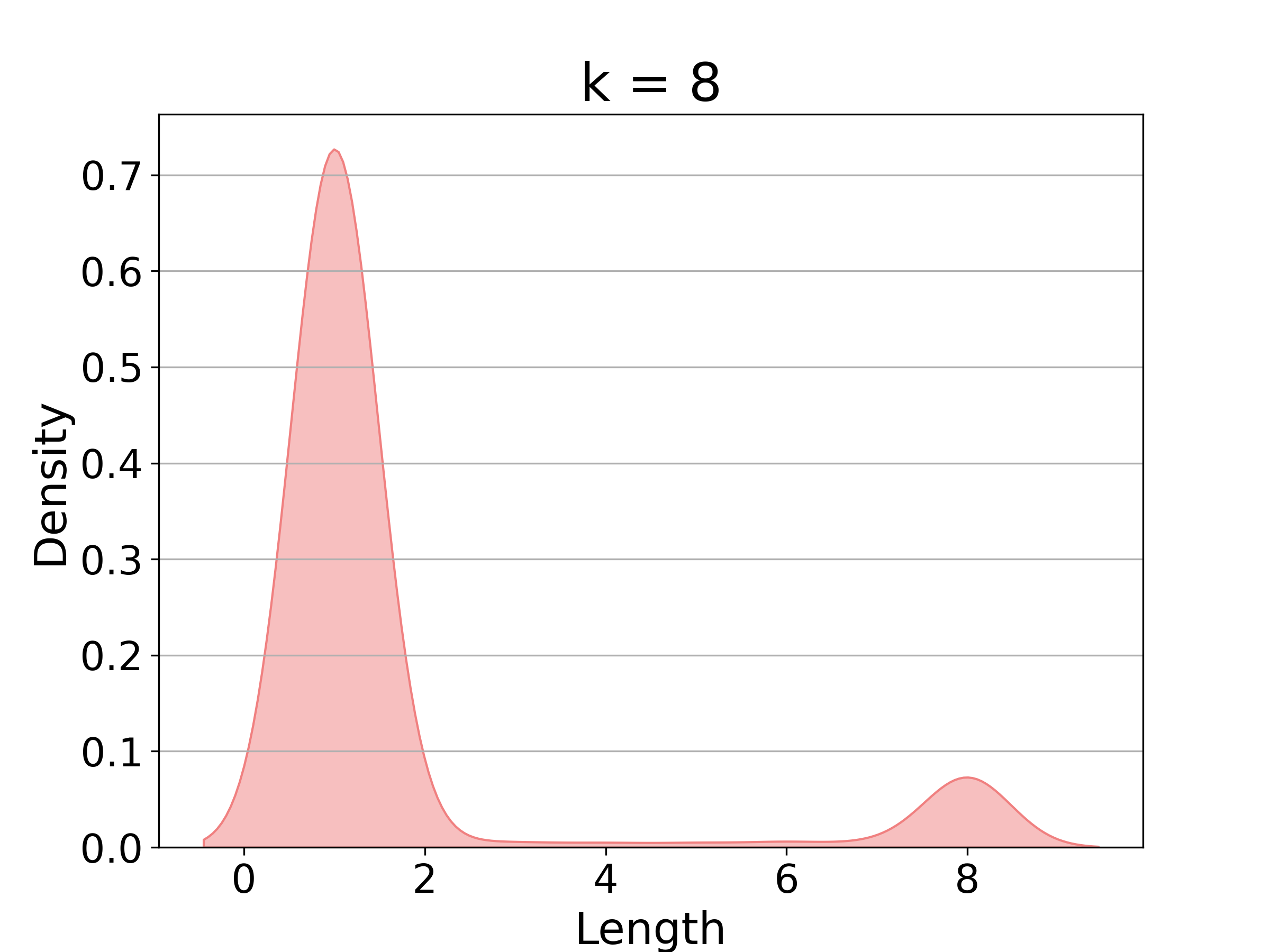}
            \caption{}
            \label{fig:plot7}
        \end{subfigure}
        \begin{subfigure}{0.24\textwidth}
            \includegraphics[width=\textwidth]{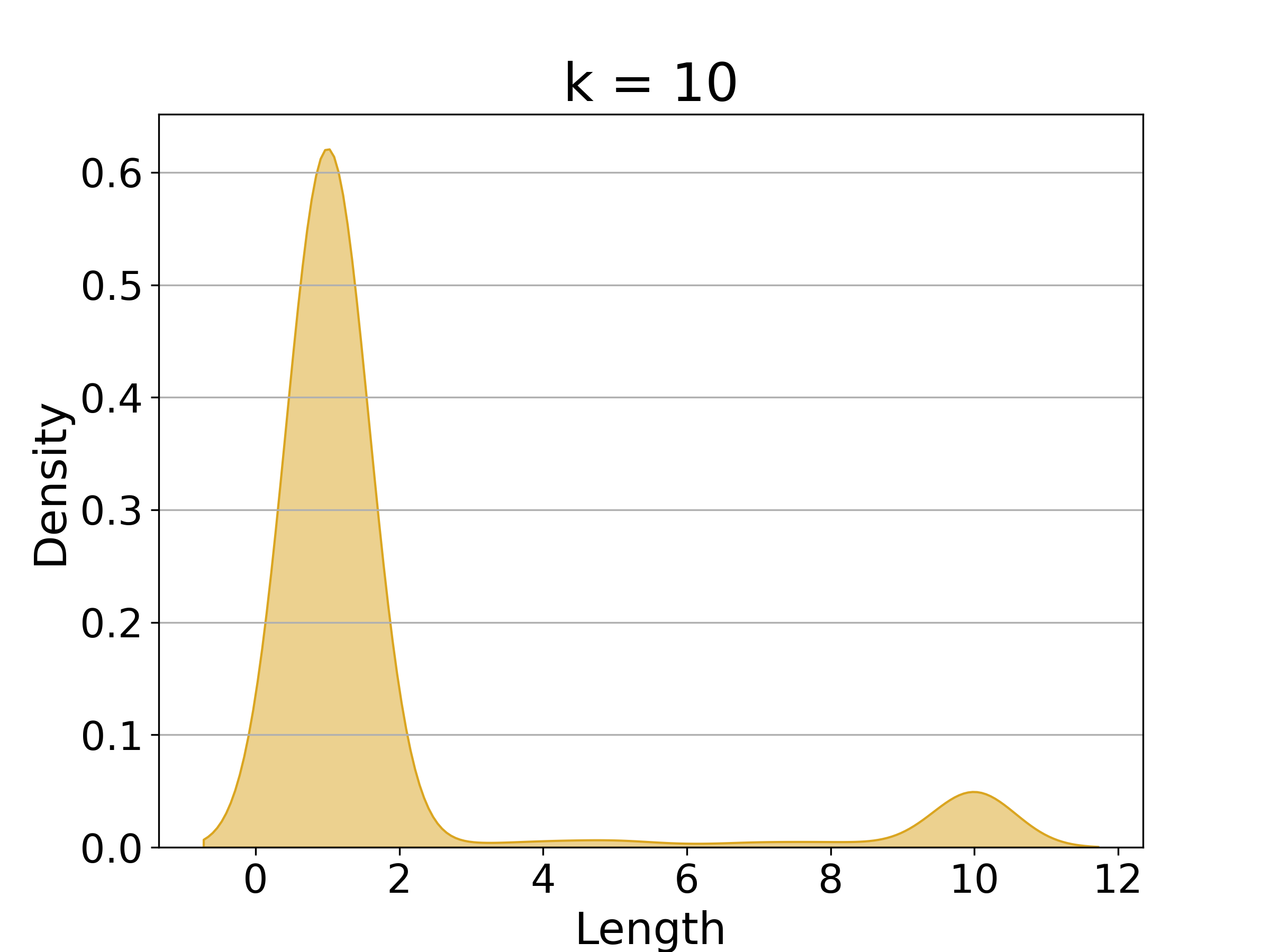}
            \caption{}
            \label{fig:plot8}
        \end{subfigure}
        \caption*{$\mu_{\text{threshold}} = 60\%$.}  % Unnumbered caption
    \end{center}

    \caption{Comparison of pathlet length distributions for different values of $k$ under two different $\mu_{\text{threshold}}$ settings (80\% and 60\%) using Kernel Density Estimation.}
    \label{fig:combined_side_by_side_plots}
\end{figure}

% \begin{figure}[t]
%     \centering
%     \includegraphics[width=0.9\linewidth]{img/parameter-sensitivty.png}
%     \caption{Parameter sensitivity analysis: the impact of $\alpha_1$, $\alpha_2$, $\alpha_3$, and $\alpha_4$ on the quality of \textsc{PathletRL}'s PD, expressed as the PD's attributes $|\mathbb{S}|$, $\phi$, $L_{traj}$, and $\mu$ respectively.}
%     \label{fig:param-sens}
% \end{figure}
\section{PathletRL Optimizations}
\label{sec:pathletrl-optimization}

Building on the parameter sensitivity analysis, which showed how variations in the four $\alpha$ parameters influence the Pathlet Dictionary (PD) in terms of the different objectives, we now shift our focus to addressing the observed limitations in the original \textsc{PathletRL} model. Specifically, the analysis highlighted the model's sensitivity to the weight assignments in the linear scalarization of the reward function, which can significantly impact the quality of the extracted PD.

To mitigate this sensitivity and further enhance the model's performance, we propose three variations of the \textsc{PathletRL} framework. The first two variations introduce alternative scalarization techniques to better balance the competing objectives: (1) \textsc{PathletRL} with Chebyshev Scalarization (\textsc{PathletRL}\textsubscript{Cheb}) and (2) \textsc{PathletRL} with a Custom Dynamic Scalarization Function (\textsc{PathletRL}\textsubscript{DS}). The third variation, \textsc{PathletRL++}, extends \textsc{PathletRL}\textsubscript{DS} by incorporating a richer state representation. This enhancement provides the reinforcement learning agent with a more detailed and informative view of the environment, enabling more nuanced decision-making and ultimately leading to improved performance in reducing the size of the candidate pathlet set.

\subsection{Scalarization Sensitivity in Multi-Objective Optimization}

The original \textsc{PathletRL} model employs a linear scalarization approach for reward calculation, equally weighting all of the objectives (\ref{eq:reward}). This linear scalarization suffers from known sensitivity issues in multi-objective optimization (MOO), as small changes in weight assignment can lead to substantially different optimization outcomes \cite{vanmoffaert2013scalarized}. Specifically, Linear scalarization only guarantees optimal solutions when the Pareto front (the set of optimal trade-offs between objectives) is convex. If the Pareto front is non-convex, this method fails to identify non-convex Pareto optimal solutions. This limitation means that some potentially desirable trade-offs between objectives may be missed entirely \cite{hayes2022practical}. To address these issues, we propose two alternative scalarization methods to improve the stability and generalization of solutions.

\subsubsection{Chebyshev Scalarization}

The first alternative is Chebyshev scalarization, which seeks to minimize the maximum deviation from the ideal point in the objective space. This approach is formally expressed as:

\begin{equation}
    R_{\text{Chebyshev}} = -\max \left( \alpha_1 \left| f_1(\mathbf{x}) - z_1^* \right|, \alpha_2 \left| f_2(\mathbf{x}) - z_2^* \right|, \alpha_3 \left| f_3(\mathbf{x}) - z_3^* \right|, \alpha_4 \left| f_4(\mathbf{x}) - z_4^* \right| \right),
\end{equation}
where \( \mathbf{x} \) represents the decision variables, \( f_1(\mathbf{x}), f_2(\mathbf{x}), f_3(\mathbf{x}), f_4(\mathbf{x}) \) are the values of the four objective functions, and \( z_1^*, z_2^*, z_3^*, z_4^* \) are the ideal values of the respective objectives. The weights \( \alpha_1, \alpha_2, \alpha_3, \alpha_4 \) are user-defined scaling factors, often set such that \( \sum_{i=1}^{4} \alpha_i = 1 \).\\
Chebyshev scalarization does not depend heavily on weight allocations and can find Pareto-optimal solutions in both convex and non-convex regions of the Pareto front \cite{vanmoffaert2013scalarized}. It ensures balanced optimization across all objectives and leads to robust solutions.

\subsubsection{Dynamic Scalarization}

The second alternative introduces a dynamic scalarization function that adjusts weights based on the agent’s proximity to predefined thresholds for trajectory representability \( \mu_{\text{traj}} \) and trajectory loss \( L_{\text{traj}} \). This method is captured by dynamically adjusting the weights for different objectives using penalties. These penalties are calculated based on the agent's proximity to critical thresholds. \\
The dynamic weight assignment is implemented in the following way:

\[
    w_{\text{traj}}(t) = \frac{1}{\text{max}\left( 0.01, \frac{M - L_{\text{traj}}(t)}{M} \right)}
\]
\[
    w_{\mu}(t) = \frac{1}{\text{max}\left( 0.01, \frac{\mu_{\text{traj}}(t) - \tau_{\mu}}{0.2} \right)}
\]
where \( w_{\text{traj}}(t) \) and \( w_{\mu}(t) \) are dynamic weights for trajectory loss and trajectory representability, respectively. The weight for each objective increases as the agent approaches the thresholds \( \tau_{\mu} \) for \( \mu_{\text{traj}} \) and \( M \) for \( L_{\text{traj}} \), encouraging the agent to prioritize these objectives when performance is critical.\\
The reward is calculated using these dynamic weights:

\begin{equation}
    R = -\alpha_1 \Delta|S| -\alpha_2 \Delta \phi -\alpha_3 \Delta L_{\text{traj}} w_{\text{traj}}(t) + \alpha_4 \Delta \bar{\mu} w_{\mu}(t),
\end{equation}
Initially, all objectives are weighted equally, but as the agent approaches critical thresholds (e.g., the trajectory representability threshold \( \tau_{\mu_{\text{traj}}} \)).
, the dynamic weights \( w_{\text{traj}}(t) \) and \( w_{\mu}(t) \) increase, forcing the agent to prioritize those objectives more.\\
At the end of an episode, an additional reward or penalty is applied based on how well the agent performed overall, considering the same objectives and thresholds. This dynamic adjustment ensures that the agent becomes more conservative when nearing critical thresholds, avoiding excessive pathlet merging that could degrade the overall performance.

\subsection{Enhanced State Representation for Improved Decision-Making}

While the first two variations improve the scalarization mechanism, they still rely on the original state representation used by \textsc{PathletRL}, which is limited to global metrics such as pathlet count, average trajectory representability, and trajectory loss. The state at time \( t \) is represented as:

\[
    \mathbf{s}_t = \left( S_1, S_2, S_3, S_4 \right),
\]
where \( S_1 \) denotes the number of pathlets in the current pathlet graph, \( S_2 \) denotes the average number of pathlets required to represent the trajectories, \( S_3 \) is the trajectory loss, and \( S_4 \) is the average trajectory representability. This restricted global view leaves the reinforcement learning agent unaware of crucial local information about individual pathlets and their neighbors, which can lead to suboptimal decision-making, especially when determining which pathlets to merge or split.\\
To address this shortcoming, we propose a third variation: \textsc{PathletRL++}. This model extends the dynamic scalarization method by incorporating a more detailed state representation, which provides the agent with weights of both the current pathlet and its neighboring pathlets. This enriched representation gives the agent a better view of its environment, enabling more informed and precise decisions during pathlet merging.

\subsubsection{Local Metrics for Improved State Representation}

The enhanced state representation at time \( t \), denoted as \( \mathbf{s}_t^{\text{enhanced}} \), is extended to include both global metrics and local pathlet weights. Specifically, the state is now formulated as:

\[
    \mathbf{s}_t^{\text{enhanced}} = \left( S_1, S_2, S_3, S_4, w_t, w_t^{\text{neighbors}} \right),
\]
where:
\begin{itemize}
    \item \( w_t \) represents the weight of the pathlet selected for processing at time \(t \) ($p_{\text{current}}$).
    
    \item \( w_t^{\text{neighbors}} \) denotes the weights of the selected pathlet's neighbours:
    \[
        w_t^{\text{neighbors}} = \left( w_p \right) \ \forall p \in \mathcal{N}(p_{\text{current}}),
    \]
    where \( \mathcal{N}(p_{\text{current}}) \) represents the set of neighboring pathlets to $p_{\text{current}}$.
\end{itemize}

\begin{figure}[t]
    \centering
    \begin{subfigure}[b]{0.49\textwidth}
        \centering
        \includegraphics[width=\textwidth]{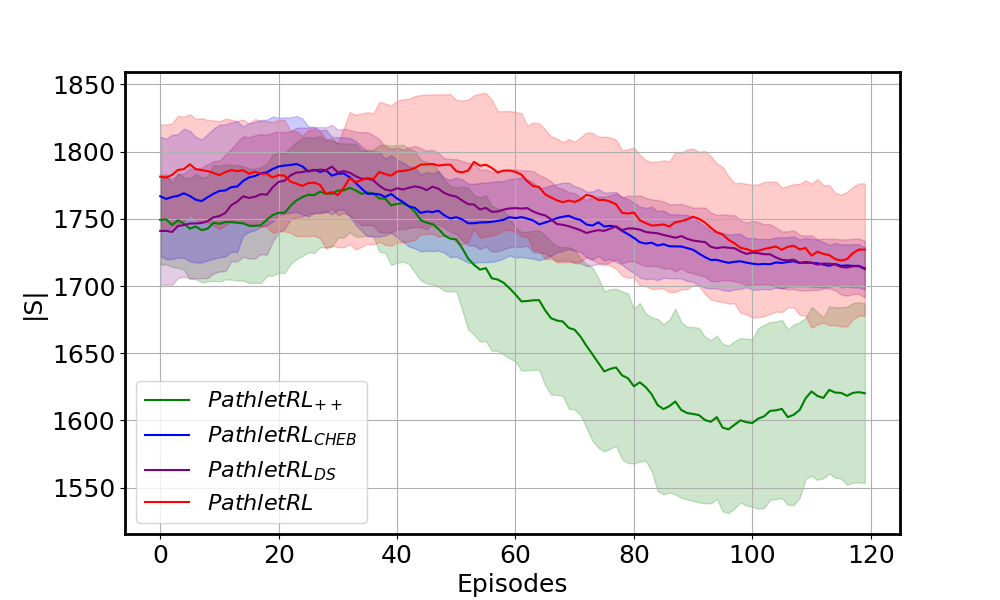}
        \subcaption{Toronto}
        \label{fig:training-toronto}
    \end{subfigure}\hfill
    \begin{subfigure}[b]{0.49\textwidth}
        \centering
        \includegraphics[width=\textwidth]{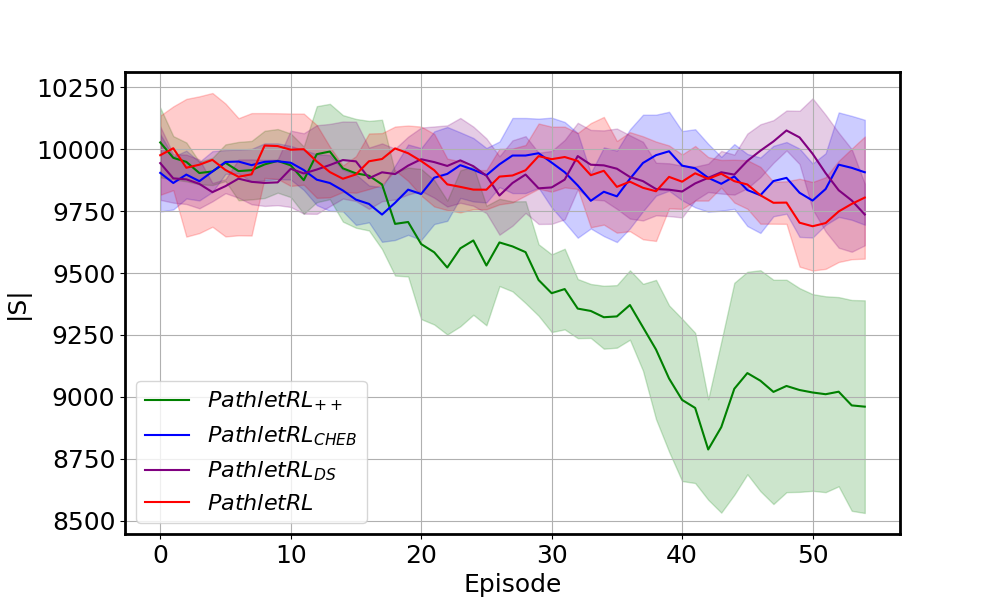}
        \subcaption{Rome}
        \label{fig:training-rome}
    \end{subfigure}
    \caption{Training performance of different \textsc{PathletRL} variations on the Toronto and Rome datasets.}
    \label{fig:training-both}
\end{figure}

\subsection{Comparative Analysis}

To maintain evaluation consistency, all models were tested using the same Toronto map and trajectory dataset. Additionally, we utilized a modified version of the Rome dataset to challenge the models with a larger set of trajectories. Typically, taxi cabs send GPS signals every 15 seconds. In our modification, the trajectory data were map matched using a shorter separation time frame of 60 seconds—if a taxi did not send a signal for over a minute, its path was considered a new trajectory. This approach resulted in a significantly larger number of trajectories (75,180).
\subsubsection{Training Performance Analysis}

To better understand the training dynamics of the proposed variations, we analyze the evolution of pathlet dictionary size with respect to the number of training episodes. Figures \ref{fig:training-toronto} and \ref{fig:training-rome} show the pathlet size reduction over the training episodes for each model variant on the Toronto and Rome datasets, respectively.\\
Focusing first on the Toronto dataset, it is evident that all models demonstrate a downward trend in pathlet size, indicating their learning to minimize the dictionary size effectively before reaching predefined thresholds. Notably, \textsc{PathletRL++} converges to a smaller pathlet dictionary size more rapidly than the other models. This swift convergence highlights the model’s enhanced training efficiency, which can be attributed to its detailed observation of the environment and robust reward system.\\
In contrast, when analyzing the Rome dataset, the original \textsc{PathletRL} variants with the standard state space representation encounter difficulties in learning meaningful actions due to the large number of trajectories. Despite this challenge, \textsc{PathletRL++} again demonstrates a rapid learning rate and achieves a substantial reduction in pathlet size.\\
Overall, it is clear that integrating local pathlet metrics within \textsc{PathletRL++} not only enhances the final performance but also accelerates the training process. This results in more optimal solutions being reached in fewer training episodes.

\begin{table}[t]
\setlength{\tabcolsep}{2.5pt}
\centering
\begin{tabular*}{0.8\textwidth}{@{\extracolsep{\fill}} c|l|cccc}
    \toprule
    \multicolumn{2}{c}{} & \multicolumn{4}{c}{\textsc{PathletRL Variations}} \\
    \cmidrule{3-6}
    \multicolumn{2}{c}{} & \textsc{PathletRL} & \textsc{PathletRL\textsubscript{Cheb}} & \textsc{PathletRL\textsubscript{DS}} & \textsc{PathletRL++} \\
    \midrule
    \parbox[t]{1mm}{\multirow{4}{*}{\rotatebox[origin=c]{90}{\textsc{\scriptsize{Toronto}}}}} & $[\downarrow] \hspace{1mm} |\mathbb{S}|$ & 1762 & 1736 & 1735 & \textbf{1646} \\
    & $[\downarrow] \hspace{1mm} \phi$ & 3.78 & 3.77 & \textbf{3.76} & 3.86 \\
    & $[\downarrow] \hspace{1mm} L_{traj}$ & \textbf{15.8}\% & 16.6\% & 16.4\% & 17.4\% \\
    & $[\uparrow] \hspace{1mm} \bar{\mu}$ & 80.0\% & 80.0\% & 80.0\% & 80.0\% \\
    \midrule
    \parbox[t]{1mm}{\multirow{4}{*}{\rotatebox[origin=c]{90}{\textsc{\scriptsize{Rome}}}}} & $[\downarrow] \hspace{1mm} |\mathbb{S}|$ & 9811 & 9871 & 9801 & \textbf{9068} \\
    & $[\downarrow] \hspace{1mm} \phi$ & 19.76 & \textbf{19.68} & 19.69 & 19.84 \\
    & $[\downarrow] \hspace{1mm} L_{traj}$ & \textbf{1.8}\% & 1.8\% & 1.8\% & 1.9\% \\
    & $[\uparrow] \hspace{1mm} \bar{\mu}$ & 80.0\% & 80.0\% & 80.0\% & 80.0\% \\
    \bottomrule 
\end{tabular*}
\caption{Performance Comparison of PathletRL Variations on Toronto and Rome Datasets.}
\label{tab:pathletrl-criteria-comparison}
\end{table}

\subsubsection{Offline Performance Comparison}
Table \ref{tab:pathletrl-criteria-comparison} summarizes the performance of the original \textsc{PathletRL} model and its variants across the four key metrics on both the Toronto and Rome datasets.
For the Toronto dataset, \textsc{PathletRL++} achieves the smallest pathlet dictionary size $|\mathbb{S}|$, comprising only 1,646 pathlets, which is notably lower than the 1,762 pathlets used by the original \textsc{PathletRL} model. When considering the average number of pathlets representing each trajectory ($\phi$), the results are relatively similar across models, with \textsc{PathletRL\textsubscript{DS}} showing a slight advantage. In terms of trajectory loss ($L_{traj}$), the original \textsc{PathletRL} model performs best with a loss rate of 15.8\%. This outcome is expected since its larger pathlet dictionary size results in fewer trajectory merges, minimizing loss. All models maintain the same average representability ($\bar{\mu}$) of 80.0\% which is equal to the $\mu$ threshold that we set for the training.

On the more complex Rome dataset, \textsc{PathletRL++} again demonstrates superior performance by achieving a smaller pathlet dictionary size of 9,068 compared to over 9,800 for the other models. This indicates its scalability and efficiency when handling larger datasets. Similar to the Toronto results, the average number of pathlets per trajectory ($\phi$) remains comparable across models, with \textsc{PathletRL\textsubscript{Cheb}} achieving the lowest value. The trajectory loss ($L_{traj}$) is nearly identical across all models, ranging from 1.8\% to 1.9\%, reflecting the large volume of trajectories in the Rome dataset. Once again, all models maintain the same average representability ($\bar{\mu}$) of 80.0\%, consistent with the training threshold.\\
Although \textsc{PathletRL++} marginally increases the trajectory loss compared to the original \textsc{PathletRL} model on both datasets, this is offset by its ability to reduce the pathlet dictionary size more significantly. This reduction in pathlet size can lead to improved scalability and computational efficiency, making \textsc{PathletRL++} a preferable choice for larger datasets or scenarios.

\section{Related Work}
\label{sec:related-work}

% shortened
Our research is related to (i) {\em trajectory data mining}, (ii) {\em pathlet mining}, (iii) {\em subtrajectory clustering}, and (iv) {\em deep learning for spatiotemporal data}. We cover below some of the most significant efforts relevant to our work. Note that some related works have already been cited throughout the manuscript to maintain a focused discussion, so they are largely omitted in this section.

% ; some have already been cited throughout the paper to keep the discussion focused, so they are omitted here.
% Note that some related work have already been cited throughout the manuscript to keep the discussion focused, so they are mostly omitted here.

% \smallskip\noindent\textbf{Trajectory Data Mining.} Trajectory data mining has been an active research direction for a long time \cite{zheng2015trajectory, alturi2018spatiotemporal, hamdi2022spatiotemporal}. This high interest can largely be attributed to the rapid development and prominence of geospatial technologies \cite{kogut2020geospatial, datta2022future}, location-based smart devices \cite{huang2018lbs, ruth2022iot}, and the abundance of Global Positioning System (GPS) based applications \cite{rash2019gps, pandey2021gps}. The focus is on popular technical problems, including trajectory similarity \cite{deng2022similarity} and trajectory clustering \cite{han2022clustering}. 
%\smallskip\noindent\textbf{Trajectory Data Mining.} 
\subsection{Trajectory Data Mining}

Trajectory data mining has been an active research direction for a long time \cite{zheng2015trajectory, alturi2018spatiotemporal, hamdi2022spatiotemporal}. This high interest can largely be attributed to the rapid development and prominence of geospatial technologies \cite{datta2022future}, location-based smart devices \cite{ruth2022iot}, 
% and the 
% Global Positioning System (GPS)
abundance of GPS-based applications \cite{pandey2021gps}, and generation of massive trajectory datasets \cite{faraji2023point2hex}. The focus is on popular technical problems, including trajectory similarity \cite{fang2022similarity} and trajectory clustering \cite{han2022clustering}. Modeling trajectory data using graphs to address complex trajectory mining tasks has also been a popular approach in recent years. Example tasks include, but are not limited to 
% clustering \cite{khalil2013clustering}, 
similarity search \cite{fang2022similarity}, recovery \cite{chen2022trajrec}, node centrality computation \cite{pechlivanoglou2018centrality}, mining spatiotemporal interactions \cite{sawas2018tensor, sawas2019versatile}, learning semantic relationships of geographic areas \cite{mehmood2020semantics} and recommendations \cite{wang2022graphenhanced, wang2022astar}.

%\smallskip\noindent\textbf{Pathlet Mining}. 
\subsection{Pathlet Mining}

Pathlet mining focuses on discovering patterns and extracting knowledge by decomposing complex trajectories into a set of fundamental building blocks known as pathlets. These pathlets serve as essential units that can efficiently represent a wide range of trajectories, aiding in tasks such as trajectory compression, classification, and route planning.\\
One of the pioneering works in this area is by Chen et al. \cite{chen2013pathlet}, who introduced the concept of a pathlet dictionary for compressing and planning trajectories. Their goal was to capture shared spatial and temporal regularities among a collection of human trajectories by learning a set of common path segments, or pathlets, that can be concatenated to reconstruct the original trajectories. They formulated the problem as an integer linear programming (ILP) task, aiming to minimize both the size of the pathlet dictionary and the number of pathlets required to represent each trajectory.
To address the computational challenges of solving the ILP on large datasets, Chen et al. \cite{chen2013pathlet} proposed an efficient decoupled approach that optimizes a lower bound of the original objective function. This method scales linearly with the number of trajectories, making it practical for large-scale applications. Their approach allows for the extraction of meaningful pathlets that reflect common movement patterns within the data. The learned pathlet dictionary can be utilized in various applications, such as route planning and trajectory analysis, by enabling efficient reconstruction and representation of trajectories.\\
Zhou et al. \cite{zhou2008bagofsegments} proposed the bag of segments method to represent motion trajectories. Inspired by the ``bag of words'' model from text mining, their method represents each trajectory as a collection of segments assigned to codewords from a predefined dictionary. This approach transforms trajectories into vectors in the codeword space, enabling the use of traditional machine learning algorithms for trajectory classification and similarity search.
While both Chen et al. \cite{chen2013pathlet} and Zhou et al. \cite{zhou2008bagofsegments} focus on representing trajectories using fundamental building blocks, their methodologies differ. Chen et al. \cite{chen2013pathlet} emphasize trajectory compression and shared pattern extraction through an optimization framework aimed at minimizing the pathlet dictionary size and reconstruction cost. Zhou et al. \cite{zhou2008bagofsegments}, on the other hand, focus on transforming trajectories into a vector space representation using clustering techniques, facilitating trajectory analysis tasks.\\
Other notable contributions to the field include the work by Panagiotakis et al. \cite{panagiotakis2012representativeness}, who developed a method for identifying representative subtrajectories using global voting, segmentation, and sampling techniques. Their approach enhances trajectory summarization and pattern discovery by focusing on the most representative segments within a dataset. Additionally, Wang et al. \cite{wang2022representative} addressed the problem of finding the top $k$ representative routes that cover as many trajectories as possible under budget constraints and distance thresholds. They proposed three near-optimal solutions that efficiently solve this problem, offering valuable strategies for trajectory analysis under resource limitations.\\
However, these top-down approaches often face challenges related to high memory usage and redundant storage due to overlapping pathlets, as they generate a large number of candidate pathlets and select a subset for the dictionary. To address these limitations, we propose a bottom-up strategy that incrementally merges basic pathlets to build a compact and efficient pathlet dictionary. Our approach begins with unit-length pathlets and iteratively merges them while optimizing utility, defined using newly introduced metrics of trajectory loss and representability. By utilizing a deep reinforcement learning framework, specifically Deep Q-Networks (DQN), the method approximates the utility function and results in significant reductions in memory requirements—up to 24,000 times compared to baseline methods.

%\smallskip\noindent\textbf{Subtrajectory Clustering.} 
\subsection{Subtrajectory Clustering}

Pathlet mining is commonly framed as a subtrajectory clustering problem. Lee et al. \cite{lee2007traculus} proposed the \textsc{Traculus} algorithm, which partitions trajectories into line segments and then groups those segments that lie in similar dense regions to form clusters. This approach effectively captures frequently traversed paths by clustering spatially close trajectory segments.\\
Van Kreveld et al. \cite{vankrevald2011wellvisited} designed a novel measure for mining median trajectories, serving as cluster centroids of (sub)trajectories. Their method focuses on discovering well-visited routes by identifying median paths that represent common movement patterns within a set of trajectories.\\
Agarwal et al. \cite{agarwal2018subtrajectory} advanced the field by introducing a comprehensive model for subtrajectory clustering using the concept of a pathlet cover. Inspired by the classical set cover problem, their method aims to find a small set of pathlets that effectively represent the shared portions of trajectories in a dataset. In their model, each trajectory is considered as a concatenation of a small set of pathlets, allowing for possible gaps to account for unique segments or noise.
They formulated the subtrajectory clustering problem with a single objective function that balances three key factors: minimizing the number of pathlets selected, ensuring high-quality representation of subtrajectories by the pathlets (by minimizing the distance between subtrajectories and their assigned pathlets), and penalizing the presence of gaps in trajectories. Recognizing the NP-hardness of the problem, Agarwal et al. \cite{agarwal2018subtrajectory} developed efficient approximation algorithms with provable guarantees on solution quality and running time. Their approach leverages geometric properties of trajectories and employs strategies like greedy selection and geometric grouping to efficiently explore the space of possible pathlets without exhaustive enumeration.
In all these subtrajectory clustering methods, the cluster centroids are seen as popular segments traversed by many trajectories and can alternatively be viewed as pathlets. This perspective facilitates the discovery of shared movement patterns and supports applications such as trajectory summarization, anomaly detection, and efficient data compression.

%\vspace{-3pt}

%\smallskip\noindent\textbf{Deep Learning for Spatiotemporal Data.} 
\subsection{Deep Learning for Spatiotemporal Data}
In recent years, deep learning methods have been proposed for learning representations of spatiotemporal data \cite{wang2022deeplearning}. Chen et al. \cite{chen2024deeplearningtrajectorydata}, for instance, introduced a comprehensive review of deep learning applications for trajectory data management and mining. This work highlights the use of advanced AI techniques for various tasks like trajectory prediction, clustering, and anomaly detection. A key contribution is the provision of an open-source repository of trajectory-related datasets, enabling researchers to access curated resources for diverse mobility analysis tasks.
Moreover, a diffusion model, TrajGDM, was developed to simulate human mobility patterns, allowing more sophisticated generation of trajectory data based on real-world movement behaviors \cite{zhu2023difftrajgeneratinggpstrajectory}. These advancements indicate the growing adoption of generative models in trajectory mining.
Other works explore spatiotemporal reachability and the use of graph neural networks (GNNs) to mine complex relationships from trajectory data \cite{ijcai2024p286}. Recent work, such as Traj-LLM (Lan et al. \cite{lan2024trajllmnewexplorationempowering}), explores the use of Large Language Models (LLMs) in trajectory prediction tasks, leveraging their ability to capture complex traffic semantics and scene understanding to enhance prediction accuracy and adaptability. Of particular interest are deep reinforcement learning based methods that are often evaluated on agents playing a specific game \cite{sun2020program}. These methods have successfully been adapted and shown promise in addressing several complex trajectory-related problems, such as route planning \cite{geng2020drl}, trajectory simplification \cite{wang2021simplification}, and adaptive vehicle navigation problem \cite{arasteh2022vehicle}.

\section{Conclusions}
\label{sec:conclusions}

Constructing trajectory pathlet dictionaries—small sets of building blocks capable of representing large numbers of trajectories—has become crucial due to applications in route planning, travel time estimation, and trajectory compression. This work introduces a deep reinforcement learning solution, \textsc{PathletRL}, which generates a dictionary 65.8\% smaller than traditional methods. Remarkably, only half of the pathlets in \textsc{PathletRL}'s dictionary are needed to reconstruct 85\% of the original trajectory dataset, whereas baselines require their entire dictionaries to reconstruct just 65\% of the trajectories. Additionally, \textsc{PathletRL} achieves significant memory savings—up to $\sim$24,000$\times$ compared to existing methods—due to reduced initial memory requirements for storing pathlets. To further enhance \textsc{PathletRL}, we proposed three new variations to address its reward function limitations and to incorporate richer and more meaningful state representations. These improvements led to a more compact dictionary, faster convergence, and better overall performance in terms of dictionary size and training efficiency, making \textsc{PathletRL++} the best-performing variant for large-scale trajectory datasets.

\bibliographystyle{ACM-Reference-Format}


\appendix
\clearpage

\huge
\smallskip\noindent \textbf{APPENDIX}

\normalsize

\section{Pathlet Dictionary's Applications}
\label{sec:applications}

\smallskip\noindent \textbf{Trajectory Compression}. Compression is the process of reducing the size of a trajectory while preserving its important spatiotemporal features \cite{muckell2010compression}; it is more commonly useful when there is a need of transmitting or storing large trajectory datasets in much smaller, more limited resources (e.g., mobile devices, wireless networks, etc.). A common technique for trajectory compression involves sampling points in trajectories that carry the most significant amount of spatiotemporal information. With a pathlet dictionary, we are able to capture the pathlets in a trajectory's pathlet-based representation set that are most useful and could represent a large number of trajectories in the dataset.

% In planning a route from $A$ to $B$, possessing a high-quality pathlet dictionary might be useful. With pre-computed pathlets, one can quickly access the relevant ones in the dictionary that is required to form the path needed to reach the destination point from the origin source. 
\smallskip\noindent \textbf{Route Planning}. While navigation services and route planning apps such as Google Maps\footnote{\url{http://maps.google.com/}} and Apple Maps\footnote{\url{https://www.apple.com/maps/}} exist, they are not necessarily accessible to users when online (internet) connection is not available (e.g., network outages, poor wifi signals in remote areas, etc.). As such, there is a need for accessing maps and loading mobility data offline while at the same time efficiently answering majority of user queries (for example, path recommendation from Point $A$ to $B$). Pathlet dictionaries can be
% of good use
useful in this
% specific use
case.

% \smallskip\noindent \textbf{Trajectory Prediction}. Prediction and forecasting of trajectories is important in several domains including transportation \cite{markovic2019transportation, li2020transportation}, urban planning \cite{hassan2020urban, li2021urban}, 
% % environmental conservation \cite{nyhan2016environment}
% and healthcare \cite{schulam2016dtm, pechlivanoglou2022microscopic, strzelecki2022apple}. Pathlets could be useful by expressing trajectories as sequences of pathlets, predicting the next pathlet(s) in the sequence and then concatenating these predicted pathlets to form the predicted trajectories.

\smallskip\noindent \textbf{Trajectory Prediction}. Prediction and forecasting of trajectories is important in several domains including transportation \cite{li2020transportation}, urban planning \cite{li2021urban}, 
% environmental conservation \cite{nyhan2016environment}
and healthcare \cite{pechlivanoglou2022microscopic, strzelecki2022apple}. Pathlets could be useful by expressing trajectories as sequences of pathlets, predicting the next pathlet(s) in the sequence and then concatenating these predicted pathlets to form the predicted trajectories.

\smallskip\noindent \textbf{Anomaly Detection}. Spotting outliers has been an active research direction. One could use pathlets to detect trajectories that move or behave anomalously by expressing each trajectory as their pathlet-based representation set and then identifying which of those trajectories has a pathlet-based representation set that deviates from other trajectories.

% \smallskip\noindent \textbf{Trajectory Similarity Search}. The trajectory similarity search task is an important problem of interest due to its numerous practical applications in domains such as traffic analysis \cite{elragal2014traffic, li2021traffic}, public safety \cite{he2020crime}, and environmental conservation \cite{nyhan2016environment}. With pathlets, the similarity of two trajectories can be calculated based on the cosine similarity (or some other similarity metric) of their pathlet-based representations.

\smallskip\noindent \textbf{Trajectory Similarity Search}. The trajectory similarity search task is an important problem of interest due to its numerous practical applications in domains such as traffic analysis \cite{li2021traffic}, public safety \cite{he2020crime}, and environmental conservation \cite{nyhan2016environment}. With pathlets, the similarity of two trajectories can be calculated based on the cosine similarity (or some other similarity metric) of their pathlet-based representations.
\section{Proof of Theorem \ref{thm:traj-rep-iter}}
\label{sec:proof-thm-traj-rep-iter}
Before proceeding with the proof, we first introduce two facts that will prove useful in our main proof for the theorem.

\smallskip\noindent \textbf{Fact 1}. The pathlet length of some pathlet $\rho_A \in \mathcal{P}$, plus the pathlet length of a neighboring pathlet $\rho_B \in \mathcal{P}$ is equal to the pathlet length of $\rho_{AB}$, where $\rho_{AB}$ is the pathlet formed when pathlets $\rho_A$ and $\rho_B$ are merged. In other words:
\begin{equation}
    \ell(\rho_A) + \ell(\rho_B) = \ell(\rho_{AB})
\end{equation}
This fact is fairly intuitive and does not require further explanation. Note that either one of $\rho_A$ or $\rho_B$ (or both) has length 1\footnote{This is because we would never find two higher-ordered (i.e., a pathlet with length greater than 1) pathlets merging together. The current pathlet in process can only merge with a neighboring pathlet that have never been processed in the algorithm (they all have length 1) and any neighboring high-ordered pathlets have already been preprocessed and added to the dictionary.}.

\smallskip\noindent \textbf{Fact 2}. The sequence of trajectory representabilities $\{ \mu_i \}$ for some trajectory $\tau \in \mathcal{T}$ is monotonically non-increasing. In other words, the trajectory representability of $\tau$ at some iteration $j$ of the algorithm, for some $i < j$ is less than or equal to its representability at iteration $i$:
\begin{equation}
    \mu_i(\tau) \geq \mu_j(\tau)
\end{equation}
This claim is straightforward and intuitive as it simply stems from the nature of Algorithm \ref{algo:pathlet-merge-algorithm}. As a remark, if $j = i + 1$, then $\mu_i(\tau) \geq \mu_{i+1}(\tau)$. This then implies that:
\begin{equation}
    \mu_i(\tau) = \mu_{i+1}(\tau) + \epsilon
    \label{eq:i-i+1}
\end{equation}
for some non-negative $\epsilon \geq 0$.

\smallskip\noindent \textbf{\underline{Proof of Theorem \ref{thm:traj-rep-iter}}}.

We are then ready to provide proof to this theorem by means of induction. First, we claim that the theorem holds at the initial base case, i.e., when $i = 0$:
    $$\mu_0(\tau) = \dfrac{\sum_{\rho' \in \Phi_0(\tau)} \ell(\rho') }{ \sum_{\rho \in \Phi_0(\tau)} \ell(\rho) } = 1 = 100\%$$
which checks out since the trajectory representabilities of all trajectories $\tau \in \mathcal{T}$ have this representability value at the beginning of the algorithm. Now, assume that the theorem holds at some iteration $i = n \geq 0$:
    $$\mu_n(\tau) = \dfrac{\sum_{\rho' \in \Phi_n(\tau)} \ell(\rho') }{ \sum_{\rho \in \Phi_0(\tau)} \ell(\rho) }$$
All it remains to complete the proof is to show that the theorem holds for when $i = n+1$, i.e.,
    $$\mu_{n+1}(\tau) = \dfrac{\sum_{\rho' \in \Phi_{n+1}(\tau)} \ell(\rho') }{ \sum_{\rho \in \Phi_0(\tau)} \ell(\rho) }$$
Now let current pathlet $\rho_A$ be the pathlet that merges with its neighbor pathlet $\rho_B$ to form $\rho_{AB}$ at the iteration $(n+1)$. At this point, there are three cases to consider.

\smallskip\noindent \textbf{\underline{Case 1}}. $\rho_A \notin \Phi_n(\tau)$ and $\rho_B \notin \Phi_n(\tau)$

This is the case when both these two pathlets that are candidate for merging are not in the pathlet-based representation set $\Phi_n(\tau)$. This means that there is no change to the pathlet-based representation of $\tau$ from iteration $n$ to $(n+1)$; i.e., $\Phi_{n+1}(\tau) = \Phi_n(\tau)$, as well as their representabilities at iteration $(n+1)$, i.e., $\mu_{n+1}(\tau) = \mu_n(\tau)$.
\begin{align*}
    \mu_{n+1}(\tau) = \mu_n(\tau) = \dfrac{\sum_{\rho' \in \Phi_n(\tau)} \ell(\rho') }{ \sum_{\rho \in \Phi_0(\tau)} \ell(\rho) } = \dfrac{\sum_{\rho' \in \Phi_{n+1}(\tau)} \ell(\rho') }{ \sum_{\rho \in \Phi_0(\tau)} \ell(\rho) }
\end{align*}

\smallskip\noindent \textbf{\underline{Case 2}}. $\rho_A \in \Phi_n(\tau)$ and $\rho_B \in \Phi_n(\tau)$

In this case, both $\rho_A$ and $\rho_B$ are in $\Phi_n(\tau)$ and therefore their merged pathlet $\rho_{AB}$ will be in the pathlet-based representation of $\tau$ at the next iteration $(n+1)$, i.e., $\rho_{AB} \in \Phi_{n+1}(\tau)$. Note that that both $\rho_A$ and $\rho_B$ will no longer be in $\Phi_{n+1}(\tau)$. Moreover in this case, it is clear as well that their representabilities of $\tau$ also do not change between iterations $n$ and $(n+1)$. Thus we have:
\begin{align*}
    \mu_{n+1}(\tau) & = \mu_n(\tau) \\
    & = \dfrac{\sum_{\rho' \in \Phi_n(\tau)} \ell(\rho') }{ \sum_{\rho \in \Phi_0(\tau)} \ell(\rho) } \tag{by the Inductive Hypothesis} \\
    & = \dfrac{\sum_{\rho' \in \Phi_n(\tau) \setminus \{\rho_A, \rho_B\} } \ell(\rho') + \ell(\rho_A) + \ell(\rho_B) }{ \sum_{\rho \in \Phi_0(\tau)} \ell(\rho) } \tag{separate $\rho_A$ and $\rho_B$} \\
    & = \dfrac{\sum_{\rho' \in \Phi_n(\tau) \setminus \{\rho_A, \rho_B\} } \ell(\rho') + \ell(\rho_{AB}) }{ \sum_{\rho \in \Phi_0(\tau)} \ell(\rho) } \tag{by Fact 1} \\
    & = \dfrac{\sum_{\rho' \in \Phi_n(\tau) \setminus \{\rho_A, \rho_B\} \cup \{ \rho_{AB} \} } \ell(\rho')  }{ \sum_{\rho \in \Phi_0(\tau)} \ell(\rho) } \tag{inclusion of the new merged pathlet $\rho_{AB}$ } \\
    & = \dfrac{\sum_{\rho' \in \Phi_{n+1}(\tau)} \ell(\rho') }{ \sum_{\rho \in \Phi_0(\tau)} \ell(\rho) } \tag{as $\Phi_{n+1}(\tau) = \Phi_n(\tau) \setminus \{\rho_A, \rho_B\} \cup \{ \rho_{AB} \}$ }          
\end{align*}
Note that in the last step, to get $\Phi_{n+1}(\tau)$, this is simply the same as the previous $\Phi_n(\tau)$ except we take out the two pathlets that are candidate for merging $\{ \rho_A, \rho_B \}$ and then add back the newly formed merged pathlet $\{ \rho_{AB} \}$.

\smallskip\noindent \textbf{\underline{Case 3}}. $\rho_A \in \Phi_n(\tau)$ and $\rho_B \notin \Phi_n(\tau)$

Without loss of generality, this case should suffice for when only one of $\{ \rho_A, \rho_B \}$ is in $\Phi_n(\tau)$. Unlike Cases 1 and 2 where the representability of $\tau$ did not change, here in Case 3 $\tau$'s representability will decrease. Following from Fact 2, there exists some positive $\epsilon > 0$ that satisfies the equality in Equation (\ref{eq:i-i+1}). This equation can be rewritten as:
\begin{equation}
    \mu_{n+1}(\tau) = \mu_n(\tau) - \epsilon
    \tag{$*$}
\end{equation}
(where $i$'s are expressed as as $n$'s). From here, it is not difficult to see that the $\epsilon$ we are after is basically the percentage of the trajectory that can no longer be represented by the pathlets in the dictionary at the next iteration $(n+1)$. Since we know that $\rho_A$, which was part of $\Phi_n(\tau)$ will no longer be part of $\Phi_{n+1}(\tau)$ due to its merge with $\rho_B$ to form $\rho_{AB}$ (and moreover, $\rho_{AB}$ is not part of $\Phi_{n+1}(\tau)$ as $\rho_B \notin \Phi_n(\tau)$), then it must be that the portion covered by $\rho_A$ in $\tau$ is the loss and is the $\epsilon$ that we want:
\begin{equation}
    \epsilon = \dfrac{ \ell(\rho_A) }{ \sum_{\rho \in \Phi_0(\tau)} \ell(\rho) } > 0
    \tag{$**$}
\end{equation}
Note that this value is positive as both $\ell(\rho_A) > 0 $ and $\sum_{\rho \in \Phi_0(\tau)} \ell(\rho) > 0$. So now we have:
\begin{align*}
    \mu_{n+1}(\tau) & = \mu_n(\tau) - \epsilon \tag{by $(*)$} \\
    & = \dfrac{\sum_{\rho' \in \Phi_n(\tau)} \ell(\rho') }{ \sum_{\rho \in \Phi_0(\tau)} \ell(\rho) } - \epsilon \tag{by the Inductive Hypothesis} \\
    & = \dfrac{\sum_{\rho' \in \Phi_n(\tau)} \ell(\rho') }{ \sum_{\rho \in \Phi_0(\tau)} \ell(\rho) } - \dfrac{ \ell(\rho_A) }{ \sum_{\rho \in \Phi_0(\tau)} } \tag{by $(**)$} \\
    & = \dfrac{\sum_{\rho' \in \Phi_n(\tau)} \ell(\rho') - \ell(\rho_A) }{ \sum_{\rho \in \Phi_0(\tau)} \ell(\rho) }   \\
    & = \dfrac{\sum_{\rho' \in \Phi_n(\tau) \setminus \{  \rho_A \} } \ell(\rho') + \ell(\rho_A) - \ell(\rho_A) }{ \sum_{\rho \in \Phi_0(\tau)} \ell(\rho) }  \tag{separate $\rho_A$ } \\
    & = \dfrac{\sum_{\rho' \in \Phi_n(\tau) \setminus \{  \rho_A \} } \ell(\rho')  }{ \sum_{\rho \in \Phi_0(\tau)} \ell(\rho) }   \\
    & = \dfrac{\sum_{\rho' \in \Phi_{n+1}(\tau)} \ell(\rho') }{ \sum_{\rho \in \Phi_0(\tau)} \ell(\rho) } \tag{as $\Phi_{n+1}(\tau) = \Phi_n(\tau) \setminus \{\rho_A \} $ } 
\end{align*}
Thus, on the basis of the inductive hypothesis, we just have shown that:
    $$\mu_{n+1}(\tau) = \dfrac{\sum_{\rho' \in \Phi_{n+1}(\tau)} \ell(\rho') }{ \sum_{\rho \in \Phi_0(\tau)} \ell(\rho) }$$
Clearly, the theorem holds true for all $i \geq 0$, which completes the proof. \hfill $\square$
\section{Space Complexity Analysis}
\label{sec:memory-proof}

% We analyze the space complexity of existing (top-down) methods that utilize overlapping pathlets vs our proposed (bottom-up) scheme that use edge-disjoint pathlets. Let $n = |\mathcal{E}|$ be the number of the road segments (also the initial length-1 pathlets) in some arbitrary road network $\mathcal{G}$. As existing methods consider initial pathlets of various sizes and configurations, then we can derive an upper bound on the amount of memory storage top-down methods require: $\mathcal{O} \left( \binom{n}{1} \right) + \mathcal{O} \left(\binom{n}{2} \right) + ... + \mathcal{O} \left( \binom{n}{n} \right) = \mathcal{O} \left(2^n \right)$. In contrast, our bottom-up approach only requires $\Theta(n)$ amount of memory that can clearly reduce this exponential bound to a linear space with respect to the number of segments in the road network. \hfill $\square$

To analyze the space complexity expressed in terms of the $n$ number of road segments on the road network, we can simply provide a lower bound on this number. First, we consider the analysis for top-down approaches. The following two facts would be useful in this proof (we will bypass the proofs of these facts as they are trivial; refer to linear algebra books for their complete proofs \cite{olver2018linearalgebra}).

\smallskip\noindent \textbf{Fact 1}. Let $a_{ij}$ be the entry situated at the $i$th row and $j$th column of an $m \times m$ square matrix $A$. Moreover, let $\vec{x}$ be a column vector of ones that has dimension $m \times 1$. Then the sum of all entries of $A$ is:
\begin{equation}
    \sum_{i=1}^m \sum_{j=1}^m a_{ij} = \vec{x}^\top A \vec{x}
    \label{eq:sum-entries}
\end{equation}

\smallskip\noindent \textbf{Fact 2}. For an $m \times m$ square symmetric matrix $A$ (i.e., $A = A^\top$) with $\lambda_{min}$ as its smallest eigenvalue, then the quadratic form $\vec{x}^\top A \vec{x}$ is lower bounded by the squared $L_2$-norm of $\vec{x}$, for all $\forall \vec{x} \in \mathbb{R}^{m \times 1}$:
\begin{equation}
     \vec{x}^\top A \vec{x} \geq \lambda_{min} ||\vec{x}||_2^2
    \label{eq:fact-symm}
\end{equation}
In other words, $\vec{x}^\top A \vec{x} \in \Omega(||\vec{x}||_2^2)$

\smallskip\noindent \textbf{Space Complexity Analysis}.
So now we focus firstly on the nodes (road intersections) of the road network. If one was to construct an adjacency matrix $A$ where entry $a_{ij} \in A$ equals 1 if there is an edge (or road segment) from $i$ to $j$ (or $j$ to $i$ as our road network is undirected by assumption) and 0 otherwise, then $a_{ij} \in A^\ell$ determines the number of paths of length $\ell$ to traverse node $i$ to $j$ (or vice-versa) on the road network represented by adjacency matrix $A$ \cite{newman2018networks}. Assuming the absence of self-loops, then the total number of paths of length $\ell$ for any pair of nodes in the road network is the sum of all entries in the upper diagonal (excluding the main diagonal) of $A^\ell$. We express it as follows. Let $Q_\ell = A^\ell - \mathcal{D}(A^\ell)$, where $\mathcal{D}(A)$ is the matrix that contains the diagonal entries of $A$ along its main diagonal and zeros elsewhere. And then the sum of the upper diagonal entries of $Q_\ell$ is simply $\frac{1}{2}\vec{x}^\top Q_{\ell} \vec{x}$, with $\vec{x}$ as $\text{dim}(A) \times 1$ column of ones (the sum of entries is in line with Fact 1 above in Equation (\ref{eq:sum-entries})). And then the half here is to avoid double counting (i.e., the undirected graph from $i$ to $j$ is the same as $j$ to $i$ as a result of symmetry). So clearly, if we want all the pathlets of at least length-1, then we can write it as follows:
\begin{equation}
    \sum_{\substack{\ell \in \mathbb{Z}^+ \\ \vec{x}^\top Q_\ell \vec{x} \neq 0}} \dfrac{1}{2} \vec{x}^\top Q_\ell \vec{x}
    \label{eq:sum-length-l}
\end{equation}
To yield a lower bound on this summation with respect to the number of road segments in the road network (instead of the nodes/road intersections), consider the smallest possible number of nodes for $n$ edges in the road network first. Each edge connects together two nodes; as a result, there can be at least $|\mathcal{V}| = n$ nodes (i.e., $\text{dim}(A) = n$). Thus, in Equation (\ref{eq:sum-length-l}), (assuming we have this least possible number of nodes) then it turns out that the $\text{dim}(Q_\ell) = n$ and $\text{dim}(\vec{x}) = n \times 1$ as a result of $\text{dim}(A) = n$.

Now, each of the terms in the summation of Equation (\ref{eq:sum-length-l}) is lower-bounded by $\Omega(||\vec{x}||_2^2) = \Omega(n)$ as a consequence of the Fact 2 in Equation (\ref{eq:fact-symm}). To see why $\Omega(||\vec{x}||_2^2)$ reduces to $\Omega(n)$, first note that $\vec{x}$ is a column vector of ones with dimension $n \times 1$. Therefore, its squared $L_2$-norm is: $\left(\sqrt{1^2 + 1^2 + ... + 1^2} \right)^2 = \left( \sqrt{n} \right)^2 = n$. Note however that the eigenvalue $\lambda_{min}$ should not be relevant towards the space complexity expressed in terms of the number of road segments. However, such space analysis is only for one of the $n$ terms in the summation of Equation (\ref{eq:sum-length-l})\footnote{It can be noted that at the worst case, the longest pathlet length in the road network is $n$; so that is why the summation can have up to $n$ terms}; thus the lower bound for the number of pathlets of the top-down scheme is $n \times \Omega(n) = \Omega(n^2)$.

% $$\left(\sqrt{\underbrace{1^2 + 1^2 + ... + 1^2}_{n}} \right)^2 = \left( \sqrt{n} \right)^2 = n$$

On the other hand, the proposed bottom-up method \textsc{PathletRL}, which consumes less memory space to initially store the pathlets, is more efficient with $\Theta(n)$ memory space complexity. It is quite easier to analyze. Since we only consider length-1 edge-disjoint pathlets, it is easy to see how ours simply relies exactly on the number of segments in the road network -- which happens to be $n$, and is therefore more space efficient compared to the top-down approaches the requires at least $\Omega(n^2)$ amount of memory space. \hfill $\square$

% \section{Summary of Notations}
% \label{sec:nomenclature}

% \setlength{\textfloatsep}{-3pt}
% \begin{table}[h]
%     \centering
%      \setlength{\tabcolsep}{1pt}
%     \begin{tabular}{cl}
%         \toprule
%          \textbf{Symbol} & \multicolumn{1}{c}{\textbf{Definition}}  \\
%          \midrule
%          $\tau$ and $\mathcal{T}$         & A trajectory $\tau$ and a set of trajectories $\mathcal{T}$ \\
%         $\mathcal{G} \langle \mathcal{V}, \mathcal{E} \rangle$ & Road network with intersections $\mathcal{V}$ and road segments  $\mathcal{E}$ \\
%          $\rho$ and $\mathcal{P}$         & A pathlet $\rho$ and a pathlet set $\mathcal{P}$ \\
%          $\mathcal{G}_p \langle \mathcal{V}_p, \mathcal{E}_p \rangle$ & The pathlet graph representation of road network $\mathcal{G}$ \\
%          $\Phi(\tau)$                     & The pathlet-based representation of a trajectory $\tau$ \\
%          $\Lambda(\rho)$                  & The trajectory traversal set of a pathlet $\rho$ \\
%          $\mu(\tau)$                      & The trajectory representability of trajectory $\tau$ \\
%          $L_{traj}$                       & The trajectory loss \\
%          $\mathbb{S}$                     & The pathlet dictionary \\
%          $\phi$                           & The avg \# of pathlets representing each $\tau$ in $\mathcal{T}$ \\
%          \bottomrule
%     \end{tabular}

%     \caption[Summary of notation]{Summary of notation used in this work}
%     \label{tab:nomenclature}
    
% \end{table}

% \vspace{-15pt}

\section{Map-Matching}
\label{sec:map-matching}

\begin{figure}[h]
    \centering
    \includegraphics[width=0.40\textwidth]{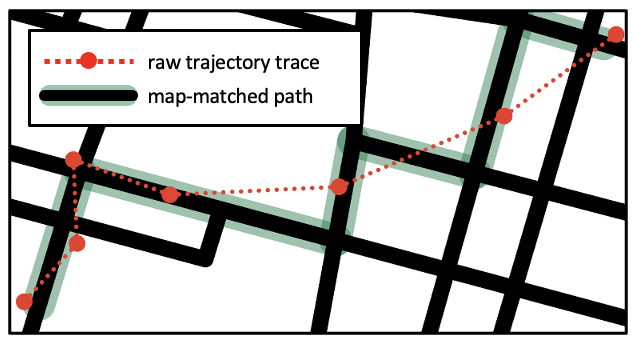}
    \caption{An example of the map-matching procedure.}
    \label{fig:map-matching-ex}
\end{figure}

Map-matching is a common task that identifies the path on the road an object has taken given a sequence of GPS locations \cite{newson2009hmm} (Fig. \ref{fig:map-matching-ex} provides an illustrative example). Ideally, we prefer highly accurate map-matched data from GPS trajectory traces; however, this task is itself involved and is outside the paper's scope. Thus, we rely on existing methods \cite{lou2009stmatching, yuan2010ivmm} to handle map-matching. 

\section{Other Deep Reinforcement Learning Policies}
\label{sec:drl-policies}

\balance

The choice of \textsc{Dqn} method over others such as Actor-Critic (\textsc{A2c}) and Policy-gradient is due to a number of reasons. For one, \textsc{Dqn}s are more sample efficient than Policy-gradient methods because learning happens using an experience replay buffer rather than learning from data collected with the current policy. Moreover, with small action spaces, \textsc{Dqn}'s are more stable and more efficient than \textsc{A2c} methods; a separate neural network computes (approximates) the target $Q$-values, which can reduce the variance in these estimates. In addition, implementing \textsc{Dqn}s is more straightforward compared to Actor-critic and Policy-gradient methods that tend to be more complex in terms of architectures and hyperparameter tuning. This makes them easier to employ in real-world settings.

\section{Dataset Statistics}
\label{sec:dataset-stats}

\setlength\belowcaptionskip{0pt}
\begin{table}[h]
\centering
  \begin{tabular}{c|lcc}
    \toprule
    \multicolumn{1}{c}{} & \multicolumn{1}{c}{\textbf{Feature}} & \textbf{\textsc{Toronto}} & \textbf{\textsc{Rome}} \\
    \midrule
    \parbox[t]{1mm}{\multirow{2}{*}{\rotatebox[origin=c]{90}{\textit{\scriptsize{roadmap}}}}} & \textbf{\# nodes} & 1.9K & 7.5K \\
    \\[-0.8em]
    & \textbf{\# edges / initial pathlets} & 2.5K & 15.4K \\
    \midrule
    \parbox[t]{1mm}{\multirow{4}{*}{\rotatebox[origin=c]{90}{\textit{\scriptsize{trajectories}}}}} & \textbf{Trajectory type} & realistic synthetic & real world  \\
    & \textbf{Object} & cars & taxis  \\
    & \textbf{\# Total geolocation data points} & 169K & 3.8M \\
    & \textbf{\# Total trajectories} & 2.9K & 75K \\
  \bottomrule 
\end{tabular}
\caption{Dataset attributes}
\label{tab:data-attributes}

\end{table}

% \vspace{-5pt}

\section{Privacy and Ethics}
\label{sec:privacy-ethics}

Our proposed methodology makes use of real-world trajectory datasets in the experiments that could potentially raise concerns about the individual (vehicle taxi cab) privacy and the potential for re-identification of such individuals. Therefore, to ensure protection of privacy and ethical considerations, all datasets used in our evaluation have been anonymized. Such datasets are derived from publicly-available sources, are publicly available, and are free of use for the intention of research purposes as outlined by their curators. In addition, all terms and conditions of use have been followed, with proper attribution and citation of works.

\section{Implementation Details}
\label{sec:imp}

\smallskip\noindent \textbf{Machine Specifications}. We conducted experiments on a server equipped with an NVIDIA RTX A6000 graphics card and 320GB of memory.

\smallskip\noindent \textbf{The RL Implementation}. We implemented our deep reinforcement learning architecture using the \verb|stable-baselines3|\footnote{\url{https://stable-baselines3.readthedocs.io}} package, a PyTorch-based library designed specifically for implementing RL methods.

\section{The Choice of Baselines}
\label{sec:baselines-choice}

While there have been a number of considerable baseline methods \cite{wang2022representative, panagiotakis2012representativeness, zhou2008bagofsegments} that can be used to compare \textsc{PathletRL} with, there are some reasons we decided to not do so. Firstly, Wang et al. \cite{wang2022representative} focuses on route representativeness discovery which is a completely different task than ours. Theirs, including Panagiotakis et al. \cite{panagiotakis2012representativeness} use a representativeness criterion that is not applicable to our case. Zhou et al. \cite{zhou2008bagofsegments} moreover focuses on motion analysis which is a substantially different setting and task. This is in contrast to Chen et al. \cite{chen2013pathlet} and Agarwal et al. \cite{agarwal2018subtrajectory}, where we have opted for their baselines methods due to a number of reasons. The former is the original and most representative work on pathlet dictionary construction, while the latter is the most representative/newest method on subtrajectory clustering that frames the pathlet dictionary construction as a subtrajectory clustering task. Both of these baselines also do not rely on learning-based methods, compared to our proposed model where we show the importance of deep learning.

As an aside, it is important to note that the technical problem of trajectory pathlet dictionary construction is novel and state-of-the-art methods are originating from the original work of Chen et al. \cite{chen2013pathlet}. We revisit the problem with a RL-based approach and consider additional constraints (such as edge-disjointness in pathlets and the $k$-order constraint).

\end{document}